\documentclass[11pt,table]{article}

\pdfoutput=1

\usepackage[utf8]{inputenc} % allow utf-8 input
\usepackage[T1]{fontenc}    % use 8-bit T1 fonts
\usepackage[in]{fullpage}
\usepackage{graphicx}
\usepackage{amsmath}
\usepackage{amssymb}
\usepackage{booktabs}
\usepackage[pagebackref,breaklinks,colorlinks]{hyperref}
\usepackage{url}            % simple URL typesetting
\usepackage{booktabs}       % professional-quality tables
\usepackage{amsfonts}       % blackboard math symbols
\usepackage{nicefrac}       % compact symbols for 1/2, etc.
\usepackage{microtype}      % microtypography
\usepackage{listings} % pseudocode, yell at Cade if this breaks something!
\usepackage{subcaption}
\usepackage{multirow}
\usepackage{hhline}
\usepackage{colortbl}
\usepackage{amsmath}
\DeclareMathOperator*{\argmax}{argmax} % thin space, limits underneath in displays
\usepackage{xcolor}
\usepackage{textcomp}

\definecolor{light-light-gray}{gray}{0.90} 
\newcommand{\citep}{\cite}

\usepackage[textsize=scriptsize]{todonotes}      % notes in the margins
\usepackage[capitalize]{cleveref}
\crefname{section}{Sec.}{Secs.}
\Crefname{section}{Section}{Sections}
\Crefname{table}{Table}{Tables}
\crefname{table}{Tab.}{Tabs.}

\makeatletter
\renewcommand\paragraph{\@startsection{paragraph}{4}{\z@}                                     {1.05ex \@plus1ex \@minus.2ex}                                {-.5em}
{\normalfont\normalsize\bfseries}}
\makeatother

\usepackage[binary-units=true]{siunitx}
\DeclareSIUnit\flop{FLOP}
\DeclareSIUnit[per-mode=symbol]\floppersec{\flop\per\second}

\title{Reproducible scaling laws for contrastive language-image learning}  % **** Enter the paper title here

\author{
\textbf{Mehdi Cherti}$^{1,5}$ §§ \quad \textbf{Romain Beaumont}$^1$ §§ \quad \textbf{Ross Wightman}$^{1,3}$ §§  \\ \textbf{Mitchell Wortsman}$^{4}$ §§ \quad \textbf{Gabriel Ilharco}$^{4}$ §§ \quad \textbf{Cade Gordon}$^{2}$ \\ 
\textbf{Christoph Schuhmann}$^1$ \quad  \textbf{Ludwig Schmidt}$^{1,4}$ °° \quad \textbf{Jenia Jitsev}$^{1,5}$ §§°° \\
LAION$^1$ \quad UC Berkeley$^2$ \quad HuggingFace$^{3}$ \quad University of Washington$^4$ \\ 
Juelich Supercomputing Center (JSC), Research Center Juelich (FZJ)$^{5}$  \\ 
\texttt{contact@laion.ai, \{m.cherti,j.jitsev\}@fz-juelich.de} \\
\texttt{§§ Equal first contributions, °° Equal senior contributions}
}

\begin{document}

\date{}

\maketitle

\begin{abstract}
Scaling up neural networks has led to remarkable performance  across a wide range of tasks.
Moreover, performance often follows reliable scaling laws as a function of training set size, model size, and compute, which offers valuable guidance as large-scale experiments are becoming increasingly expensive.
However, previous work on scaling laws has primarily used private data \& models or  focused on uni-modal language or vision learning.
To address these limitations, we investigate scaling laws for contrastive language-image pre-training (CLIP) with the public LAION dataset and the open-source OpenCLIP repository.
Our large-scale experiments involve models trained on up to two billion image-text pairs and identify power law scaling for multiple downstream tasks including zero-shot classification, retrieval, linear probing, and end-to-end fine-tuning.
We find that the training distribution plays a key role in scaling laws as the OpenAI and OpenCLIP models exhibit different scaling behavior despite identical model architectures and similar training recipes.
We open-source our evaluation workflow and all models, including the largest public CLIP models, to ensure reproducibility and make scaling laws research more accessible. Source code and instructions to reproduce this study will be available at \url{https://github.com/LAION-AI/scaling-laws-openclip}
\end{abstract}

\section{Introduction}
\label{sect:introduction}

Large pre-trained models now achieve state-of-the-art performance on a wide range of tasks. 
In particular, large models have led to substantial advances in speech \citep{radford2022robust}, language \citep{devlin2019bert,t5,Brown2020,hoffmann2022training}, vision \citep{Kolesnikov2020,Zhai2022scaling}, and multi-modal language-vision settings \citep{radford2021learning,ALIGN,basic,ramesh2022hierarchical,rombach2021highresolution}.
A key ingredient in these breakthroughs has been self- or weakly-supervised learning, which enabled the use of Internet-harvested training sets and reduced the need for explicit human annotation of training data.
In addition, recent pre-trained models relied on increasing the compute, model, and data scale by orders of magnitude.
% Importantly, these models are trained with self-supervised losses that do not require manual labeling of the dataset samples. This enables the use of substantially larger datasets than what is possible when using conventional supervised training, which relies on costly human-annotated labels.

\begin{table}[t!]
\setlength\tabcolsep{3.0pt}
    \centering
    \small
    \begin{tabular}{lllccc}\toprule
      & Data & Arch. & ImageNet & VTAB+ & COCO  \\\midrule
    CLIP \cite{radford2021learning} & WIT-400M & L/14 & 75.5  & 55.8  & 61.1\\
    %\midrule
%    Ours & LAION-400M & L/14 &  &  &  \\
    Ours & LAION-2B & L/14 & 75.2 &  54.6	 & 71.1 \\
    Ours & LAION-2B & H/14 & \underline{78.0} & \underline{56.4} & \underline{73.4} \\\bottomrule
    \end{tabular}
    \caption{
    We study the scaling behavior of large CLIP models using fully open-source training code and data.
    % NOTE! This next sentence here has very careful tense as to not reveal who we are.
    All models in our investigation will be made available and include the largest public CLIP models.
    %This table considers zero-shot models at 224 pixel resolution.
    This table shows zero-shot performance at 224 pixel resolution, displaying accuracy on ImageNet \cite{Deng2009a}, average accuracy on 35 VTAB+ datasets \cite{Schuhmann2022,zhai2019large}, and image retrieval recall at 5 on MS-COCO image retrieval \cite{lin2014microsoft}.
    \vspace{-1em}
    }
    \label{tab:best_perf}
\end{table}

When varying model size, compute amount, and data quantity, several papers have empirically observed that both pre-training loss and downstream task performance reliably improve with scale.
Specifically, researchers have postulated \emph{scaling laws} in the form of power law relationships between model performance and model compute, or data scale \citep{Kaplan2020,tay2021scale,Riquelme2021,Zhai2022scaling}. % which postulate a power law relationship between scale and model performance. %indicating that performance improves when increasing model, data, and training compute scale hand in hand. 
Such scaling laws allow practitioners to predict model performance for a given model and compute scale, extrapolate to larger scales, and can be used to determine pre-training regimes that obtain optimal model performance for a fixed amount of compute \citep{Kaplan2020,hoffmann2022training}.

\begin{figure}[!t]
\begin{subfigure}{\textwidth}
  \centering
  \includegraphics[width=\textwidth]{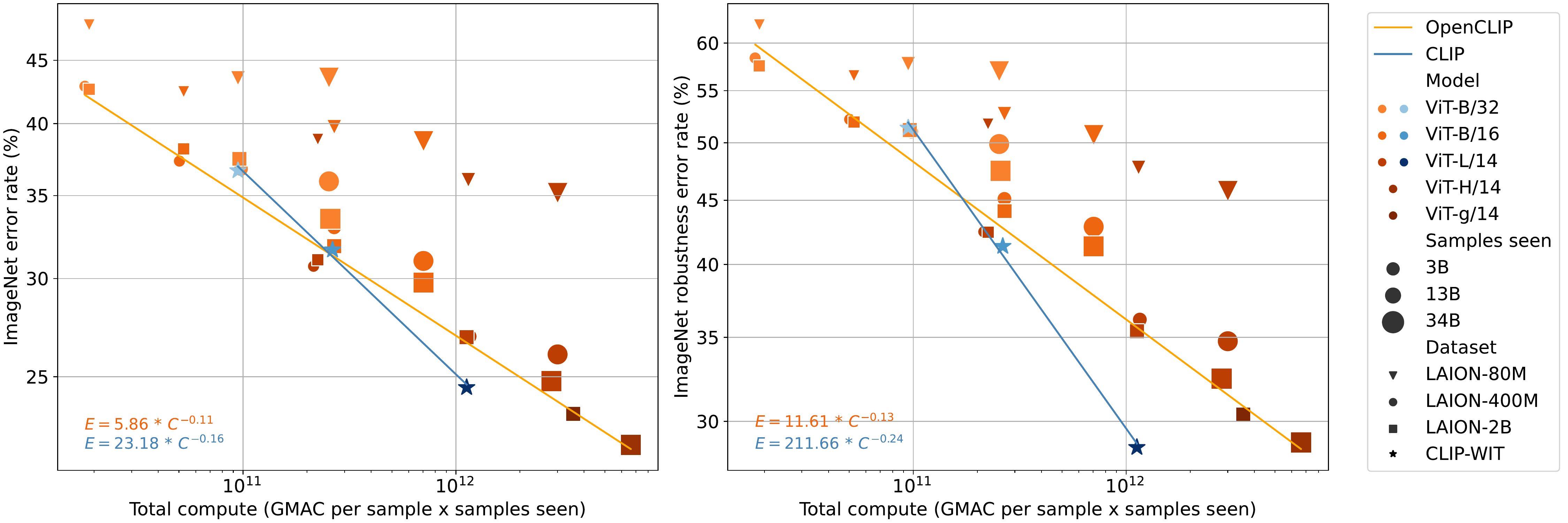}
  %\caption{Put your sub-caption here}
  \caption{Relationship between total training compute and zero-shot classification performance on downstream tasks. \textbf{Left}: ImageNet performance. \textbf{Right}: average performance on five ImageNet robustness datasets (ImageNet-V2~\cite{pmlr-v97-recht19a}, ImageNet-R~\cite{imagenetr}, ImageNet-Sketch~\cite{imagenetsketch}, ObjectNet~\cite{objectnet}, and ImageNet-A~\cite{imageneta}). Scaling model size, data size, and samples seen leads to better performance on zero-shot classification. Models trained on OpenAI's WebImageText (WIT) show a stronger scaling than models trained on LAION.}
  \label{fig:gmacs_vs_perf_imagenet_and_robustness}
\end{subfigure}
\begin{subfigure}{\textwidth}
  \centering
  % include second image
  \includegraphics[width=\textwidth]{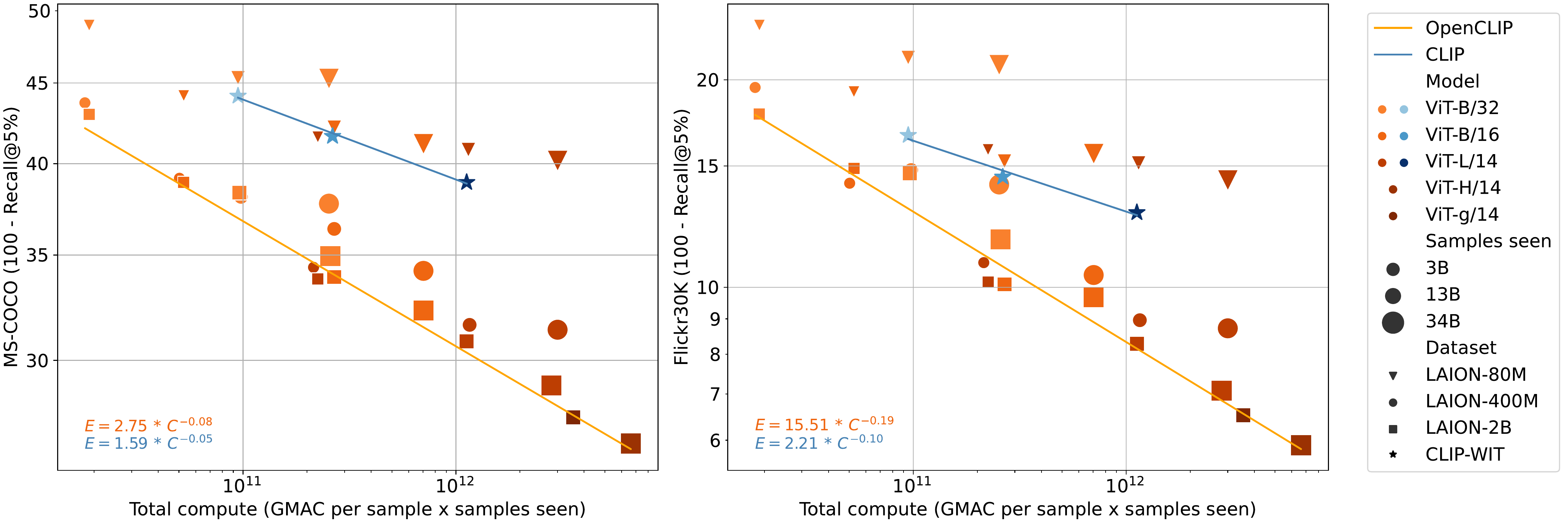}
  \caption{
  Relationship between total training compute and zero-shot image retrieval performance on MS-COCO (\textbf{Left}) and Flickr30K (\textbf{Right}). Scaling model size, data size, and samples seen leads to better performance on zero-shot image retrieval. Interestingly, in contrast to zero-shot classification (Figure ~\ref{fig:gmacs_vs_perf_imagenet_and_robustness}), models trained on LAION show a stronger scaling trend than OpenAI CLIP models trained on OpenAI's WebImageText (WIT) dataset.}
  \label{fig:gmacs_vs_perf_retrieval}
\end{subfigure}
\caption{Relationship between total training compute and performance in zero-shot classification (\ref{fig:gmacs_vs_perf_imagenet_and_robustness}) and retrieval (\ref{fig:gmacs_vs_perf_retrieval}).
We fit a power-law on the Pareto frontier of the available models. Since total compute budgets (measured in GMAC) of different trained models are not exactly aligned,  we  divide the total compute scale into bins and select the best model performance from each bin.
\vspace{-0.8em}}
\label{fig:fig1}
\end{figure}

So far, the literature on empirical scaling laws has focused on language-only \citep{Kaplan2020,tay2021scale,hoffmann2022training} or vision-only models \citep{Zhai2022scaling,Henighan2020,Riquelme2021}.
In the multimodal domain of language and vision, contrastive language-image models such as CLIP \citep{radford2021learning} have recently achieved large performance gains in zero-image classification, for instance improving zero-shot ImageNet accuracy from the prior state-of-the-art of 12\% to 76\%.
Moreover, these models demonstrate unprecedented robustness to distribution shifts compared to prior supervised models \citep{Taori2020, radford2021learning, wortsman2022robust}.
However, there is currently no systematic investigation for scaling trends in contrastive language-image learning.
One substantial challenge in this direction is that until recently, there were no datasets of sufficiently large scale openly available for the research community to undertake such experiments.

%In this work, we make use of the publicly available large-scale LAION-400M\citep{schuhmann2021laion} and LAION-5B\citep{Schuhmann2022} datasets obtained from Common Crawl\citep{common_crawl}, and the open-source OpenCLIP\citep{ilharco_gabriel_2021_5143773} implementation%,
%and compute resources obtained from academic supercomputing facilities
%to conduct a reproducible scaling laws study for contrastive language-vision learning. We pre-train OpenCLIP models, while varying model, data, and samples seen. Pre-trained models are evaluated on several downstream tasks, using zero-shot transfer, retrieval, few-shot and full dataset linear probing and fine-tuning.  We observe a consistent increase in performance when scaling model, data, and samples seen, and derive scaling laws of power law form across different downstream tasks. Interestingly, when comparing OpenCLIP and OpenAI CLIP\citep{radford2021learning} models, we find larger scaling coefficients for OpenCLIP models on zero-shot retrieval, while OpenAI CLIP models show stronger scaling for zero-shot classification. Table \ref{tab:best_perf} shows a highlight of two of our models and their results on image classification and retrieval benchmarks.

In this work, we conduct a scaling laws study for contrastive language-vision learning by utilizing the recently released LAION-5B \citep{Schuhmann2022} dataset of 5 billion image-text pairs.
To ensure that our experiments are fully reproducible, we use the open source OpenCLIP \citep{ilharco_gabriel_2021_5143773} code to train CLIP models while varying model, data, and samples seen.
We evaluate our CLIP models on several downstream tasks, including zero-shot classification, image retrieval, and fine-tuning via linear probing and end-to-end optimization.
We observe a consistent increase in performance when scaling model, data, and compute, and derive scaling laws of power law form across different downstream tasks (Figure~\ref{fig:gmacs_vs_perf_imagenet_and_robustness}, ~\ref{fig:gmacs_vs_perf_retrieval}).
Interestingly, when comparing our OpenCLIP and OpenAI’s original CLIP models, we find larger scaling coefficients for OpenCLIP models on zero-shot retrieval, while OpenAI CLIP models show stronger scaling for zero-shot classification.
Table \ref{tab:best_perf} shows two of our models and their results on image classification and retrieval benchmarks.

% We observe consistent increase of performance across various downstream tasks when increasing model, data, and samples seen. We observe scaling laws of power law form across different downstream tasks.

% By comparing OpenCLIP and OpenAI CLIP\citep{radford2021learning} models, we observe task-dependent scaling laws of power law form. Scaling coefficients are larger for OpenCLIP models on retrieval, while OpenAI CLIP models show higher slope on zero-shot transfer. 

%Since model architectures and pre-training recipes are largely matched, we hypothesize that the training dataset is responsible for the task-dependent differences in scaling behavior we observe.
%We discuss the results and outline further opportunities provided by large-scale open datasets for reproducible studies of CLIP models at various scales. We conclude with an outlook on constructing foundation datasets\citep{sorscher2022beyond} that can be extracted from unfiltered data crawled from the internet.

We hypothesize that the training dataset is responsible for the task-dependent differences in scaling behavior between the OpenCLIP and OpenAI models.
Our experiments have used the same ViT architectures as the OpenAI models, and the training recipes are largely matched.
The main difference in training recipes is the batch size due to different compute environments, and our experiments with varying batch sizes suggest that the batch size changes do not explain the change in scaling trends.

Overall our findings highlight the design of pre-training datasets as an important direction to further improve image-text models.
Dataset designers should measure scaling behavior so that the generalization capabilities of image-text models can continue to improve as we increase model size and the amount of compute.
Moreover, pre-training datasets should be evaluated on a broad range of downstream tasks because model scaling can differ substantially by task with different pre-training sources leading to different scaling behavior by task.
We hope that our open-source and reproducible scaling trends offer concrete starting points for improving current image-text datasets and models.

\section{Background and related work}
\label{sec:background}

\paragraph{Scaling laws for generalization and transfer.} Strong empirical evidence that increasing model or data scale is beneficial was initially studied in the context of deep learning and computer vision \citep{Sun2017,hestness2017deep}. For instance, in~\citep{hestness2017deep}, the power law relation between scale and model performance was highlighted. Empirical work stimulated theoretical studies that provided justification for the observed generalization boost with scale, investigating generalization error in overparameterized networks in the interpolation regime \citep{Belkin2019,Bubeck2021}.

Early empirical studies focused on the effect of training scale on upstream performance, measuring the test loss from the same distribution used for training. 
Subsequent studies of large language models such as GPT-3 \citep{Brown2020} demonstrated broad generalization capabilities in models with substantially larger scale. Moreover, neural scaling laws of the power law form were derived for language models, connecting model, data, and training compute scale to performance \citep{Kaplan2020,tay2021scale,hoffmann2022training}. This also allowed accurate prediction of model performance at larger scales, and researchers were able to determine the scale parameters for achieving optimal performance given a fixed amount of compute \citep{hoffmann2022training,koppula2022should}. Scaling law studies were then also studied in the vision domain \citep{Riquelme2021,Zhai2022scaling}, also observing a power law dependency of performance on scale.   

Scaling law studies were also conducted for transfer and out-of-distribution performance \citep{Kaplan2020,tay2021scale,Zhai2022scaling}. In these studies, researchers observed that performance on downstream tasks benefits from increasing model, data, and training compute scale \citep{Kolesnikov2020, Brown2020, Kaplan2020, Zhai2022scaling}. Interestingly, upstream performance does not always correlate with downstream performance \citep{tay2021scale,tay2022scaling}. Since downstream performance most accurately reflects a practical use cases, examining scaling behavior on downstream tasks is increasingly important. Recent work has also studied the effect of scale on other model characteristics, such as performance after pruning and compression \citep{pmlr-v139-rosenfeld21a, chen2021lottery} and on susceptibility to catastrophic forgetting \citep{ramasesh2021effect}.   

%Also Transfer, effect of scale on downstream tasks; out of distribution.  Big Transfer;  Transfer Laws; up stream and downstream NOT necessarily related! paper  other functionality effect of scale. Beyond scaling laws ...

% Consistent reports by empirical studies that larger model and data scale help tremendously with generalization and robustness motivated further theoretical studies. 

% Evidence that increasing training scale may be beneficial for performance of obtained models came already from early empirical studies in vision learning. Then GPT language, then scaling laws motivated by observed strong boost on generalization. Upstream test loss. But also on transfer, out of distribution, downstream performance. Optimal compute allocation for a performance, or prediction of performance when compute is a free choice. 

\paragraph{Scaling up language-vision learning.} Learning from very large amounts of weakly aligned image-text pairs has led to the development of models with broad generalization capabilities.
Notably, work on contrastive language-image pre-training (CLIP \citep{radford2021learning}) showed dramatic improvement compared to the previous state-of-the-art in zero-shot transfer and unprecendented robustness to distribution shift \citep{Taori2020,miller2021accuracy,nguyen2022quality,fang2022data}. 
The success of the initial CLIP study, which used a private WIT-400M image-text pairs dataset and ViT-L/14 as the largest scale vision encoder, motivated further developments and numerous extensions that increased model and data scale. ALIGN \citep{ALIGN} used a private dataset of 1.8B text-image pairs and a large EfficientNet-L2 as an image encoder. BASIC \citep{basic} employed a large CoAttNet-7 model with 2.4B parameters for the image encoder, also further increasing dataset size up to 6.6B image-text pairs, using supervised visual encoder pre-training and private datasets (ALIGN and JFT-5B). LiT \citep{zhai2021lit} used a private dataset of 4B image-text samples for contrastive learning on a total of 18B samples, scaling the visual encoder up to ViT-g/14, which was pre-trained in a supervised manner using another private dataset (JFT-3B). CoCa \citep{yu2022coca} used ViT-g/14 as a visual encoder and both the ALIGN and JFT private datasets, and an additional text captioning loss based on autoregressive language modeling during pre-training. LiMoE \citep{mustafa2022multimodal} trained a sparse mixture-of-experts (MoE) single tower architecture that share a common backbone for both vision and text using both private 3.6B image-text data from LiT and JFT-4B \citep{Zhai2022scaling}, obtaining a ViT H/14 model at the largest scale. Flamingo \citep{alayrac2022flamingo} uses a large private interleaved image-text dataset, using NFNet-F6 as a visual encoder while scaling up the text encoder from 1.4B to 70B parameters. PaLI \citep{chen2022pali} trained a multi-language multi-task text-image model using ViT-e (4B parameters) as a visual encoder and mT5-XXL (13B parameters) as a text encoder, trained on a private dataset (WebLI) with 29B image-text pairs. While these studies already show clear merits of scaling up, they do not conduct a thorough scaling investigation by systematically scaling model, data and, training compute. Moreover, most studies involve a customized multi-stage training procedure, where encoders may be pre-trained separately with uni-modal losses, and then tuned further with a contrastive image-text loss, while also potentially freezing one of the encoders \citep{basic,zhai2021lit}. This makes it difficult to derive conclusions about the effect of scale as pre-training procedures are heterogeneous. In addition, the private nature of the employed datasets impairs reproduction and validation of the results, especially in cases where pre-trained models are also not publicly available.

\paragraph{Open large-scale language-vision datasets.} Conducting scaling law studies requires sufficiently large pre-training datasets. Earlier efforts to provide open image-text datasets like MS-COCO \citep{lin2014microsoft}, Visual Genome \citep{krishna2017visual}, YFCC-100M \citep{thomee2016yfcc100m}, Conceptual Captions CC3M and CC12M \citep{sharma-etal-2018-conceptual,hu2021scaling} do not match the current scale of private data used to train large-scale language vision models. More recently, larger image-text datasets have been collected from Common Crawl \citep{common_crawl}. The resulting datasets, LAION-400M \citep{schuhmann2021laion} and LAION-5B \citep{Schuhmann2022} are publicly available, enabling training language-vision models at larger scale \citep{rombach2021highresolution, mustafa2022multimodal, ho2022imagen}. Using the LAION toolset \citep{Schuhmann2022}, it also became possible to construct additional open datasets, such as COYO-700M \citep{kakaobrain2022coyo-700m}. 

\section{Datasets and Methods} 
\label{sec:datasets_methods}

\subsection{Open large-scale datasets LAION-400M/2B} 
\label{subsect:pretrain_laion}
% Re-iterate here the content regarding dataset origins and its content (re-phrasing already existing content we did). Refer to NeurIPS paper.
% TODO : Ludwig, Romain, Richard, Christoph

We use the LAION-400M \citep{schuhmann2021laion} and LAION-5B \citep{Schuhmann2022} datasets which are open, public image-text datasets validated by the pre-training of state-of-the art multi-modal models such as CLIP \citep{radford2021learning} and Stable Diffusion \citep{rombach2021highresolution}. LAION-5B contains an English image-text subset of 2.32 billion samples, which we refer to as LAION-2B in this work.  Due to its scale, transparency and open-source nature, LAION has already been adopted by various works on language-vision modelling, validating its suitability for systematic scaling law studies.

\subsection{Pre-training OpenCLIP across various scales}
\label{subsect:pretrain}
%Re-iterate here the content regarding openCLIP pre-training, distributed training (re-phrasing already existing content we did). Refer to NeurIPS paper.
%TODO : Mehdi, Ross, Cade, Ludwig, Romain

To systematically vary model scale, data scale and the number of samples seen during pre-training, we selected a scale range for each dimension. For model scale, we choose CLIP architectures with ViT-B/32, ViT-B/16, ViT-L/14, ViT-H/14 and ViT-g/14 as visual encoders, scaling the text encoder in accord (see Appendix Table ~\ref{table:openclip_architectures}). For data scale, we use LAION-80M (an 80M subset of LAION-400M), LAION-400M, and LAION-2B. For training duration, we choose 3B, 13B and 34B samples seen scales. Due to compute constraints, for the larger H/14 and g/14 model scales, we conduct only restricted measurements (done for LAION-2B, with 34B samples seen for H/14, and with 13B samples seen for g/14). This selection provides coverage at the scale where we cannot afford to sample with the same density as at the intermediate  and lower model scales. To verify that LAION-80M and LAION-400M are valid subsets of LAION-2B, we conduct a control experiment by extracting a random 400M subset of LAION-2B and comparing our reference OpenCLIP ViT-B/32 models pre-trained on both datasets. When doing so, we found no significant difference (see Appendix Sec. \ref{sec:appendix:control_exp}).

% \citep{ilharco_gabriel_2021_5143773}

Compared to the original CLIP training procedure \citep{radford2021learning}, we work with larger batch sizes and adapt the learning rate accordingly.
We opt for larger batch sizes to allow for more efficient distributed training; maximizing the local batch size per GPU and using close to one thousand GPUs lead us to global batch sizes in the range of 86-88K samples.
% There are two motivations for larger batch sizes. First, larger batch sizes allow for more efficient distributed training; maximizing the local batch size per GPU and using close to one thousand GPUs lead us to global batch sizes in the range of 86-88k samples. Second, larger batch sizes tend to improve performance for contrastive image-text learning\cite{radford2021learning, basic}.
In order to assess the validity of re-using measurements obtained with different batch sizes, we perform a number of control experiments varying batch size from 32K to 86-88K, and observe a difference of $0.2-0.5\%$ across different settings (see Appendix Sec. \ref{sec:appendix:control_exp}), which is small enough not to confound observations on the effect of scale.

For each number of samples seen scale, we execute a separate training experiment with a cosine annealing learning schedule adapted to the number of samples. This allows us to assess performance of models pre-trained with different training durations and avoid suboptimal training when using the same schedule for runs of different length~\citep{hoffmann2022training}. We tune a small number of hyper-parameters (see Appendix Table ~\ref{table:openclip_hyper_parameters}), each scale point to optimize validation loss and prevent training instabilities, and otherwise closely follow the original CLIP training procedure \citep{radford2021learning}, using the InfoNCE loss, Adam with decoupled weight regularization \citep{loshchilov2018decoupled} (i.e., AdamW) as an optimizer, with $\beta_1=0.9$, $\beta_2=0.98$ and weight decay of $0.2$.
We train the models using mixed precision.
%\todo{Mehdi: I think we should mention what we observed with loss explosion and fixing it with bfloat16, it was a good insight}
%\todo{Jenia: Agree - but depending on space constraint, we might shift it to Appendix}
%With the largest models (ViT-H/14, ViT-g/14), we observed loss spikes during training which affected training performance heavily. We first tried to fix the issue by modifying the learning rate schedule, the learning rate, and by adding gradient clipping, without success. 
For larger model scales (ViT-L/14, H/14, g/14), we observed loss spikes during training which had an adverse effect on performance. We fixed the issue by switching from mixed precision with float16 to mixed precision with bfloat16.\footnote{We also tried to reduce the learning rate, change the learning rate schedule, and use gradient clipping but none of these changes helped to avoid the training instabilities.} We hypothesize that bfloat16 fixed the issue due to larger models typically showing larger activation values as observed by~\cite{dettmers2022llm}, making bfloat16 more suitable with its wider dynamic range (8 exponent bits). %We then adopted bfloat16 for the remaining models that we train. To make sure there is no performance difference between float16 and bfloat16 on the smallest models, we do a control experiment with ViT-B/32 and observe no performance difference (see Appendix).

% all experiments are full runs with AN OWN LR ANNEALING SCHEDULE adapted to the number of samples seen (3B, 13B or 34B samples).  It avoids the mistake pointed out in Chinchilla paper, where Neural Scaling Laws work was taking early points from the same schedule of a longer run to provide measurements for performance with less samples seen. We do for each samples seen scenario a separate run with ITS OWN LR SCHEDULE

%%% PLUGGING MAIN FIGURE HERE IS POSSIBLE

CLIP pre-training experiments on larger scales require distributed training, as otherwise experiment execution times are intractable. We use OpenCLIP~\citep{ilharco_gabriel_2021_5143773}, an open source software that was adapted for distributed training on supercomputers. Using data parallel training via PyTorch DDP~\cite{li2020pytorch, paszke2019pytorch}, we conduct experiments with up to 1520 NVIDIA A100 GPUs. Distributed training was executed on JUWELS Booster~\citep{JUWELSBooster2020}, the supercomputer at Juelich Supercomputing Center (JSC, Germany), and partly also at Stability AI AWS supercomputer~\citep{stability_hpc} For more details on distributed training procedure and on experiment compute budgets and runtimes, see Appendix Sec.\ref{appendix:distributed_training} and Sec. \ref{sec:appendix:exp_further_results}.

\section{Scaling laws for different downstream tasks}
\label{subsect:scaling_laws}

\subsection{Zero-shot transfer and robustness}
\label{subsect:scaling_laws_zeroshot_transfer}

One of the biggest advantages of open-vocabulary models like CLIP is that they can be used on downstream classification tasks by carefully designing text prompts corresponding to class descriptions, without requiring any labeled training example. Moreover, pre-trained CLIP models are observed to excel on out-of-distribution robustness benchmarks \citep{radford2021learning,miller2021accuracy}. In this section, we study the effect of scale on zero-shot classification, including an investigation on robustness benchmarks. We evaluate the models on ImageNet \citep{Deng2009a}, ImageNet distribution shift datasets \citep{imagenetr,imagenetc,imageneta,imagenetsketch,objectnet}, and the visual task adaptation benchmark (VTAB)~\cite{zhai2019large}. We conduct a simple duplication check for downstream datasets based on the perceptual image hash library pHash~\cite{zauner2010implementation}, revealing no or very little overlap with pre-training datasets (see Appendix Sec. \ref{sec:appendix:datasets}).

\paragraph{Evaluation setup.}
%TODO : Mehdi, Mitchell, Gabriel, Ludwig
We follow the setup of Radford \textit{et al.}~\cite{radford2021learning}. For each downstream dataset, we use a set of pre-defined prompts for each class, which we collected from prior works~\cite{radford2021learning,zhai2021lit}.
We compute the embedding of each class by averaging over the embeddings of the prompts obtained using the text tower, then we L2-normalize them. 
Given a dataset $\{(x_i,y_i)\}_{i=1}^n$, we classify each image as the class that has the largest cosine similarity with the (L2-normalized) image embedding, $\hat{y}_i = \argmax_j (\phi(x_i)^Tc_j)$. We evaluate the models using top-1 accuracy. For comparison to OpenAI CLIP, we take ViT-B/32, B/16, and L/14 models pre-trained on the private WIT-400M dataset.

\textbf{Effect of scale.} Accuracy consistently improves when increasing model, data and samples seen scale hand-in-hand. Accuracy follows power laws, such that larger models benefit from larger data and samples seen scale (Figure \ref{fig:gmacs_vs_perf_imagenet_and_robustness}). The strongest ImageNet accuracy (78\%) is obtained with the largest total pre-training compute, using ViT-H/14 pre-trained on LAION-2B data scale and 34B samples seen. For additional results, see Appendix Sec. \ref{sec:appendix:exp_further_results} and \ref{sec:appendix:further_results} (for instance, additional evidence showing consistency of scaling trends for CLIP and openCLIP observed here across various tasks independent of vision tower architecture in Appendix Fig. \ref{fig:resnet_scaling_trends}, further supporting the results from Fig.~\ref{fig:fig1})

%Im1k
%C=-0.12 (openCLIP)
%C=-0.16 (openAI)

%ImageNet-Robustness
%C=-0.12 (openCLIP)
%C=-0.24 (openAI)
% General message: scale improves performance as expected. Power law form observed. CLIP higher slope/stronger scaling than openCLIP. Improvement on accuracy on ImageNet goes hand in hand with improvement in (effective?) robustness. 

Fitting power-law ($E = \beta C^{\alpha}$) on the Pareto frontier of the available models, we measure scaling coefficients $\alpha_{\text{openCLIP}}=-0.11$ and $\alpha_{\text{CLIP}}=-0.16$ for zero-shot top-1 ImageNet and $\alpha_{\text{openCLIP}}=-0.13$ and $\alpha_{\text{CLIP}}=-0.24$ for ImageNet robustness datasets performance \citep{imagenetr,imagenetc,imageneta,imagenetsketch}. For those tasks, we observe a scaling advantage for CLIP pre-trained on WIT-400M over OpenCLIP pretrained on LAION-400M/2B. $\alpha_{\text{openCLIP}}$ is similar for ImageNet and robustness datasets, suggesting that improving accuracy with scale leads to corresponding improvement on robustness benchmarks for OpenCLIP pre-trained on LAION.

We also find bottleneck effects when scaling. For instance, OpenCLIP ViT-B/32 and ViT-B/16 models show no change or deterioration of performance when increasing data scale from 400M to 2B when using a smaller samples seen scale (3B or 13B). Moving to the largest samples seen scale (34B) then shows clear improvement for the larger 2B data scale, indicating that the number samples seen is a bottleneck (see also Appendix Table \ref{table:zeroshot_im1k}).

% Sec. \ref{sec:appendix:further_results})

% : drop from 58.73\% to 57.60\%;
% : drop from 62.90\% to 

Using the obtained power law, we can make a prediction for the performance of a well-tuned ViT-g/14 model when using the largest data scale of 2B and samples seen scale of 34B, giving us error estimate of 20.9\% (79.1\% top-1 accuracy) on ImageNet. We predict even stronger performance at larger scales. For instance, assuming 68B samples seen we estimate top-1 accuracies of 79.7\%, 80.7\%, and 81.9\% for ViT-H/14, ViT-g/14 and ViT-G/14, respectively (see also Appendix Sec. \ref{sec:appendix:predictions}).

% TODO: make a statement on when same performance can be reached by a smaller model, how much do we predict for samples seen / data scale / compute to be necessary (say L/14 ? 68B? 34B samples seen?, or H/14)
%TODO : Mehdi, Mitchell, Gabriel, Ludwig

%\begin{figure}[]
%  \centering
%  \includegraphics[width=\textwidth]{images/gmacs_vs_perf_imagenet_and_robustness.pdf}
%  \caption{Relationship between total training compute and zero-shot classification performance on downstream tasks. \textbf{Left}: ImageNet performance. %\textbf{Right}: average performance on five ImageNet robustness datasets. Scaling model size, data size, and samples seen budget lead to better performance on %zero-shot classification.}
%  \label{fig:gmacs_vs_perf_imagenet_and_robustness}
%\end{figure}
\subsection{Retrieval}
\label{subsect:scaling_laws_retrieval}

Retrieval is another common way to evaluate zero-shot capabilities of the models. In this section, we study the effect of scale on both text and image zero-shot retrieval.

\paragraph{Evaluation setup.}
%TODO : Mehdi, Romain, Mitchell, Gabriel, Ludwig
We compute text-image scores using the cosine similarity between image and text embeddings and rank the top-K images (resp. text captions) for each text caption (resp. images) when evaluating on image (resp. text) retrieval.
%We rank the top-K images (resp. text) for each text caption (resp. image) when evaluating image (resp. text) retrieval.
We evaluate on MS-COCO \citep{lin2014microsoft} and Flickr30K \citep{flickr30_young2014image}, following the evaluation setup and test splits from~\cite{karpathy2015deep}. We use Recall@K as an evaluation metric where $K=5$.%, measuring the fraction of times the correct item is retrived among the top-k results.

\paragraph{Effect of scale.} Again we observe performance consistently improves when increasing scale following power law trends (Figure \ref{fig:gmacs_vs_perf_retrieval}). We measure scaling coefficients $\alpha_{\text{openCLIP}}=-0.08$ and $\alpha_{\text{CLIP}}=-0.05$ for zero-shot retrieval on MS-COCO and $\alpha_{\text{openCLIP}}=-0.19$ and $\alpha_{\text{CLIP}}=-0.10$ for Flickr30K. In contrast to zero-shot accuracy, retrieval performance shows a scaling advantage for OpenCLIP pre-trained on LAION-400M/2B over CLIP pre-trained on WIT-400M. We also observe scale bottleneck effects. For instance, OpenCLIP ViT-L/14 model shows almost no improvement on LAION-400M when increasing the number of samples seen scale from 13B to 34B, indicatating a data scale bottleneck. When increasing data scale to 2B, we then observe clear improvements when going from 13B to 34B samples (see also Appendix Table \ref{table:zeroshot_mscoco_image_retrieval} and \ref{table:zeroshot_mscoco_text_retrieval}).

\begin{figure}
    \centering
    \includegraphics[width=\textwidth]{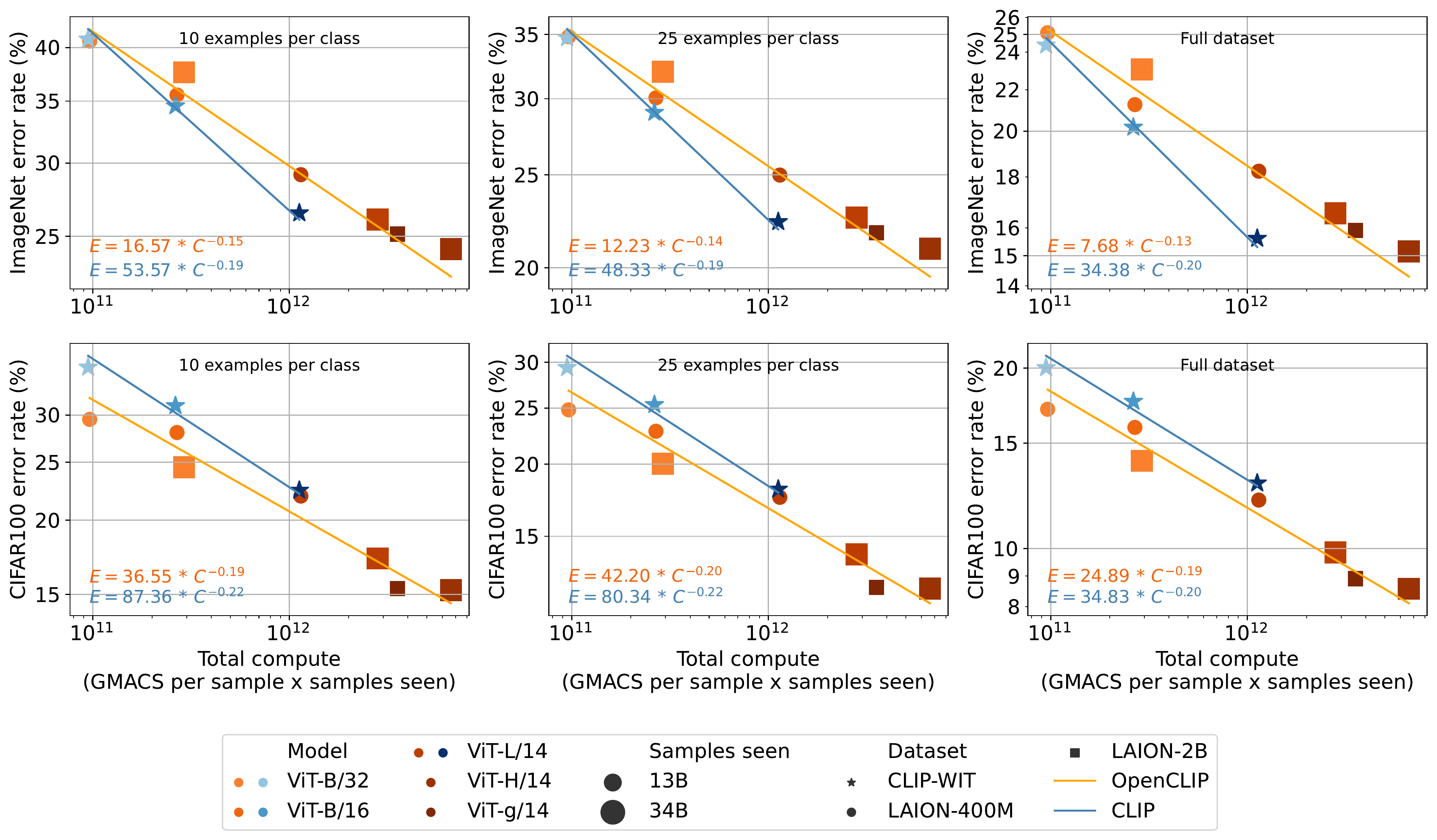}
    \caption{Scaling model and data size leads to lower error linear classifers on ImageNet~\cite{Deng2009a} and CIFAR-100~\cite{Krizhevsky2009} in both the few-shot and full dataset regime.
    We train linear probes for models with at least 13B samples seen (also see corresponding Table~\ref{tab:lp}).
    As discussed in Figure~\ref{fig:fig1}, we fit a power-law on the Pareto frontier of the available models.}
    \label{fig:lp1}
\end{figure}
\begin{figure}
    \centering
    \includegraphics[width=0.6\textwidth]{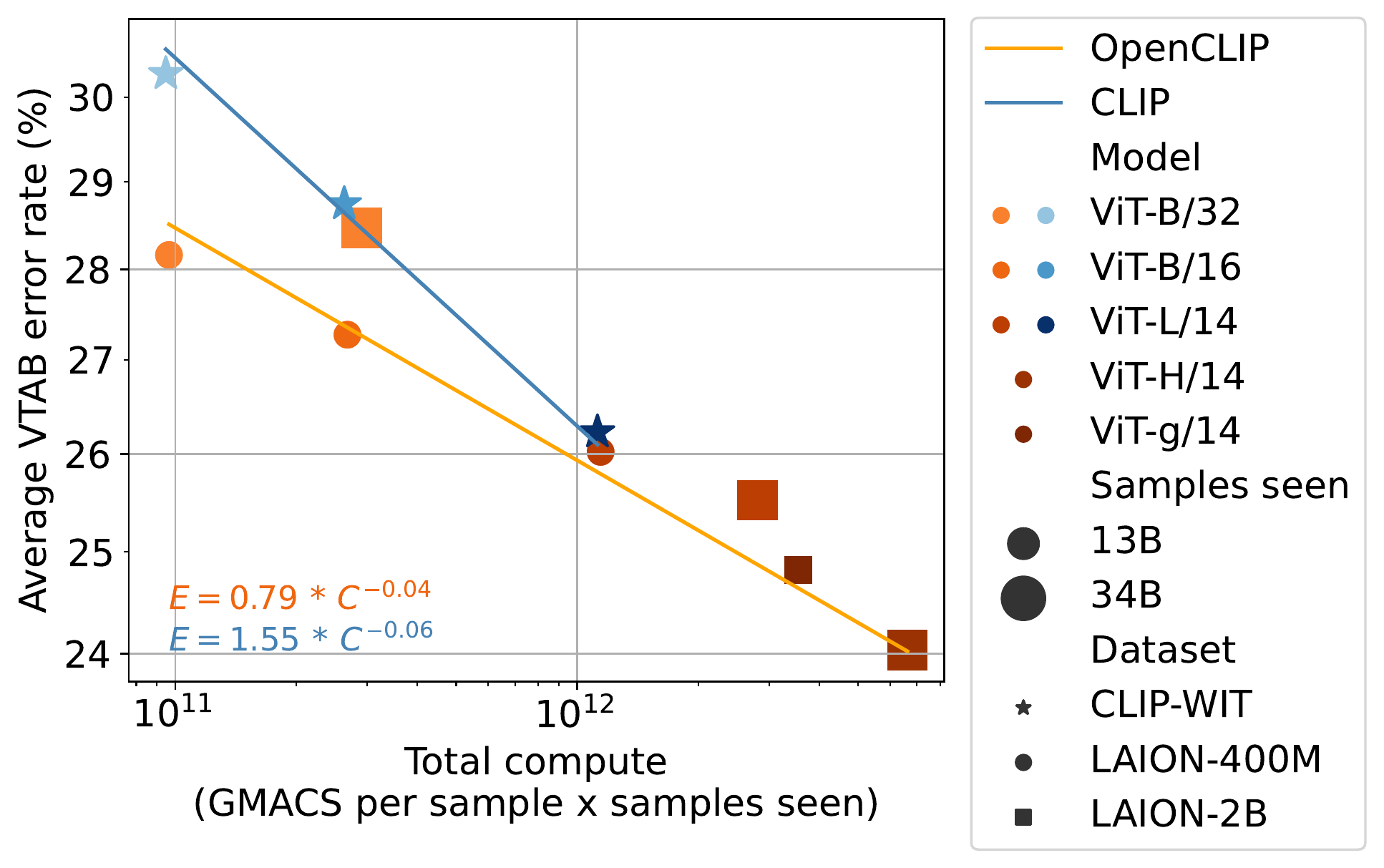}
    \caption{Scaling model and data size leads to lower error linear classifers on the visual task adaptation benchmark (VTAB)~\cite{zhai2019large}.
     We train linear probes for models with at least 13B samples seen (also see corresponding Table~\ref{tab:lp}).
    As discussed in Figure~\ref{fig:fig1}, we fit a power-law on the Pareto frontier of the available models.}
    \label{fig:lp2}
\end{figure}
%TODO : Mehdi, Romain, Mitchell, Gabriel, Ludwig

% 
%Im1k
%C=-0.09 (openCLIP)
%C=-0.05 (openAI)

%ImageNet-Robustness
%C=-0.20 (openCLIP)
%C=-0.10 (openAI)

% General message: scale improves performance as expected. Power law form observed. This time openCLIP higher slope/stronger scaling than CLIP - in contrast to zero-shot classification. 
% Sec. \ref{sec:appendix:further_results}). 

% The strongest performance is measured when going to largest total pre-training compute (ViT-H/14, LAION-2B data scale and 34B samples seen).   

%\begin{figure}[]
%  \centering
%  \includegraphics[width=\textwidth]{images/gmacs_vs_perf_retrieval.pdf}
%  \caption{
%  Relationship between total training compute and zero-shot retrieval performance on MS-COCO (left) and Flickr30K (right). Scaling model size, data size, and %samples seen budget lead to better performance on zero-shot retrieval. Interestingly, contrary to zero-shot classification (Figure ~\ref{fig:gmacs_vs_perf_imagenet_and_robustness}), models trained on LAION consistently improve over OpenAI CLIP models trained on OpenAI's WIT.}
%  \label{fig:gmacs_vs_perf_imagenet_and_robustness}
%\end{figure}

\subsection{Full and few-shot linear probing}
\label{subsect:scaling_laws_linearprobe}

Another common way to measure the quality of learned representations is by training a linear classifier.
While this technique underperforms end-to-end fine-tuning, it is often preferred because it requires far less compute~\cite{kornblith2019better, radford2021learning}.
In this section we train linear classifiers, also referred to as linear probes, on the frozen representations of various CLIP models and examine the effect of data and model scale.

\paragraph{Evaluation setup.}
Given a CLIP model with an image tower $\phi$, our goal is to learn $W$ such that $W^\top\phi(x)$ classifies $x$ as its label $y$.
Given a dataset $\{(x_i,y_i)\}_{i=1}^n$, we begin by saving the image features and labels for the dataset.
That is, for all image label pairs $(x, y)$ in the dataset we cache $(\phi(x), y)$.
We then train a linear classifier $W$ to minimize the cross entropy loss between $\mathsf{softmax}\left(W^\top \phi(x) \right)$ and $y$.
In preliminary experiments we found that this softmax regression achieved higher accuracy than linear regression.
We use mini-batch stochastic optimization with the Adam optimizer~\cite{kingma2014adam}.
We use batch size 256 and select the best result in a hyper-parameter sweep over learning rate $\{0.1, 0.01, 0.001\}$ and epochs $\{10, 20, 40\}$ individually for each model and dataset.
For the ImageNet~\cite{Deng2009a} and CIFAR100 datasets~\cite{Krizhevsky2009} we consider 10-shot, 25-shot, and full-dataset linear classifers (Figure~\ref{fig:lp1}).
Additionally, we train linear classifiers on the visual task adaptation benchmark (VTAB)~\cite{zhai2019large} (Figure~\ref{fig:lp2}).

\paragraph{Effect of scale.}
For ImageNet, CIFAR100, and VTAB, scaling up consistently improves the accuracy of a linear classifier (Figure~\ref{fig:lp1}, \ref{fig:lp2}).
For ImageNet and CIFAR100, this is true in both the few-shot and full regimes.
Moreover, among models trained on the same data distribution, scaling up follows a linear trend on a log-log plot.
These results are perhaps not too surprising given similar observations for power laws on zero-shot downstream tasks in Section~\ref{subsect:scaling_laws_zeroshot_transfer} as well as the correlation between zero-shot and linear probe performance observed by Radford \textit{et al.}~\cite{radford2021learning}.
Nonetheless, this result re-affirms that scaling up model and data size leads to contunied accuracy improvements.

\subsection{Fine-tuning}
\label{subsect:scaling_laws_finetune}

Next, we evaluate the effect of scale on fine-tuning performance. Since fine-tuning is much more compute-intensive than zero-shot and linear probing, we only evaluate a subset of the pre-trained models.

\begin{figure}[!th]
    \centering
    \includegraphics[width=\linewidth]{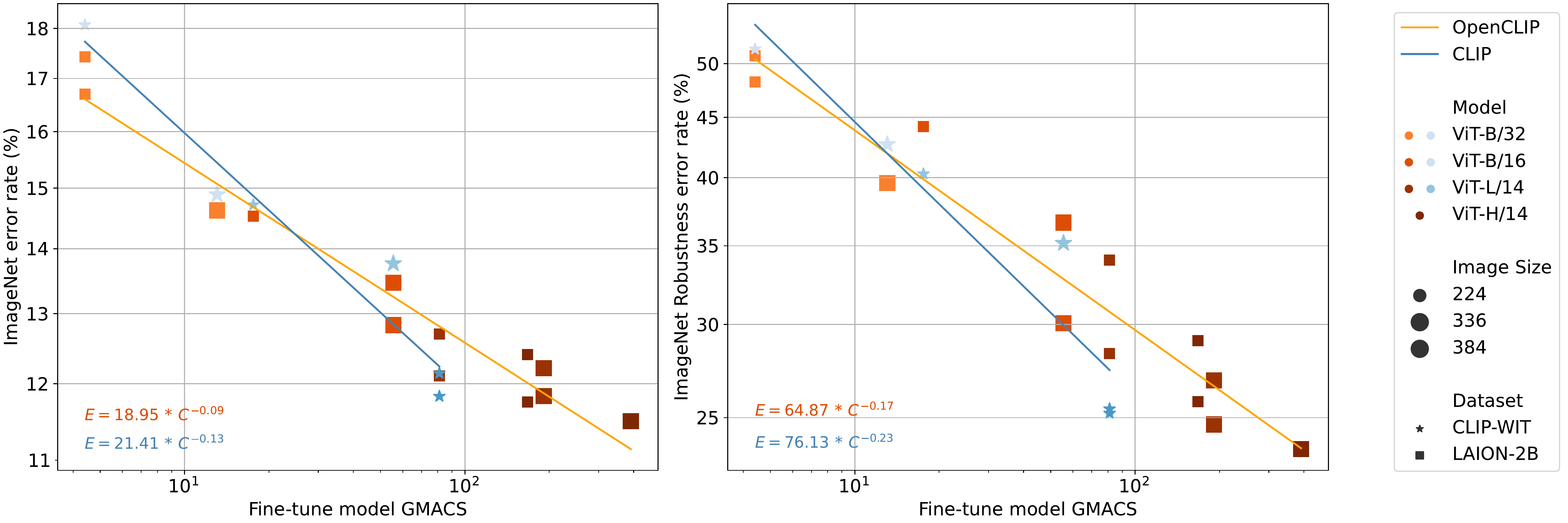}
    % \vspace{-1em}
    \caption{
    ImageNet and ImageNet robustness datasets classification performance for fine-tuned models.}
    \label{fig:ft-imagenet}
\end{figure}
\begin{figure}[!th]
    \centering
    \includegraphics[width=\linewidth]{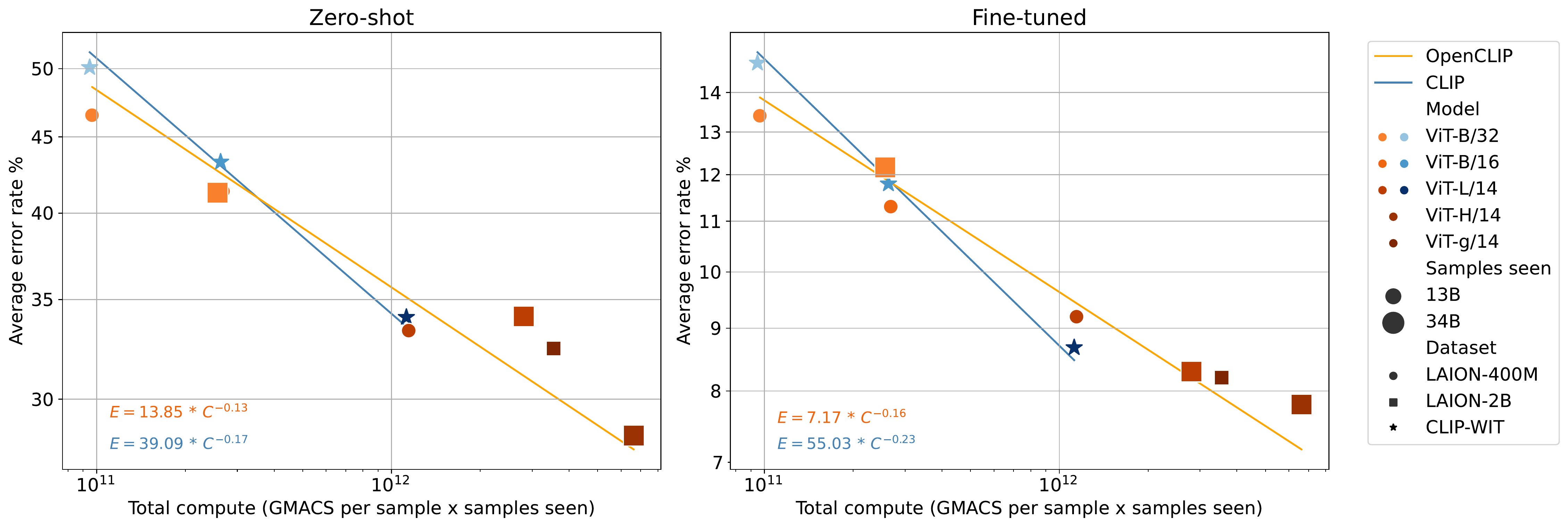}
    \caption{Scaling model and data size leads to lower error after jointly fine-tuning on eight downstream image classification tasks.
    In this experiment, we fine-tune a single model jointly on all eight tasks, alternating batches from each task.
    We fine-tune only the parameters of the vision encoder, using a fixed classification head for each task initialized with the weights from the zero-shot model.}
    \label{fig:ft-eight}
\end{figure}

%TODO : Ross, Gabriel, Mitchell, Mehdi, Ludwig
\paragraph{Evaluation setup.} We fine-tune and evaluate on ImageNet with the \textbf{timm} \citep{timm} library, using the image encoder from CLIP models trained on 2B data, 34B samples seen scale. To get the best results, we consider two different schemes, (A) fine-tune directly on ImageNet (B) first fine-tune on a subset of the full ImageNet-22k we call ImageNet-12k\footnote{We filter classes with few examples from the full ImageNet-22k with 14M examples to get a better balanced subset and we end up with 12K classes, 12M training examples, 470K validation examples.} then continue fine-tuning on ImageNet, similar to~\cite{bao2021beit}. We compare the results with OpenAI CLIP models fine-tuned with the same settings, evaluating the models using top-1 accuracy on ImageNet and the ImageNet distribution shift datasets \citep{imagenetr,imagenetc,imageneta,imagenetsketch,objectnet}. The OpenCLIP models range from 82.6 to 88.5\% top-1 on ImageNet, comparable to the best released ImageNet models pretrained on public datasets\cite{bao2021beit}. For additional details, including strong supervised baselines, see Appendix sec. \ref{sec:appendix:finetuning}.

%\todo{Ross: should we mention interesting tidbits like the relative stength on ImageNet-R and Sketch vs weakness on A and this being similar to the zero-shot case?  }
%\todo{Mehdi: this is just from my understanding, @Ross / @Gabriel please update/correct/extend. also, patching description is missing}
% Model patching is explained in Appendix, Section B.2.2 Fine Tuning

In addition, we fine-tune and evaluate on eight diverse datasets where zero-shot models perform poorly \cite{radford2021learning,ilharco2022patching}: Cars \citep{cars}, DTD \citep{dtd}, EuroSAT \citep{eurosat}, GTSRB \citep{gtsrb}, MNIST \citep{lecun1998mnist}, RESISC45 \citep{cheng2017remote}, SUN397 \citep{sun397}, and SVHN \citep{svhn}. We fine-tune a single model jointly on the eight downstream tasks following Ilharco et al. \cite{ilharco2022patching}, fine-tuning only the parameters of the vision encoder. The classification heads for each task are obtained using the zero-shot text encoder, and are kept frozen during fine-tuning.
We fine-tune for 2000 iterations with a batch size of 128, learning rate 1e-5 and a cosine annealing learning rate schedule with 200 warm-up steps and the AdamW optimizer \citep{loshchilov2018decoupled}, with weight decay 0.1.
We further explore the effect of fine-tuning on zero-shot ImageNet accuracy in the Appendix Sec. \ref{sec:appendix:finetuning}. 

\paragraph{Effect of scale.} For ImageNet fine-tuning, only the models with the largest data and samples seen were fine-tuned. Despite the narrower scale range, a similar relationship in the slope of the OpenAI CLIP vs OpenCLIP fit lines is observed across the model scales (Figure \ref{fig:ft-imagenet}). Moreover, scale consistently improves accuracy when fine-tuning on other downstream tasks (Figure~\ref{fig:ft-eight}).
While trends vary with the task, we find that the slope of the linear trend relating accuracy and total compute used for pre-training depends on the pre-training dataset, typically favors CLIP WIT-400M, as we observe in zero-shot experiments.

% TODO : Gabriel, Ross, Mitchell, Mehdi, Ludwig

%\subsubsection{Zero-shot transfer and robustness}
%\textbf{Evaluation setup.}
%\textbf{Results.}

%\subsection{Evaluation on different downstream tasks}
%\label{subsect:eval}

%\textbf{Zero-shot transfer.} 

%\textbf{Retrieval.}

%\textbf{Linear probing.}
%Including few-shot

%\textbf{Fine-tuning.}
%Including few-shot

%\subsection{Results.}
%\label{subsect:results}

%\textbf{Effect of scale on zero-shot transfer.}

%\textbf{Effect of scale on retrieval.}

%\textbf{Effect of scale on linear probing and fine-tuning.} Including few-shot.

\section{Discussion}
\label{sec:discussion_and_conclusion}

\paragraph{Larger scale improves performance across different downstream tasks.} In line with previous studies~\citep{Kaplan2020,tay2021scale,Riquelme2021,Zhai2022scaling}, our work observes scaling laws of power law form across various downstream tasks. We empirically find that scaling model, data and training samples seen results in consistent improvements on downstream zero-shot classification, retrieval, linear probing, and fine-tuning performance.

We also observe bottleneck behaviors \citep{Kaplan2020,Zhai2022scaling} that occur when fixing one scaling dimension while increasing others. For instance, OpenCLIP ViT-B/32 and ViT-B/16 are bottlenecked by the number of samples seen at the 13B scale. Increasing the number of samples seen to 34B reveals that LAION-2B brings clear improvement over LAION-400M, which would remain hidden when fixing the number of samples seen scale to a lower value. Similar observations may occur along other scaling dimensions. OpenCLIP ViT L/14 shows an example of data scale bottleneck on LAION-400M scale, as increasing the number of samples seen from 13B to 34B does not lead to improvements. The benefit of using a larger number of samples seen is then revealed when going to the larger LAION-2B dataset. 

Having derived scaling laws from our experimental observations, we are able to make predictions for both smaller and larger scales. Extrapolation has its limits, as saturation effects at both lower and higher scale ranges have been previously observed. We can however extrapolate to scales close to the ones we have already measured. A prediction for larger ViT-g/14 trained on LAION-2B with 34B samples delivers an estimate of 79.1\% ImageNet top-1 accuracy. This may appear at first sight modest compared to results reported by BASIC (85.7\% \citep{basic}), LiT (85.2\% \citep{zhai2021lit}) or CoCA (86.1\% \citep{yu2022coca}).
However, these works leverage an internal JFT dataset with labels which can be used for supervised pre-training. Moreover, for 973/1000 ImageNet classes, researchers were able to manually identify a correspondance from a JFT class~\cite{wortsman2022robust}. 
%However, one has to keep in mind that all those works made use of private data labeled with ImageNet class descriptions (JFT dataset), which makes it hard to make straightforward comparison in terms of zero-shot classification, as those models have been exposed to large amount of exact ImageNet labels during pre-training.
These works also use larger encoders, larger private data, and pre-train the encoders in multiple stages. Nonetheless, we estimate based on our empirical findings that further increasing model and data scale could result in competitive models even without using labeled data, additional supervised pre-training stages or additional losses.
Finally, we observe that the improvement of zero-shot ImageNet accuracy due to scaling up is accompanied by closely aligned improvements on robustness benchmarks.
%When scaling up, robustness thus increases accordingly and larger CLIP models are able to further improve on their beneficial robustness properties.   

%General statement - increasing scales hand in hand results in consistent improvement across broad range of downstream tasks. Mention robustness improvement. Note the bottleneck situations (like B/32  or L/14 with fixed samples seen budget of 13B);
% Larger scale improves generalization and transfer

% Effect of scale shows dependence on task type and pre-training dataset.

\paragraph{Scaling behavior depends on task type and pre-training dataset.} 
When measuring scaling coefficients for the observed power laws, we see that OpenAI CLIP and OpenCLIP have distinct scaling advantages over each other depending on the downstream task. OpenCLIP pre-trained on LAION-400M/2B data has stronger scaling trends for zero-shot retrieval, while OpenAI CLIP pre-trained on private WIT-400M data shows stronger scaling for zero-shot ImageNet classification.  We hypothesize that the observed differences are due to differences in the pre-training data, as we closely follow the architectures and pre-training recipes used for the OpenAI CLIP models. WIT-400M may have a stronger affinity to ImageNet as a result of the curation procedure, while LAION-400M/2B was filtered by a pre-trained OpenAI ViT-B/32 model relying on its similarity measurements for image-text pairs, which may have rendered the dataset more suitable for retrieval based tasks. This hypothesis can be tested by systematically varying dataset composition procedure (for example by using a stronger L/14 model for filtering crawled data) and observing the effect on scaling behavior across various task types.

% opens venues

% The hypothesis opens venues for further experiments, where dataset composition is systematically varied

% General statement: depending on task, we observe stronger scaling for one or another pre-training dataset, which indicates influence of pre-training dataset type on scaling behavior for specific tasks. 

% Effect of scale depends on task type and pre-training dataset

\paragraph{Limitations of the current study.} Observed scaling laws are based on points we were able to obtain with available compute resources. Therefore, the density of sampling the scales space is low. It is also not possible to conduct full hyper-parameter tuning, especially on larger scales, due to high compute costs. We rely thus on control experiments that look at few hyper-parameters at early pre-training stages and on tuning already performed in previous work to suggest that pre-training for each scale is not far from optimal. It was also not possible to obtain more points for OpenAI CLIP due to the private nature of the WIT-400M dataset. Moreover, we conduct only a simple duplication check for downstream data, which may leave few duplicates undetected. Previous studies \citep{radford2021learning, zhai2021lit} also reported that duplication in test sets do not significantly alter most results, potentially due to the very large scale and diversity of pre-training data.

% No systematic de-dup. No systematic hyperparam scan, single experiments for sanity checking. Points sampled are not dense due to limit. It is not possible to obtain more points from OpenAI CLIP for comparison as dataset and its composition routines are private.

% Batch size discussion?
\section{Conclusion}

We present a systematic study of scaling laws for contrastive language-image learning, investigating how scale affects performance on several downstream tasks and across adaptation methods. We find---in accord with previous works on uni-modal learning \citep{Kaplan2020,Zhai2022scaling}---a power law relation between scale (model, data and the number of samples seen) and downstream performance in a broad range of settings, including zero-shot classification, retrieval, few- and full-shot linear probing and fine-tuning. Interestingly, the scaling behavior for OpenCLIP-LAION pre-trained models and for OpenAI-WIT-400M pre-trained models differ, showing distinct benefits of one over another on different downstream tasks. We hypothesize that such task-specific scaling differences originate from the different pre-training datasets. Predictions for model performance on larger scales made on the basis of the scaling laws estimate 81.9\% zero-shot top-1 accuracy on ImageNet for a ViT-G/14 CLIP model trained on 68B image-text samples from scratch.

% Here we study purely contrastive pre-trained models, and it may be hypothesized based on current predictions that further increasing model and data scale to match scales of other works could also come close to the performance reported there, without using labeled data, additional supervised pre-training stages or additional losses.
% Predictions that we make on the grounds of derived scaling laws 

% Using the obtained power law form, we can make prediction for a performance of a well tuned pre-trained ViT-g/14 model when using largest data scale of 2B and largest samples seen budget of 34B, giving us error estimate of 20.8\%, thus predicting 79.2\% zero-shot top-1 ImageNet-1K accuracy for ViT-g/14. Predictions of stronger performance can be obtained for larger samples seen and model scales, for instance assuming 68B samples seen and getting estimates for ViT-H/14 (80\%), ViT-g/14 (81\%) and ViT-G/14 (82.4\%) (see also Appendix Sec. \ref{sec:appendix:predictions}).

% benefiting one over another  
% one being stronger than another on distinct tasks depending on 
Our study opens many directions for further investigations. Obtaining more data points for smaller and intermediate scales can provide enough sampling density to better understand the optimal configuration of model size, dataset size and number of samples seen given a fixed compute, similar to works such as~\cite{hoffmann2022training, koppula2022should}. Scaling laws for robustness benchmarks \citep{Taori2020} can be derived when controlling for larger accuracies observed at larger scales. Further, treating vision and text encoder scales separately may lead to modality specific scaling laws. A promising direction is to study the effect of the pre-training dataset on scaling behavior. Our observations so far hint that the data source may strongly influence task-specific scaling. This paves the road for studies on foundation datasets \citep{sorscher2022beyond}. % which composition aims towards strong scaling across broad range of downstream tasks. 
Having open datasets \citep{schuhmann2021laion,Schuhmann2022} and open source tools \citep{ilharco_gabriel_2021_5143773} at hand, such experiments can be conducted and reproduced in a common effort by the broader research community.   

% towards achieving strong scaling 

% Further, studying behavior of effective robustness can provide
% vision text encoder separately
% Important property of effective robustness can be studied by deriving scaling laws. 
% Power laws across various tasks. Robustness goes along. Predictions see larger scales approaching SOTA. Dataset dependency of scaling behavior for specific tasks. Outlook: Scaling laws for Effective out-of-distribution robustness; More dense point sample on lower-mid-scales to get better prediction for larger scales and optimal model selection given fixed compute. Dataset composition and its effect on scaling behavior. Foundation datasets to obtain models with strong scaling on a broad range of downstream tasks. Democratizing scaling law studies and studies of larger models with strong performance and emergent functions via open source data and models.

% WARNING! Acknowledgments sections to be REMOVED for the anonymous submission !
\section*{Acknowledgments.}
We would like to express gratitude to all the people who are working on making code, models and data publicly available, advancing community based research and making research more reproducible. Specifically, we would like to thank all the members of the LAION discord server\footnote{\url{https://discord.gg/BZqhreFazY}} community that was pivotal for the effort to compose LAION-400m and LAION-5B datasets without which this study would be impossible, and openAI for making their pre-trained CLIP models publicly available. We want to thank Hugging Face for providing hosting space for open datasets and models and Stability AI for providing supercomputing resources and storage space.

The authors gratefully acknowledge the Gauss Centre for Supercomputing e.V. \footnote{\url{https://gauss-centre.eu}} for funding this work by providing computing time through the John von Neumann Institute for Computing (NIC) on the GCS Supercomputer JUWELS Booster \citep{JUWELSBooster2020} at Jülich Supercomputing Centre (JSC). We also acknowledge storage resources on JUST \citep{graf2021just} granted and operated by JSC, as well as computing resources from the Helmholtz Data Federation (HDF). Further thank goes for support provided by JSC supercomputing facility administration team, especially to Damian Alvarez for his endurance and patience during the long "de-micing" sessions on JUWELS Booster.

MC and JJ acknowledge partial funding by the Federal Ministry of Education and Research of Germany under BMBF grant no. 01IS22094B WestAI - AI Service Center West.

Special thanks goes also to Richard Vencu (LAION, Stability AI) for his on-going dedication towards enabling a HPC system and infrastructure around it that can be used by broad community of researchers and citizen scientists. 

{\small
\bibliographystyle{ieee_fullname}
\bibliography{bibliography}

\begin{thebibliography}{10}\itemsep=-1pt

\bibitem{common_crawl}
Common {C}rawl.
\newblock \url{https://commoncrawl.org}.

\bibitem{stability_hpc}
Stability {AI} {HPC} facility, https://hpc.stability.ai.

\bibitem{alayrac2022flamingo}
Jean-Baptiste Alayrac, Jeff Donahue, Pauline Luc, Antoine Miech, Iain Barr,
  Yana Hasson, Karel Lenc, Arthur Mensch, Katie Millican, Malcolm Reynolds,
  et~al.
\newblock Flamingo: a visual language model for few-shot learning.
\newblock {\em arXiv preprint arXiv:2204.14198}, 2022.

\bibitem{bao2021beit}
Hangbo Bao, Li Dong, and Furu Wei.
\newblock Beit: Bert pre-training of image transformers.
\newblock {\em arXiv preprint arXiv:2106.08254}, 2021.

\bibitem{objectnet}
Andrei Barbu, David Mayo, Julian Alverio, William Luo, Christopher Wang, Dan
  Gutfreund, Josh Tenenbaum, and Boris Katz.
\newblock Objectnet: A large-scale bias-controlled dataset for pushing the
  limits of object recognition models.
\newblock In {\em Advances in Neural Information Processing Systems (NeurIPS)},
  2019.

\bibitem{Belkin2019}
Mikhail Belkin, Daniel Hsu, Siyuan Ma, and Soumik Mandal.
\newblock Reconciling modern machine-learning practice and the classical
  bias-variance trade-off.
\newblock {\em Proceedings of the National Academy of Sciences of the United
  States of America}, 116:15849--15854, Aug. 2019.

\bibitem{beyer2020we}
Lucas Beyer, Olivier~J H{\'e}naff, Alexander Kolesnikov, Xiaohua Zhai, and
  A{\"a}ron van~den Oord.
\newblock Are we done with imagenet?
\newblock {\em arXiv preprint arXiv:2006.07159}, 2020.

\bibitem{Brown2020}
Tom Brown, Benjamin Mann, Nick Ryder, Melanie Subbiah, Jared~D Kaplan, Prafulla
  Dhariwal, Arvind Neelakantan, Pranav Shyam, Girish Sastry, Amanda Askell,
  Sandhini Agarwal, Ariel Herbert-Voss, Gretchen Krueger, Tom Henighan, Rewon
  Child, Aditya Ramesh, Daniel Ziegler, Jeffrey Wu, Clemens Winter, Chris
  Hesse, Mark Chen, Eric Sigler, Mateusz Litwin, Scott Gray, Benjamin Chess,
  Jack Clark, Christopher Berner, Sam McCandlish, Alec Radford, Ilya Sutskever,
  and Dario Amodei.
\newblock Language models are few-shot learners.
\newblock In H. Larochelle, M. Ranzato, R. Hadsell, M.~F. Balcan, and H. Lin,
  editors, {\em Advances in Neural Information Processing Systems}, volume~33,
  pages 1877--1901. Curran Associates, Inc., 2020.

\bibitem{Bubeck2021}
Sebastien Bubeck and Mark Sellke.
\newblock A universal law of robustness via isoperimetry.
\newblock In M. Ranzato, A. Beygelzimer, Y. Dauphin, P.S. Liang, and J.~Wortman
  Vaughan, editors, {\em Advances in Neural Information Processing Systems},
  volume~34, pages 28811--28822. Curran Associates, Inc., 2021.

\bibitem{kakaobrain2022coyo-700m}
Minwoo Byeon, Beomhee Park, Haecheon Kim, Sungjun Lee, Woonhyuk Baek, and
  Saehoon Kim.
\newblock Coyo-700m: Image-text pair dataset.
\newblock \url{https://github.com/kakaobrain/coyo-dataset}, 2022.

\bibitem{chen2021lottery}
Tianlong Chen, Jonathan Frankle, Shiyu Chang, Sijia Liu, Yang Zhang, Michael
  Carbin, and Zhangyang Wang.
\newblock The lottery tickets hypothesis for supervised and self-supervised
  pre-training in computer vision models.
\newblock In {\em Proceedings of the IEEE/CVF Conference on Computer Vision and
  Pattern Recognition}, pages 16306--16316, 2021.

\bibitem{chen2022pali}
Xi Chen, Xiao Wang, Soravit Changpinyo, AJ Piergiovanni, Piotr Padlewski,
  Daniel Salz, Sebastian Goodman, Adam Grycner, Basil Mustafa, Lucas Beyer,
  et~al.
\newblock Pali: A jointly-scaled multilingual language-image model.
\newblock {\em arXiv preprint arXiv:2209.06794}, 2022.

\bibitem{cheng2017remote}
Gong Cheng, Junwei Han, and Xiaoqiang Lu.
\newblock Remote sensing image scene classification: Benchmark and state of the
  art.
\newblock {\em Proceedings of the Institute of Electrical and Electronics
  Engineers (IEEE)}, 2017.
\newblock \url{https://ieeexplore.ieee.org/abstract/document/7891544}.

\bibitem{dtd}
Mircea Cimpoi, Subhransu Maji, Iasonas Kokkinos, Sammy Mohamed, and Andrea
  Vedaldi.
\newblock Describing textures in the wild.
\newblock In {\em Conference on Computer Vision and Pattern Recognition
  (CVPR)}, 2014.
\newblock
  \url{https://openaccess.thecvf.com/content_cvpr_2014/html/Cimpoi_Describing_Textures_in_2014_CVPR_paper.html}.

\bibitem{Deng2009a}
J. {Deng}, W. {Dong}, R. {Socher}, L. {Li}, {Kai Li}, and {Li Fei-Fei}.
\newblock Imagenet: A large-scale hierarchical image database.
\newblock In {\em Proc. IEEE Conf. Computer Vision and Pattern Recognition},
  pages 248--255, June 2009.

\bibitem{dettmers2022llm}
Tim Dettmers, Mike Lewis, Younes Belkada, and Luke Zettlemoyer.
\newblock Llm. int8 (): 8-bit matrix multiplication for transformers at scale.
\newblock {\em arXiv preprint arXiv:2208.07339}, 2022.

\bibitem{devlin2019bert}
Jacob Devlin, Ming-Wei Chang, Kenton Lee, and Kristina Toutanova.
\newblock {BERT}: Pre-training of deep bidirectional transformers for language
  understanding.
\newblock In {\em Proceedings of the 2019 Conference of the North {A}merican
  Chapter of the Association for Computational Linguistics: Human Language
  Technologies, Volume 1 (Long and Short Papers)}, pages 4171--4186,
  Minneapolis, Minnesota, June 2019. Association for Computational Linguistics.

\bibitem{fang2022data}
Alex Fang, Gabriel Ilharco, Mitchell Wortsman, Yuhao Wan, Vaishaal Shankar,
  Achal Dave, and Ludwig Schmidt.
\newblock Data determines distributional robustness in contrastive language
  image pre-training (clip).
\newblock {\em arXiv preprint arXiv:2205.01397}, 2022.

\bibitem{fang2022eva}
Yuxin Fang, Wen Wang, Binhui Xie, Quan Sun, Ledell Wu, Xinggang Wang, Tiejun
  Huang, Xinlong Wang, and Yue Cao.
\newblock Eva: Exploring the limits of masked visual representation learning at
  scale.
\newblock {\em arXiv preprint arXiv:2211.07636}, 2022.

\bibitem{graf2021just}
Stephan Graf and Olaf Mextorf.
\newblock Just: Large-scale multi-tier storage infrastructure at the j{\"u}lich
  supercomputing centre.
\newblock {\em Journal of large-scale research facilities JLSRF}, 7:180, 2021.

\bibitem{eurosat}
Patrick Helber, Benjamin Bischke, Andreas Dengel, and Damian Borth.
\newblock Eurosat: A novel dataset and deep learning benchmark for land use and
  land cover classification.
\newblock {\em Journal of Selected Topics in Applied Earth Observations and
  Remote Sensing}, 2019.
\newblock \url{https://arxiv.org/abs/1709.00029}.

\bibitem{imagenetr}
Dan Hendrycks, Steven Basart, Norman Mu, Saurav Kadavath, Frank Wang, Evan
  Dorundo, Rahul Desai, Tyler Zhu, Samyak Parajuli, Mike Guo, Dawn Song, Jacob
  Steinhardt, and Justin Gilmer.
\newblock The many faces of robustness: A critical analysis of
  out-of-distribution generalization.
\newblock {\em International Conference on Computer Vision (ICCV)}, 2021.
\newblock \url{https://arxiv.org/abs/2006.16241}.

\bibitem{imagenetc}
Dan Hendrycks and Thomas Dietterich.
\newblock Benchmarking neural network robustness to common corruptions and
  perturbations.
\newblock {\em International Conference on Learning Representations (ICLR)},
  2019.
\newblock \url{https://arxiv.org/abs/1903.12261}.

\bibitem{imageneta}
Dan Hendrycks, Kevin Zhao, Steven Basart, Jacob Steinhardt, and Dawn Song.
\newblock Natural adversarial examples.
\newblock {\em Conference on Computer Vision and Pattern Recognition (CVPR)},
  2021.
\newblock \url{https://arxiv.org/abs/1907.07174}.

\bibitem{Henighan2020}
Tom Henighan, Jared Kaplan, Mor Katz, Mark Chen, Christopher Hesse, Jacob
  Jackson, Heewoo Jun, Tom~B Brown, Prafulla Dhariwal, Scott Gray, Chris
  Hallacy, Benjamin Mann, Alec Radford, Aditya Ramesh, Nick Ryder, Daniel~M.
  Ziegler, John Schulman, Dario Amodei, and Sam McCandlish.
\newblock Scaling laws for autoregressive generative modeling.
\newblock {\em arXiv preprint arXiv:2010.14701}, 2020.

\bibitem{hestness2017deep}
Joel Hestness, Sharan Narang, Newsha Ardalani, Gregory Diamos, Heewoo Jun,
  Hassan Kianinejad, Md Patwary, Mostofa Ali, Yang Yang, and Yanqi Zhou.
\newblock Deep learning scaling is predictable, empirically.
\newblock {\em arXiv preprint arXiv:1712.00409}, 2017.

\bibitem{ho2022imagen}
Jonathan Ho, William Chan, Chitwan Saharia, Jay Whang, Ruiqi Gao, Alexey
  Gritsenko, Diederik~P Kingma, Ben Poole, Mohammad Norouzi, David~J Fleet,
  et~al.
\newblock Imagen video: High definition video generation with diffusion models.
\newblock {\em arXiv preprint arXiv:2210.02303}, 2022.

\bibitem{hoffmann2022training}
Jordan Hoffmann, Sebastian Borgeaud, Arthur Mensch, Elena Buchatskaya, Trevor
  Cai, Eliza Rutherford, Diego de~Las Casas, Lisa~Anne Hendricks, Johannes
  Welbl, Aidan Clark, et~al.
\newblock Training compute-optimal large language models.
\newblock {\em arXiv preprint arXiv:2203.15556}, 2022.

\bibitem{howard2018universal}
Jeremy Howard and Sebastian Ruder.
\newblock Universal language model fine-tuning for text classification.
\newblock In {\em Proceedings of the 56th Annual Meeting of the Association for
  Computational Linguistics (Volume 1: Long Papers)}, pages 328--339,
  Melbourne, Australia, July 2018. Association for Computational Linguistics.

\bibitem{hu2021scaling}
Xiaowei Hu, Zhe Gan, Jianfeng Wang, Zhengyuan Yang, Zicheng Liu, Yumao Lu, and
  Lijuan Wang.
\newblock Scaling up vision-language pre-training for image captioning.
\newblock In {\em Proceedings of the IEEE/CVF Conference on Computer Vision and
  Pattern Recognition (CVPR)}, pages 17980--17989, June 2022.

\bibitem{ilharco2022patching}
Gabriel Ilharco, Mitchell Wortsman, Samir~Yitzhak Gadre, Shuran Song, Hannaneh
  Hajishirzi, Simon Kornblith, Ali Farhadi, and Ludwig Schmidt.
\newblock Patching open-vocabulary models by interpolating weights.
\newblock {\em arXiv preprint arXiv:2208.05592}, 2022.

\bibitem{ilharco_gabriel_2021_5143773}
Gabriel Ilharco, Mitchell Wortsman, Ross Wightman, Cade Gordon, Nicholas
  Carlini, Rohan Taori, Achal Dave, Vaishaal Shankar, Hongseok Namkoong, John
  Miller, Hannaneh Hajishirzi, Ali Farhadi, and Ludwig Schmidt.
\newblock Openclip, July 2021.

\bibitem{ALIGN}
Chao Jia, Yinfei Yang, Ye Xia, Yi-Ting Chen, Zarana Parekh, Hieu Pham, Quoc Le,
  Yun-Hsuan Sung, Zhen Li, and Tom Duerig.
\newblock Scaling up visual and vision-language representation learning with
  noisy text supervision.
\newblock In {\em International Conference on Machine Learning}, pages
  4904--4916. PMLR, 2021.

\bibitem{JUWELSBooster2020}
{Juelich Supercomputing Center}.
\newblock {JUWELS} {B}ooster {S}upercomputer, 2020.
\newblock
  \url{https://apps.fz-juelich.de/jsc/hps/juwels/configuration.html\#hardware-configuration-of-the-system-name-booster-module}.

\bibitem{Kaplan2020}
Jared Kaplan, Sam McCandlish, Tom Henighan, Tom~B Brown, Benjamin Chess, Rewon
  Child, Scott Gray, Alec Radford, Jeffrey Wu, and Dario Amodei.
\newblock Scaling laws for neural language models.
\newblock {\em arXiv preprint arXiv:2001.08361}, 2020.

\bibitem{karpathy2015deep}
Andrej Karpathy and Li Fei-Fei.
\newblock Deep visual-semantic alignments for generating image descriptions.
\newblock In {\em Proceedings of the IEEE conference on computer vision and
  pattern recognition}, pages 3128--3137, 2015.

\bibitem{kingma2014adam}
Diederik~P Kingma and Jimmy Ba.
\newblock Adam: A method for stochastic optimization.
\newblock {\em arXiv preprint arXiv:1412.6980}, 2014.

\bibitem{Kolesnikov2020}
Alexander Kolesnikov, Lucas Beyer, Xiaohua Zhai, Joan Puigcerver, Jessica Yung,
  Sylvain Gelly, and Neil Houlsby.
\newblock Big transfer (bit): General visual representation learning.
\newblock In Andrea Vedaldi, Horst Bischof, Thomas Brox, and Jan-Michael Frahm,
  editors, {\em Computer Vision -- ECCV 2020}, pages 491--507, Cham, 2020.
  Springer International Publishing.

\bibitem{koppula2022should}
Skanda Koppula, Yazhe Li, Evan Shelhamer, Andrew Jaegle, Nikhil Parthasarathy,
  Relja Arandjelovic, Jo{\~a}o Carreira, and Olivier H{\'e}naff.
\newblock Where should i spend my flops? efficiency evaluations of visual
  pre-training methods.
\newblock {\em arXiv preprint arXiv:2209.15589}, 2022.

\bibitem{kornblith2019better}
Simon Kornblith, Jonathon Shlens, and Quoc~V Le.
\newblock Do better imagenet models transfer better?
\newblock In {\em Conference on Computer Vision and Pattern Recognition
  (CVPR)}, 2019.
\newblock \url{https://arxiv.org/abs/1805.08974}.

\bibitem{cars}
Jonathan Krause, Michael Stark, Jia Deng, and Li Fei-Fei.
\newblock 3d object representations for fine-grained categorization.
\newblock In {\em International Conference on Computer Vision Workshops
  (ICML)}, 2013.
\newblock
  \url{https://www.cv-foundation.org/openaccess/content_iccv_workshops_2013/W19/html/Krause_3D_Object_Representations_2013_ICCV_paper.html}.

\bibitem{krishna2017visual}
Ranjay Krishna, Yuke Zhu, Oliver Groth, Justin Johnson, Kenji Hata, Joshua
  Kravitz, Stephanie Chen, Yannis Kalantidis, Li-Jia Li, David~A Shamma, et~al.
\newblock Visual genome: Connecting language and vision using crowdsourced
  dense image annotations.
\newblock {\em International journal of computer vision}, 123(1):32--73, 2017.

\bibitem{Krizhevsky2009}
Alex Krizhevsky.
\newblock Learning multiple layers of features from tiny images.
\newblock Technical report, 2009.

\bibitem{lecun1998mnist}
Yann LeCun.
\newblock The mnist database of handwritten digits, 1998.
\newblock \url{http://yann.lecun.com/exdb/mnist/}.

\bibitem{li2020pytorch}
Shen Li, Yanli Zhao, Rohan Varma, Omkar Salpekar, Pieter Noordhuis, Teng Li,
  Adam Paszke, Jeff Smith, Brian Vaughan, Pritam Damania, et~al.
\newblock Pytorch distributed: Experiences on accelerating data parallel
  training.
\newblock {\em arXiv preprint arXiv:2006.15704}, 2020.

\bibitem{lin2014microsoft}
Tsung-Yi Lin, Michael Maire, Serge Belongie, James Hays, Pietro Perona, Deva
  Ramanan, Piotr Doll{\'a}r, and C~Lawrence Zitnick.
\newblock Microsoft coco: Common objects in context.
\newblock In {\em European conference on computer vision}, pages 740--755.
  Springer, 2014.

\bibitem{loshchilov2018decoupled}
Ilya Loshchilov and Frank Hutter.
\newblock Decoupled weight decay regularization.
\newblock In {\em International Conference on Learning Representations}, 2019.

\bibitem{miller2021accuracy}
John~P Miller, Rohan Taori, Aditi Raghunathan, Shiori Sagawa, Pang~Wei Koh,
  Vaishaal Shankar, Percy Liang, Yair Carmon, and Ludwig Schmidt.
\newblock Accuracy on the line: on the strong correlation between
  out-of-distribution and in-distribution generalization.
\newblock In {\em International Conference on Machine Learning}, pages
  7721--7735. PMLR, 2021.

\bibitem{mustafa2022multimodal}
Basil Mustafa, Carlos Riquelme, Joan Puigcerver, Rodolphe Jenatton, and Neil
  Houlsby.
\newblock Multimodal contrastive learning with limoe: the language-image
  mixture of experts.
\newblock {\em arXiv preprint arXiv:2206.02770}, 2022.

\bibitem{svhn}
Yuval Netzer, Tao Wang, Adam Coates, Alessandro Bissacco, Bo Wu, and Andrew~Y
  Ng.
\newblock Reading digits in natural images with unsupervised feature learning.
\newblock In {\em Advances in Neural Information Processing Systems (NeurIPS)
  Workshops}, 2011.
\newblock
  \url{https://storage.googleapis.com/pub-tools-public-publication-data/pdf/37648.pdf}.

\bibitem{nguyen2022quality}
Thao Nguyen, Gabriel Ilharco, Mitchell Wortsman, Sewoong Oh, and Ludwig
  Schmidt.
\newblock Quality not quantity: On the interaction between dataset design and
  robustness of clip.
\newblock {\em arXiv preprint arXiv:2208.05516}, 2022.

\bibitem{oord2018representation}
Aaron van~den Oord, Yazhe Li, and Oriol Vinyals.
\newblock Representation learning with contrastive predictive coding.
\newblock {\em arXiv preprint arXiv:1807.03748}, 2018.

\bibitem{paszke2019pytorch}
Adam Paszke, Sam Gross, Francisco Massa, Adam Lerer, James Bradbury, Gregory
  Chanan, Trevor Killeen, Zeming Lin, Natalia Gimelshein, Luca Antiga, et~al.
\newblock Pytorch: An imperative style, high-performance deep learning library.
\newblock {\em Advances in neural information processing systems}, 32, 2019.

\bibitem{basic}
Hieu Pham, Zihang Dai, Golnaz Ghiasi, Hanxiao Liu, Adams~Wei Yu, Minh-Thang
  Luong, Mingxing Tan, and Quoc~V Le.
\newblock Combined scaling for zero-shot transfer learning.
\newblock {\em arXiv preprint arXiv:2111.10050}, 2021.

\bibitem{radford2021learning}
Alec Radford, Jong~Wook Kim, Chris Hallacy, Aditya Ramesh, Gabriel Goh,
  Sandhini Agarwal, Girish Sastry, Amanda Askell, Pamela Mishkin, Jack Clark,
  et~al.
\newblock Learning transferable visual models from natural language
  supervision.
\newblock In {\em International Conference on Machine Learning}, pages
  8748--8763. PMLR, 2021.

\bibitem{radford2022robust}
Alec Radford, Jong~Wook Kim, Tao Xu, Greg Brockman, Christine McLeavey, and
  Ilya Sutskever.
\newblock Robust speech recognition via large-scale weak supervision.
\newblock Technical report, Tech. Rep., Technical report, OpenAI, 2022.

\bibitem{t5}
Colin Raffel, Noam Shazeer, Adam Roberts, Katherine Lee, Sharan Narang, Michael
  Matena, Yanqi Zhou, Wei Li, and Peter~J. Liu.
\newblock Exploring the limits of transfer learning with a unified text-to-text
  transformer.
\newblock {\em Journal of Machine Learning Research}, 21(140):1--67, 2020.

\bibitem{ramasesh2021effect}
Vinay~Venkatesh Ramasesh, Aitor Lewkowycz, and Ethan Dyer.
\newblock Effect of scale on catastrophic forgetting in neural networks.
\newblock In {\em International Conference on Learning Representations}, 2021.

\bibitem{ramesh2022hierarchical}
Aditya Ramesh, Prafulla Dhariwal, Alex Nichol, Casey Chu, and Mark Chen.
\newblock Hierarchical text-conditional image generation with clip latents.
\newblock {\em arXiv preprint arXiv:2204.06125}, 2022.

\bibitem{pmlr-v97-recht19a}
Benjamin Recht, Rebecca Roelofs, Ludwig Schmidt, and Vaishaal Shankar.
\newblock Do {I}mage{N}et classifiers generalize to {I}mage{N}et?
\newblock In {\em International Conference on Machine Learning (ICML)}, 2019.
\newblock \url{https://arxiv.org/abs/1902.10811}.

\bibitem{Riquelme2021}
Carlos Riquelme, Joan Puigcerver, Basil Mustafa, Maxim Neumann, Rodolphe
  Jenatton, Andr{\'e} Susano~Pinto, Daniel Keysers, and Neil Houlsby.
\newblock Scaling vision with sparse mixture of experts.
\newblock {\em Advances in Neural Information Processing Systems}, 34, 2021.

\bibitem{rombach2021highresolution}
Robin Rombach, Andreas Blattmann, Dominik Lorenz, Patrick Esser, and Bj{\"o}rn
  Ommer.
\newblock High-resolution image synthesis with latent diffusion models.
\newblock In {\em Proceedings of the IEEE/CVF Conference on Computer Vision and
  Pattern Recognition}, pages 10684--10695, 2022.

\bibitem{pmlr-v139-rosenfeld21a}
Jonathan~S Rosenfeld, Jonathan Frankle, Michael Carbin, and Nir Shavit.
\newblock On the predictability of pruning across scales.
\newblock In Marina Meila and Tong Zhang, editors, {\em Proceedings of the 38th
  International Conference on Machine Learning}, volume 139 of {\em Proceedings
  of Machine Learning Research}, pages 9075--9083. PMLR, 18--24 Jul 2021.

\bibitem{Schuhmann2022}
Christoph Schuhmann, Romain Beaumont, Richard Vencu, Cade~W Gordon, Ross
  Wightman, Mehdi Cherti, Theo Coombes, Aarush Katta, Clayton Mullis, Mitchell
  Wortsman, Patrick Schramowski, Srivatsa~R Kundurthy, Katherine Crowson,
  Ludwig Schmidt, Robert Kaczmarczyk, and Jenia Jitsev.
\newblock {LAION}-5{B}: An open large-scale dataset for training next
  generation image-text models.
\newblock In {\em Thirty-sixth Conference on Neural Information Processing
  Systems (NeurIPS), Datasets and Benchmarks Track}, 2022.

\bibitem{schuhmann2021laion}
Christoph Schuhmann, Richard Vencu, Romain Beaumont, Robert Kaczmarczyk,
  Clayton Mullis, Aarush Katta, Theo Coombes, Jenia Jitsev, and Aran
  Komatsuzaki.
\newblock {LAION}-400{M}: Open dataset of {CLIP}-filtered 400 million
  image-text pairs.
\newblock {\em arXiv preprint arXiv:2111.02114}, 2021.

\bibitem{sharma-etal-2018-conceptual}
Piyush Sharma, Nan Ding, Sebastian Goodman, and Radu Soricut.
\newblock Conceptual captions: A cleaned, hypernymed, image alt-text dataset
  for automatic image captioning.
\newblock In {\em Proceedings of the 56th Annual Meeting of the Association for
  Computational Linguistics (Volume 1: Long Papers)}, pages 2556--2565,
  Melbourne, Australia, July 2018. Association for Computational Linguistics.

\bibitem{sorscher2022beyond}
Ben Sorscher, Robert Geirhos, Shashank Shekhar, Surya Ganguli, and Ari~S
  Morcos.
\newblock Beyond neural scaling laws: beating power law scaling via data
  pruning.
\newblock {\em arXiv preprint arXiv:2206.14486}, 2022.

\bibitem{gtsrb}
Johannes Stallkamp, Marc Schlipsing, Jan Salmen, and Christian Igel.
\newblock The german traffic sign recognition benchmark: a multi-class
  classification competition.
\newblock In {\em International Joint Conference on Neural Networks (IJCNN)},
  2011.
\newblock \url{https://ieeexplore.ieee.org/document/6033395}.

\bibitem{Sun2017}
Chen Sun, Abhinav Shrivastava, Saurabh Singh, and Abhinav Gupta.
\newblock Revisiting unreasonable effectiveness of data in deep learning era.
\newblock In {\em Proceedings of the IEEE international conference on computer
  vision}, pages 843--852, 2017.

\bibitem{Taori2020}
Rohan Taori, Achal Dave, Vaishaal Shankar, Nicholas Carlini, Benjamin Recht,
  and Ludwig Schmidt.
\newblock Measuring robustness to natural distribution shifts in image
  classification.
\newblock {\em Advances in Neural Information Processing Systems},
  33:18583--18599, 2020.

\bibitem{tay2022scaling}
Yi Tay, Mostafa Dehghani, Samira Abnar, Hyung~Won Chung, William Fedus, Jinfeng
  Rao, Sharan Narang, Vinh~Q Tran, Dani Yogatama, and Donald Metzler.
\newblock Scaling laws vs model architectures: How does inductive bias
  influence scaling?
\newblock {\em arXiv preprint arXiv:2207.10551}, 2022.

\bibitem{tay2021scale}
Yi Tay, Mostafa Dehghani, Jinfeng Rao, William Fedus, Samira Abnar, Hyung~Won
  Chung, Sharan Narang, Dani Yogatama, Ashish Vaswani, and Donald Metzler.
\newblock Scale efficiently: Insights from pretraining and finetuning
  transformers.
\newblock In {\em International Conference on Learning Representations}, 2021.

\bibitem{thomee2016yfcc100m}
Bart Thomee, David~A Shamma, Gerald Friedland, Benjamin Elizalde, Karl Ni,
  Douglas Poland, Damian Borth, and Li-Jia Li.
\newblock Yfcc100m: The new data in multimedia research.
\newblock {\em Communications of the ACM}, 59(2):64--73, 2016.

\bibitem{imagenetsketch}
Haohan Wang, Songwei Ge, Zachary Lipton, and Eric~P Xing.
\newblock Learning robust global representations by penalizing local predictive
  power.
\newblock In {\em Advances in Neural Information Processing Systems (NeurIPS)},
  2019.
\newblock \url{https://arxiv.org/abs/1905.13549}.

\bibitem{wang2022internimage}
Wenhai Wang, Jifeng Dai, Zhe Chen, Zhenhang Huang, Zhiqi Li, Xizhou Zhu,
  Xiaowei Hu, Tong Lu, Lewei Lu, Hongsheng Li, et~al.
\newblock Internimage: Exploring large-scale vision foundation models with
  deformable convolutions.
\newblock {\em arXiv preprint arXiv:2211.05778}, 2022.

\bibitem{timm}
Ross Wightman.
\newblock Pytorch image models.
\newblock \url{https://github.com/rwightman/pytorch-image-models}, 2019.

\bibitem{wortsman2022robust}
Mitchell Wortsman, Gabriel Ilharco, Jong~Wook Kim, Mike Li, Simon Kornblith,
  Rebecca Roelofs, Raphael~Gontijo Lopes, Hannaneh Hajishirzi, Ali Farhadi,
  Hongseok Namkoong, et~al.
\newblock Robust fine-tuning of zero-shot models.
\newblock In {\em Proceedings of the IEEE/CVF Conference on Computer Vision and
  Pattern Recognition}, pages 7959--7971, 2022.

\bibitem{sun397}
Jianxiong Xiao, Krista~A Ehinger, James Hays, Antonio Torralba, and Aude Oliva.
\newblock Sun database: Exploring a large collection of scene categories.
\newblock {\em International Journal of Computer Vision (IJCV)}, 2016.
\newblock \url{https://link.springer.com/article/10.1007/s11263-014-0748-y}.

\bibitem{flickr30_young2014image}
Peter Young, Alice Lai, Micah Hodosh, and Julia Hockenmaier.
\newblock From image descriptions to visual denotations: New similarity metrics
  for semantic inference over event descriptions.
\newblock {\em Transactions of the Association for Computational Linguistics},
  2:67--78, 2014.

\bibitem{yu2022coca}
Jiahui Yu, Zirui Wang, Vijay Vasudevan, Legg Yeung, Mojtaba Seyedhosseini, and
  Yonghui Wu.
\newblock {C}o{C}a: Contrastive captioners are image-text foundation models.
\newblock {\em Transactions on Machine Learning Research}, 2022.

\bibitem{zauner2010implementation}
Christoph Zauner.
\newblock Implementation and benchmarking of perceptual image hash functions.
\newblock 2010.

\bibitem{Zhai2022scaling}
Xiaohua Zhai, Alexander Kolesnikov, Neil Houlsby, and Lucas Beyer.
\newblock Scaling vision transformers.
\newblock In {\em Proceedings of the IEEE/CVF Conference on Computer Vision and
  Pattern Recognition}, pages 12104--12113, 2022.

\bibitem{zhai2019large}
Xiaohua Zhai, Joan Puigcerver, Alexander Kolesnikov, Pierre Ruyssen, Carlos
  Riquelme, Mario Lucic, Josip Djolonga, Andre~Susano Pinto, Maxim Neumann,
  Alexey Dosovitskiy, et~al.
\newblock A large-scale study of representation learning with the visual task
  adaptation benchmark.
\newblock {\em arXiv preprint arXiv:1910.04867}, 2019.

\bibitem{zhai2021lit}
Xiaohua Zhai, Xiao Wang, Basil Mustafa, Andreas Steiner, Daniel Keysers,
  Alexander Kolesnikov, and Lucas Beyer.
\newblock Li{T}: Zero-shot transfer with locked-image text tuning.
\newblock In {\em Proceedings of the IEEE/CVF Conference on Computer Vision and
  Pattern Recognition}, pages 18123--18133, 2022.

\end{thebibliography}
}

% WARNING! Appendix sections to be REMOVED for the anonymous submission !
% (Supplementary has to be submitted as separate pdf)
\clearpage
% ------------------------ Appendix ------------------------ 

\begin{appendix}

\begin{center}
{\Large\bf Supplementary: Reproducible scaling laws for contrastive language-image learning}
\end{center}

\section{Further details on distributed training}
\label{appendix:distributed_training}

% Details in distributed training - amount of GPUs, time spent, scaling plots

%\subsection{JUWELS Booster Supercomputer}
\subsection{Supercomputer specifications}
\label{appendix:booster}

The JUWELS Booster~\citep{JUWELSBooster2020} supercomputer used for training  consists of \num{936} compute nodes that host four NVIDIA A100 GPUs each, providing \num{3744}~GPUs in total. The installed A100 Tensor Core GPUs~(\SI{40}{\giga\byte}) provide \SI{19.5}{\tera\floppersec} of $\text{FP64}_\text{TC}$ computing performance each. The GPUs are hosted by AMD EPYC 7402 CPUs with $2\times 24$~cores (SMT-2) per node, clocked with \SI{2.8}{\giga\Hz}. Each node is diskless and is equipped with \SI{512}{\giga\byte} of RAM. The network is based on Mellanox HDR200 InfiniBand, with four Mellanox ConnectX~6 devices per node, each providing \SI{200}{\giga\bit\per\second} bandwidth per direction.

% remove for the submission ~\citep{JUWELSBooster2020}, re-introduce in final version

% Installed in November 2020, JUWELS Booster~\citep{JUWELSBooster2020} features \num{936} compute nodes that host four NVIDIA A100 GPUs each, providing \num{3744}~GPUs in total. The installed A100 Tensor Core GPUs~(\SI{40}{\giga\byte}) provide \SI{19.5}{\tera\floppersec} of $\text{FP64}_\text{TC}$ computing performance each. The GPUs are hosted by AMD EPYC 7402 CPUs with $2\times 24$~cores (SMT-2) per node, clocked with \SI{2.8}{\giga\Hz}. Each node is diskless and is equipped with \SI{512}{\giga\byte} of RAM. The network of JUWELS Booster is based on Mellanox HDR200 InfiniBand, with four Mellanox ConnectX~6 devices per node, each providing \SI{200}{\giga\bit\per\second} bandwidth per direction.

The NVIDIA A100 GPUs reach peak efficiency of \SI{48,75}{\giga\floppersec\per\watt} when utilizing the $\text{FP64}$ Tensor Cores. This made the employed machine rank highest in the Green500 list as of November 2020 as the most energy efficient supercomputer among the first 100 machines of the Top500 list with \SI{25}{\giga\floppersec\per\watt}.

% The NVIDIA A100 GPUs installed into JUWELS Booster reach peak efficiency of \SI{48,75}{\giga\floppersec\per\watt} when utilizing the $\text{FP64}$ Tensor Cores. This made JUWELS Booster rank highest in the Green500 list as of November 2020 as the most energy efficient supercomputer among the first 100 machines of the Top500 list with \SI{25}{\giga\floppersec\per\watt}.

%\subsection{JUWELS Booster Energy Efficiency}

\subsection{Scaling and training time}
\label{appendix:scaling}

\begin{figure}[ht]
\begin{subfigure}{.5\textwidth}
  \centering
  \includegraphics[width=\textwidth]{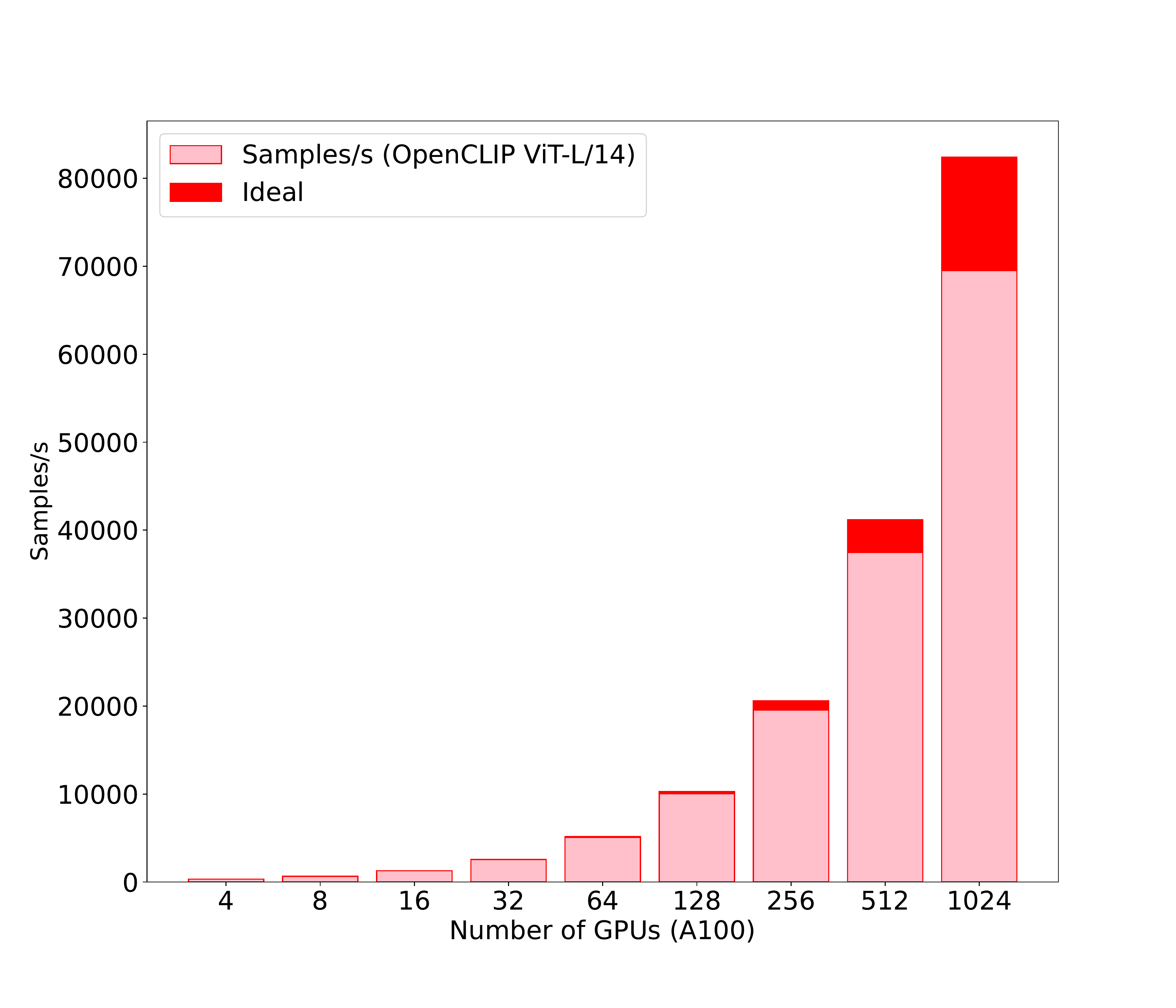}
  %\caption{Put your sub-caption here}
  \caption{Throughput in Im/s}
  \label{fig:throughput_clip}
\end{subfigure}
\begin{subfigure}{.5\textwidth}
  \centering
  % include second image
  \includegraphics[width=\textwidth]{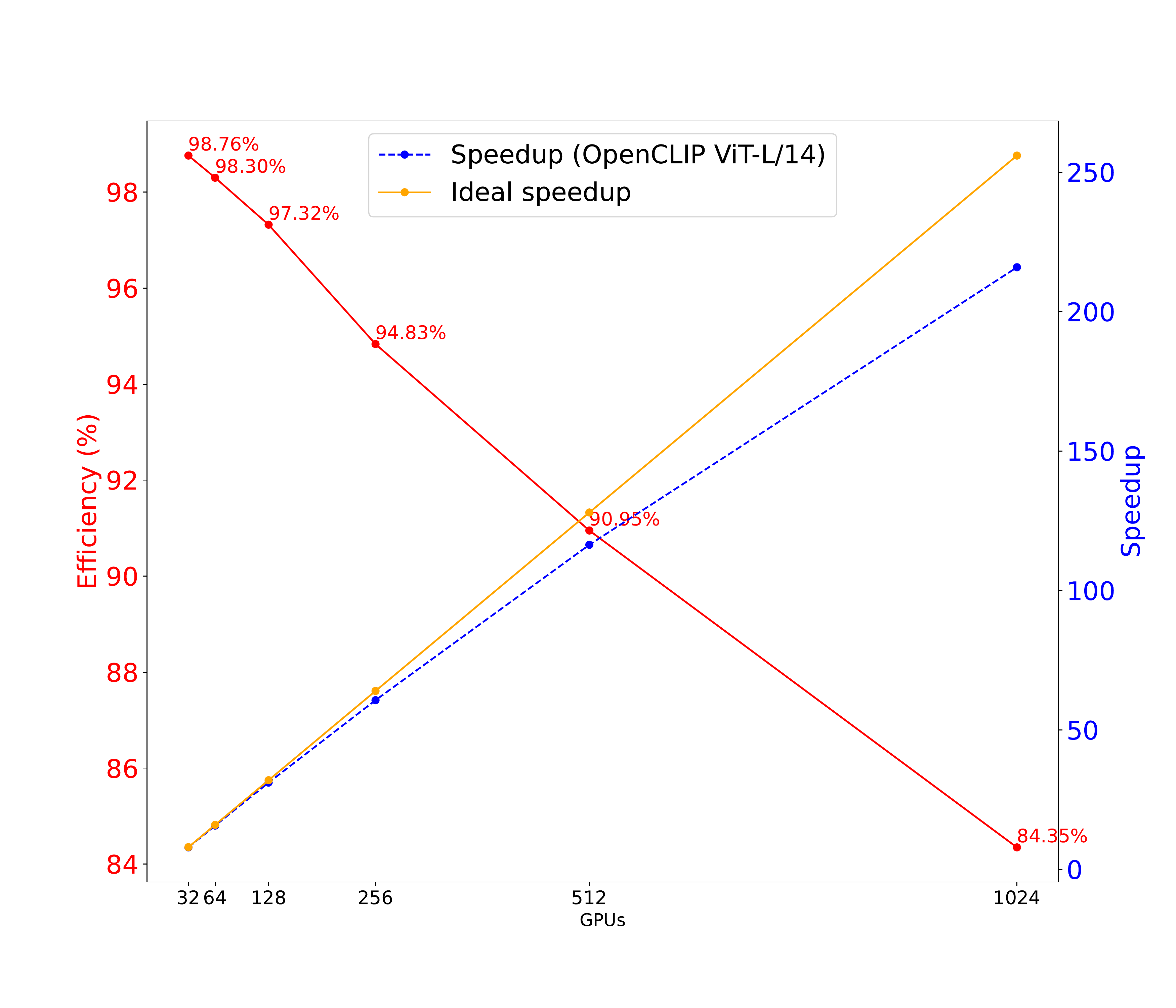}
  %\caption{Put your sub-caption here}
  \caption{Speedup and efficiency}
  \label{fig:speedup_clip}
\end{subfigure}
%\vspace*{-0.2cm}
\caption{Distributed training for OpenCLIP ViT-L/14, scaling behavior on the supercomputer using A100 GPUs while varying the number of GPUs. In Figure~\ref{fig:throughput_clip},  we  show  the  raw throughputs and in Figure~\ref{fig:speedup_clip} we show speedup and efficiency we obtain in the same setup, relative to training with a single node (each node contains 4 GPUs).}
\label{fig:scaling_R152}
\end{figure}

Here, we report scaling behavior during large-scale pre-training  using ViT-L/14 as a vision backbone with OpenCLIP \citep{ilharco_gabriel_2021_5143773}.
We performed scaling experiments to assess the scalability of data parallel training distributed across many GPUs on multiple nodes using PyTorch DDP.
The efficiency in Figure \ref{fig:speedup_clip} is computed using the following formula: $E(N) = 100 \times \frac{T(N)}{N \times T(1)}$. $T(N)$ is the total measured throughput in Im/s for $N$ GPUs. The best achievable efficiency, when scaling is perfect, is 100\%.We observe that scaling is sufficiently close to ideal linear, staying above $\approx 84\%$ for 1024 GPUs (256 nodes).
We also provide the raw throughput (Im/s) numbers in Figure \ref{fig:throughput_clip}. 
    
% \caption{Distributed training for OpenCLIP ViT-L/14, scaling behavior on JUWELS Booster using A100 GPUs while varying the number of GPUs. In Figure~\ref{fig:throughput_clip},  we  show  the  raw throughputs and in Figure~\ref{fig:speedup_clip} we show speedup and efficiency we obtain in the same setup, relative to training with a single node (each node contains 4 GPUs).}

% A pre-training with large R152x4 on takes about 81 hours on ImageNet-21k and x hours on ImageNet-1K with $256$ A100 GPUs, while pre-training with small R50x1 

\subsection{Sharding contrastive loss}

The InfoNCE loss \cite{oord2018representation} used by CLIP can be thought of as a method to maximize the mutual information between text and image representations. Formally, Oord et al. express that $I(X;Y)\geq \log(N)-\mathcal{L}_N$, $N$ denoting batch size and $\mathcal{L}_N$ representing the InfoNCE loss. As a result of this lower bound, maximizing the batch size will maximize our mutual information.

Radford et al. \citep{radford2021learning} take advantage of this bound and use $N=32,768$ to train CLIP. Such a batch size necessitates the sharding of computation. Although the original CLIP paper points towards this notion, the implementation details are nontrivial.

Before sharding, the similarity scores will take up $\mathcal{O}(N^2)$ memory on each worker, totalling to 4 GB of VRAM in FP32. After sharding memory reduces to instantiating two $n\times N$ matrices, $n$ being the batch size allocated to each worker. Using a local batch size of 256, the similarity matrices now occupy 64 MB of memory in FP32.

To achieve this memory reduction, we can eliminate redundant computations and compute the similarities of local features versus all features. When aggregated across all machines, this achieves identical gradients. However, it should be noted that the all-gather method is imperative for correct gradient calculation. PyTorch's standard \texttt{torch.distributed.all\_gather} can not be differentiated through, while \texttt{torch.distributed.nn.functional.all\_gather} can be. Thus, we require the use of the latter to correctly calculate the gradients in a distributed manner.

\subsection{Training instabilities}

As parameterization increased within our training runs, so did model model instability. Half-way through the runs of ViT L/14 H/14 and g/14, NaN values and loss spikes began occurring.

To address these issues, we attempted to use extra normalization layers, add scaled cosine attention, resume many steps before crashes, and implement other architecture tweaks with no success. What ended up solving the stability issues was increasing precision.

Using Automatic Mixed Precision (AMP) with bfloat16 over float16, or float32 with tensor-float32 resolved the issues mentioned above. We also have observed that even the smaller ViT-B models with AMP can become unstable when learning rate and batch size become sufficiently large, suggesting a generic scheme behind the phenomenon where frequency of instabilities occurring during the training is a function of model scale and global batch size.

\section{Experimental details}
\label{sec:appendix:experiments}

\subsection{Datasets employed in experiments.}
\label{sec:appendix:datasets}

\textbf{LAION-400M and LAION-5B.} Both LAION-400M \citep{schuhmann2021laion} and LAION-5B \citep{Schuhmann2022} are open, public image-text datasets that were composed by obtaining links from Common Crawl \citep{common_crawl}. While LAION-400M contains 414M english image-text pairs, LAION-5B is currently the largest public image-text dataset containing over 5.8 billion multi-lingual image-text examples. In both cases, samples are obtained by filtering a subset of Common Crawl with a pre-trained OpenAI ViT B/32 model. LAION-5B contains an English image-text subset of 2.32 billion samples, to which we refer as LAION-2B in this work. Besides the open nature of the datasets, a further advantage is full transparency about the dataset composition and assembly, with software stack and tools around LAION-400M and LAION-5B released as open-source, increasing reproducibility of experiments. This already resulted in numerous works using the datasets for training state-of-the-art language-vision models \citep{rombach2021highresolution, mustafa2022multimodal, ho2022imagen, wang2022internimage, fang2022eva}, %which also provides a validation for using those datasets for studying scaling laws in this work.  
validating the usage of those datasets for studying scaling laws in this work.

\begin{table}
\centering
        \begin{footnotesize}
        \rowcolors{2}{light-light-gray}{white}
        \begin{tabular}{cc}\hline
        \textbf{Dataset}  & \textbf{\# English Img-Txt Pairs} \\
        \hline
        \multicolumn{2}{c}{\textbf{Public Datasets}}         \\ \hline
        {LAION-400M} & 407M \\
        {LAION-2B} & 2.3B                    \\ \hline
        \multicolumn{2}{c}{\textbf{Private Datasets}}         \\ \hline
        CLIP WIT (OpenAI) & 400M                    \\
        ALIGN    & 1.8B                    \\
        BASIC    & 6.6B                      \\ \hline
        \end{tabular}
        \end{footnotesize}

  \caption{\textbf{Open LAION datasets used for pre-training in this study.} Adapted from \citep{Schuhmann2022}. LAION-2B is a subset of multi-lingual LAION-5B and is more than 20 times larger than other public English image-text datasets. The scale of LAION-2B is comparable to the largest private dataset used for language-vision model training.}%
  \label{fig:laion_datasets}
\end{table}

\textbf{Downstream transfer and fine-tuning datasets.} For downstream classification tasks, in addition to standard ImageNet, we follow~\cite{Schuhmann2022} and use VTAB+, a collection of datasets in VTAB together with ImageNet derived robustness datasets and additional datasets, forming a comprehensive set of 35 tasks. For evaluating retrieval, we make use of MS-COCO and Flickr30K. For fine-tuning, we make use of a dedicated ImageNet-12k dataset  (12M training examples, 470K validation examples) which is a subset of the full ImageNet-22k (14M examples) that we employ for the multi-stage fine tuning procedure described in Sec. \ref{subsect:scaling_laws_finetune}. For more details on downstream datasets, refer to Table \ref{tab:datasets_list}.

\begin{figure}[ht]
\centering
\includegraphics[width=\textwidth]{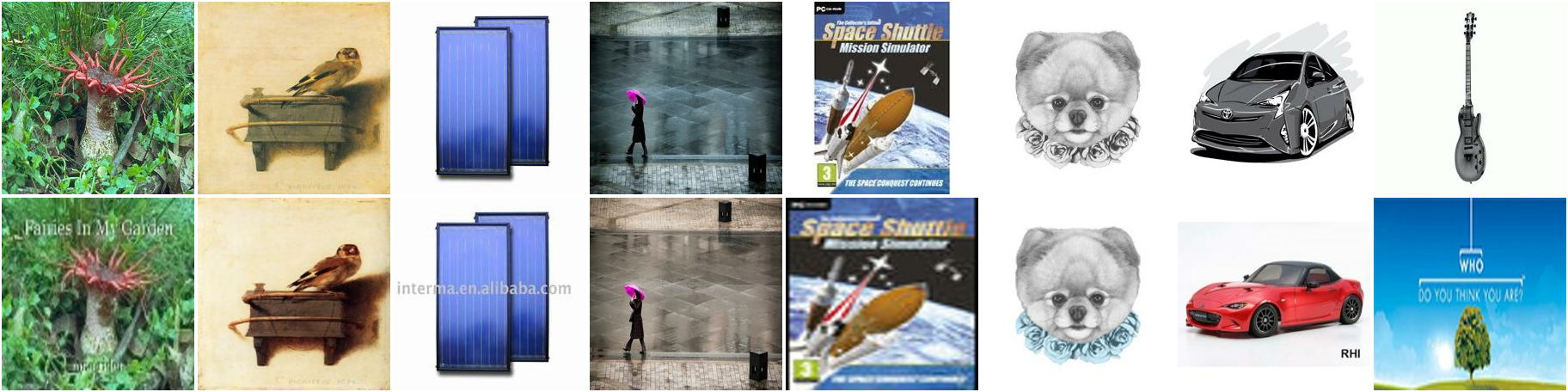}
%\caption{Put your sub-caption here}
\caption{Duplicate images detected using pHash\cite{zauner2010implementation} between downstream datasets and LAION-400M. Top row shows images from downstream datasets, while bottom row show corresponding detected duplicates in LAION-400M. We observe near-duplicate detection for a variety of image transformations: blurring, text blitting, color transformations, cropping, and scaling.
Last two columns show false positive examples detected on ImageNet-Sketch dataset. In general, we observed that most of false positive cases had a uniform background, which pHash seems to be sensitive to.
}
\label{fig:dups_viz}
\end{figure}

%ToDo Mehdi (DONE)
\textbf{Duplication check for pre-training and downstream datasets.}. To ensure that images from downstream datasets are not contained in LAION, we conduct a simple
duplication check based on the perceptual image hash library pHash~\cite{zauner2010implementation}.
We apply pHash's discrete cosine transform (DCT) method on LAION-400M images and images from downstream datasets. Afterwards, for each downstream dataset, we count the number of duplicates by finding the hashes that are also present in LAION-400M. We provide the overlap percentage found on a subset of  downstream datasets in Table~\ref{table:dups}. In Figure~\ref{fig:dups_viz}, we also provide a sample of images from downstream datasets detected as duplicates in LAION-400M. Overall, the ratio of detected duplicates is around 1\%, except on ImageNet-R (3.80\%) and ImageNet-Sketch (5.15\%). We investigate further and re-evaluate zero-shot performance of our pre-trained Vit-H/14 on ImageNet-R and ImageNet-Sketch by removing duplicates from their test sets. For ImageNet-R, zero-shot top-1 accuracy goes from 89.32\% to 89.21\% after removing duplicates. For ImageNet-Sketch, zero-shot top-1 accuracy goes from 66.57\% to 66.59\% after removing duplicates.
%This amount of duplicates is still low, and the two datasets do not have significant impact on scaling laws trends observed here. 
We conclude, based on those results, that it is unlikely that downstream results would be affected by the duplicates. This would be in line with previous works\citep{radford2021learning, zhai2021lit} which explicitly measured and compared performance on deduplicated downstream datasets, reporting that duplication in test sets do not significantly alter most results. This is likely due to the very large scale and diversity of pre-training data. We leave more elaborated duplication detection procedures for future work. 

% We investigate further those two datasets and find XXX.

\begin{table}
\centering
\rowcolors{2}{light-light-gray}{white}
\begin{tabular}{ll}
\toprule
           \textbf{Dataset} &    \textbf{Overlap\%} \\
\midrule
       ImageNet & 1.02 \\
    ImageNet-v2 & 1.35 \\
     ImageNet-R & 3.80 \\
ImageNet Sketch & 5.15 \\
     ImageNet-A & 0.40 \\
      ObjectNet & 0.10 \\
      CIFAR-100 & 0.02 \\
       CIFAR-10 & 0.03 \\
        MS-COCO & 1.12 \\
      Flickr30K & 1.30 \\
\bottomrule
\end{tabular}
\caption{Ratio of images (\%) on downstream datasets that were detected on LAION-400M, using pHash~\cite{zauner2010implementation}.}
  \label{table:dups}
\end{table}

\subsection{Further experimental results}
\label{sec:appendix:exp_further_results}

\subsubsection{Predictions derived from scaling laws}
\label{sec:appendix:predictions}
%TODO: here predictive scaling plots, at least for zero-shot imagenet-1K and retrieval (DONE)

\begin{figure}[ht]
\begin{subfigure}{.5\textwidth}
  \centering
  \includegraphics[width=\textwidth]{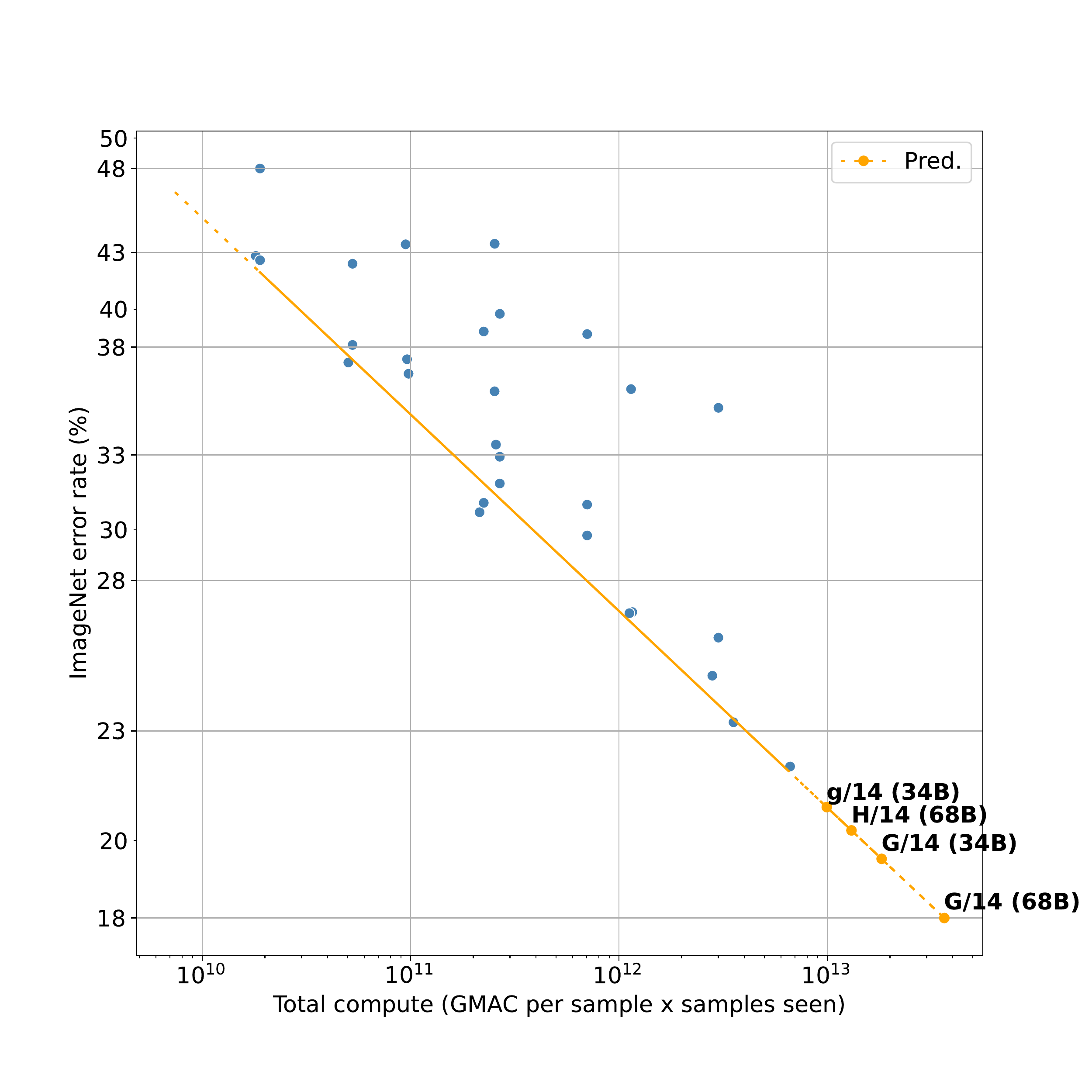}
  %\caption{Put your sub-caption here}
  \caption{Predictions on ImageNet}
  \label{fig:predictionplot_imagenet}
\end{subfigure}
\begin{subfigure}{.5\textwidth}
  \centering
  % include second image
  \includegraphics[width=\textwidth]{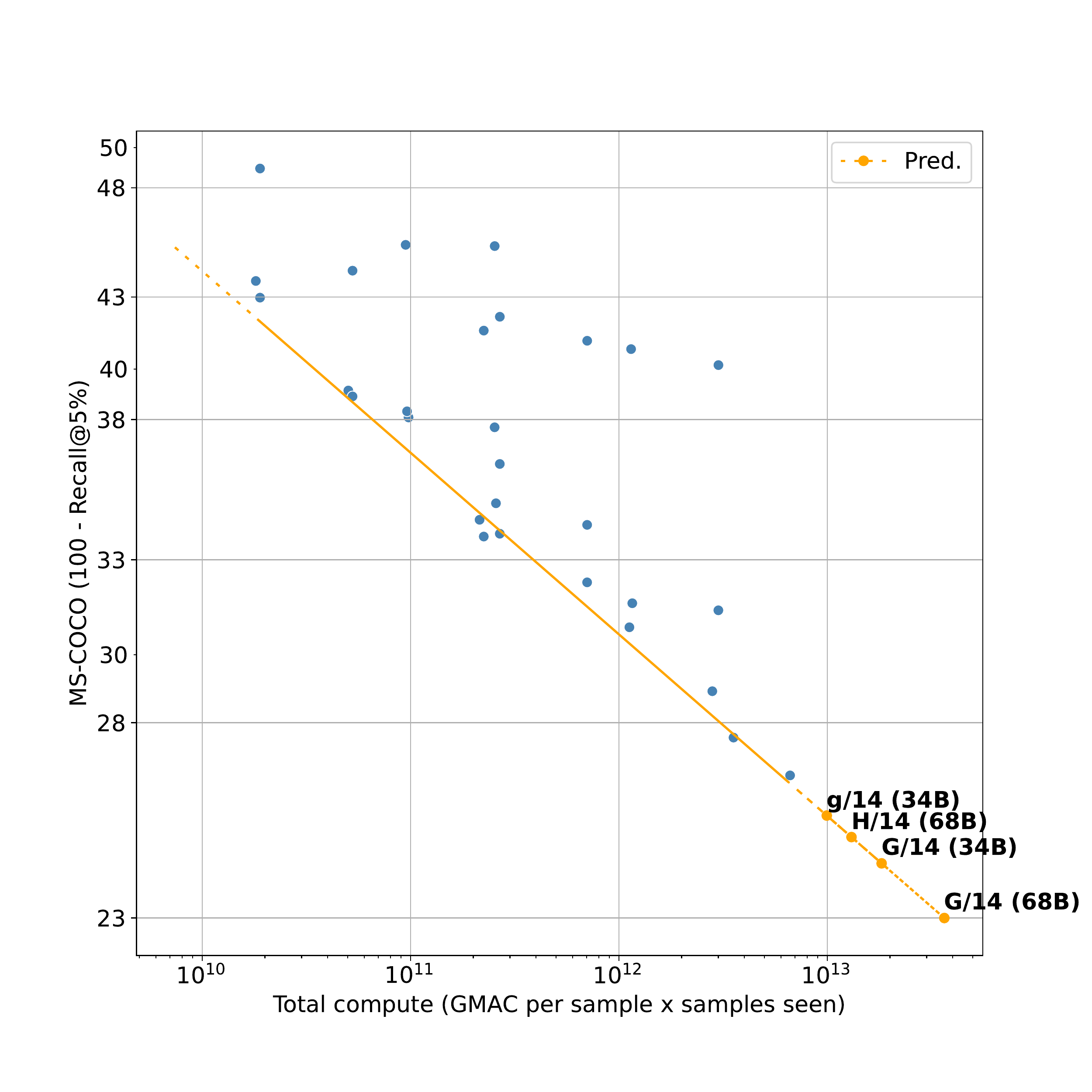}
  %\caption{Put your sub-caption here}
  \caption{Predictions on MS-COCO}
  \label{fig:predictionplot_coco}
\end{subfigure}
%\vspace*{-0.2cm}
\caption{Zero-shot performance extrapolation of g/14, H/14 and G/14 on larger scales. We fit a power-law on the Pareto frontier of available models. In Fig.\ref{fig:predictionplot_imagenet} we show the predictions for ImageNet classification, while in Fig.\ref{fig:predictionplot_coco} we show the predictions for MS-COCO image retrieval.}
\label{fig:predictionplot}
\end{figure}

\begin{table}
%\footnotesize
\centering
\rowcolors{2}{light-light-gray}{white}
\begin{tabular}{cll}
\toprule
\textbf{Model} & \textbf{ImageNet top-1 (\%)} & \textbf{MS-COCO Recall@5 (\%)} \\
\midrule
%H/14 (68B) &              79.73 &             75.03 \\
%g/14 (34B) &              79.11 &             74.48 \\
%g/14 (68B) &              80.66 &             75.85 \\
%G/14 (13B) &              78.26 &             73.75 \\
%G/14 (34B) &              80.47 &             75.68 \\
%G/14 (68B) &              81.92 &             76.99 \\
 H/14 (3B) &       70.78 &            67.58 \\
 g/14 (3B) &       72.11 &            68.65 \\
 G/14 (3B) &       73.93 &            70.12 \\
H/14 (13B) &       75.62 &            71.52 \\
g/14 (13B) &       76.66\textbf{*} &            72.40\textbf{*} \\
G/14 (13B) &       78.26 &            73.75 \\
H/14 (34B) &       77.97\textbf{*} &            73.43\textbf{*} \\
g/14 (34B) &       79.11 &            74.48 \\
G/14 (34B) &       80.47 &            75.68 \\
H/14 (68B) &       79.73 &            75.03 \\
g/14 (68B) &       80.66 &            75.85 \\
G/14 (68B) &       81.92 &            76.99 \\

\bottomrule
\end{tabular}
\caption{Performance extrapolation of g/14, H/14 and G/14 on larger scales corresponding to Fig.\ref{fig:predictionplot}. We fit a
power-law on the Pareto frontier of available models. We show the zero-shot top-1 accuracy predictions for ImageNet and zero-shot retrieval image retrieval Recall@5 predictions for MS-COCO. We denote by  (*) actual measured model performance values, while the rest are predictions.}
\label{table:predictions}
\end{table}

We can use scaling laws derived from our measurements to predict model performance for larger scales on different downstream tasks. To perform predictions, we fit a power-law on the Pareto frontier\footnote{Since total compute budget (measured in GMAC) of different trained models are not exactly aligned, we adopt a binning approach. We bin the GMAC compute budget axis and compute the optimal performance within each bin, then fit a line in log-log space on the resulting bins.}.  Fig.\ref{fig:predictionplot_imagenet} and Fig.\ref{fig:predictionplot_coco} show extrapolation of performance for ImageNet and MS-COCO, respectively. According to the predictions, H/14 (68B samples seen) would achieve 79.73\% (+1.76\%) zero-shot top-1 accuracy on ImageNet and 75.10\% (+1.60\%) image retrieval Recall@5 on MS-COCO, compared to our trained H/14 (34B samples seen).  See also Tab.\ref{table:predictions} for more detailed extrapolation numbers.
For g/14 (68B samples seen), we predict 80.66\% (+4\%) zero-shot top-1 accuracy on ImageNet and 75.85\% (+3.45\%) image retrieval Recall@5 on MS-COCO, compared to our trained g/14 (13B samples seen). 
On the largest compute budget we consider, G/14 (68B samples seen), we predict 81.92\%  zero-shot top-1 accuracy on ImageNet and 76.99\% image retrieval Recall@5 on MS-COCO.
%One way is to predict model performance for a certain model, data and sample seen scale combination. Another way is the predict optimal model performance for a given total compute scale that may correspond to different combinations of model and samples seen scale, across different data scales.

% (Yet Another way is the predict optimal model performance for a given scale (eg model scale) while assuming that other scales (eg data scale and samples seen) can be freely varied )

\subsubsection{Fine-tuning}
\label{sec:appendix:finetuning}

\begin{figure}
    \centering
    \includegraphics[width=\textwidth]{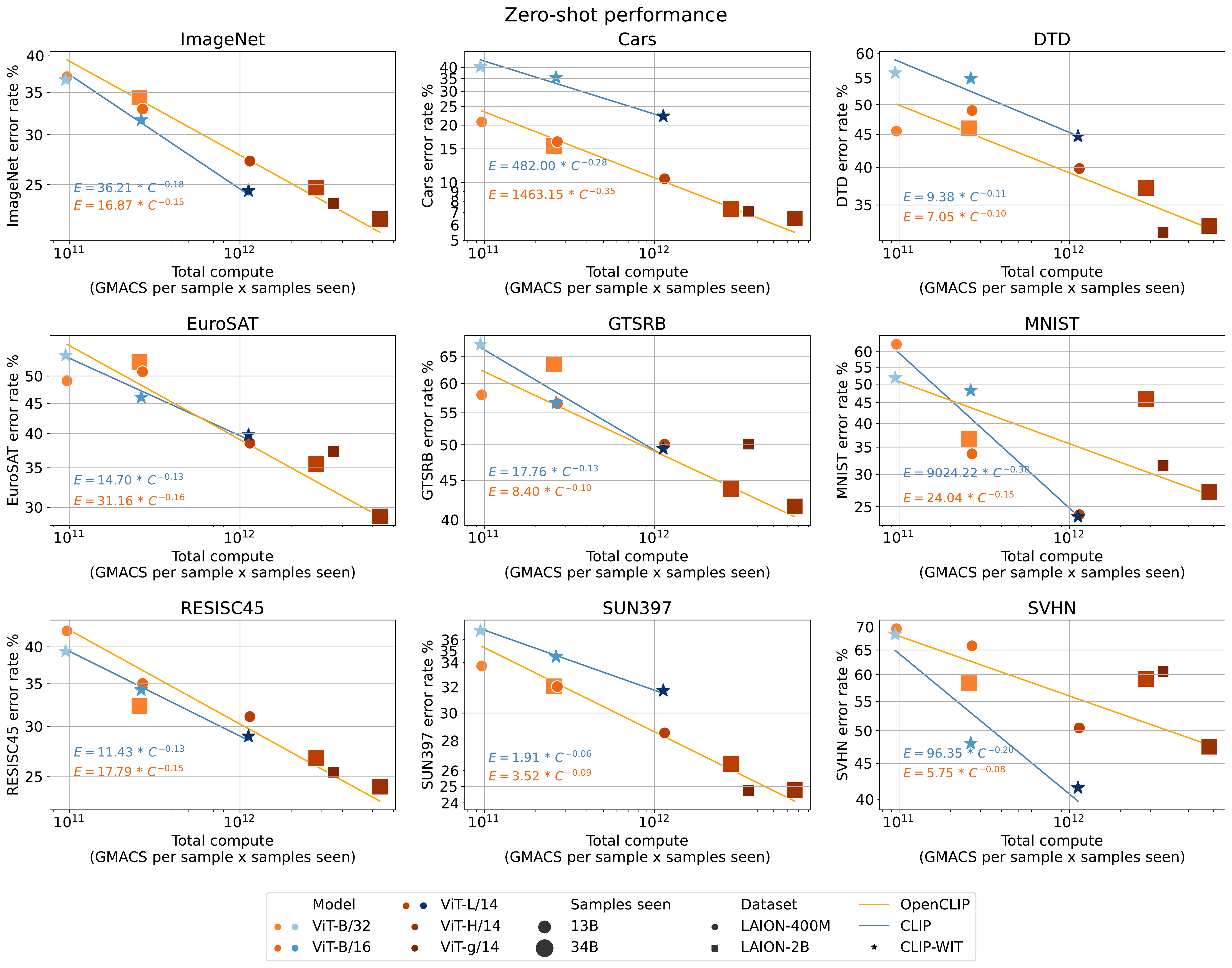}
    \caption{Scaling trends of zero-shot models on the eight other downstream tasks used for the fine-tuning experiments in Section \ref{subsect:scaling_laws_finetune} and on ImageNet.}
    \label{fig:ft-eight-zs}
\end{figure}

\begin{figure}
    \centering
    \includegraphics[width=\linewidth]{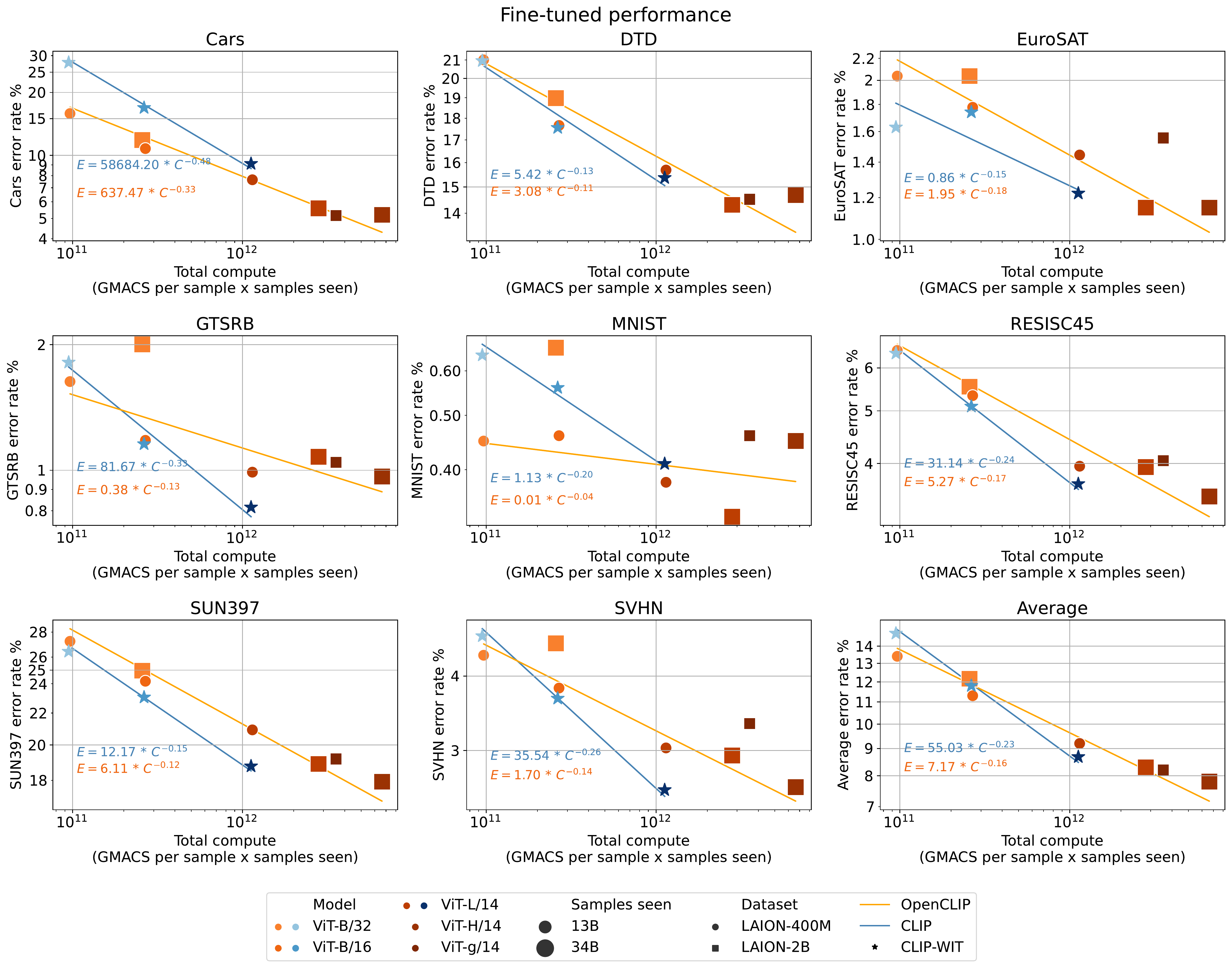}
    \caption{Scaling trends of fine-tuned models on the eight other downstream tasks used for the fine-tuning experiments in Section \ref{subsect:scaling_laws_finetune}.}
    \label{fig:ft-eight-ft}
\end{figure}

In Table~\ref{table:finetune_results}, we show detailed results of fine-tuning on ImageNet with and without extra data (Imagenet-12k), and show results of the fine-tuned models on five ImageNet robustness test sets.
Also, complementing the results shown in Figure \ref{fig:ft-eight} in Section \ref{subsect:scaling_laws_finetune}, we show a per-task breakdown of the the zero-shot and fine-tuned performance on the eight classification tasks in Figures \ref{fig:ft-eight-zs} and  \ref{fig:ft-eight-ft}. Exact numbers are shown in Tables \ref{tab:appendix:joint_ft_zs}, \ref{tab:appendix:joint_ft_ft}, and \ref{tab:appendix:joint_ft_patch}.


\begin{figure}
    \centering
    \includegraphics[width=\linewidth]{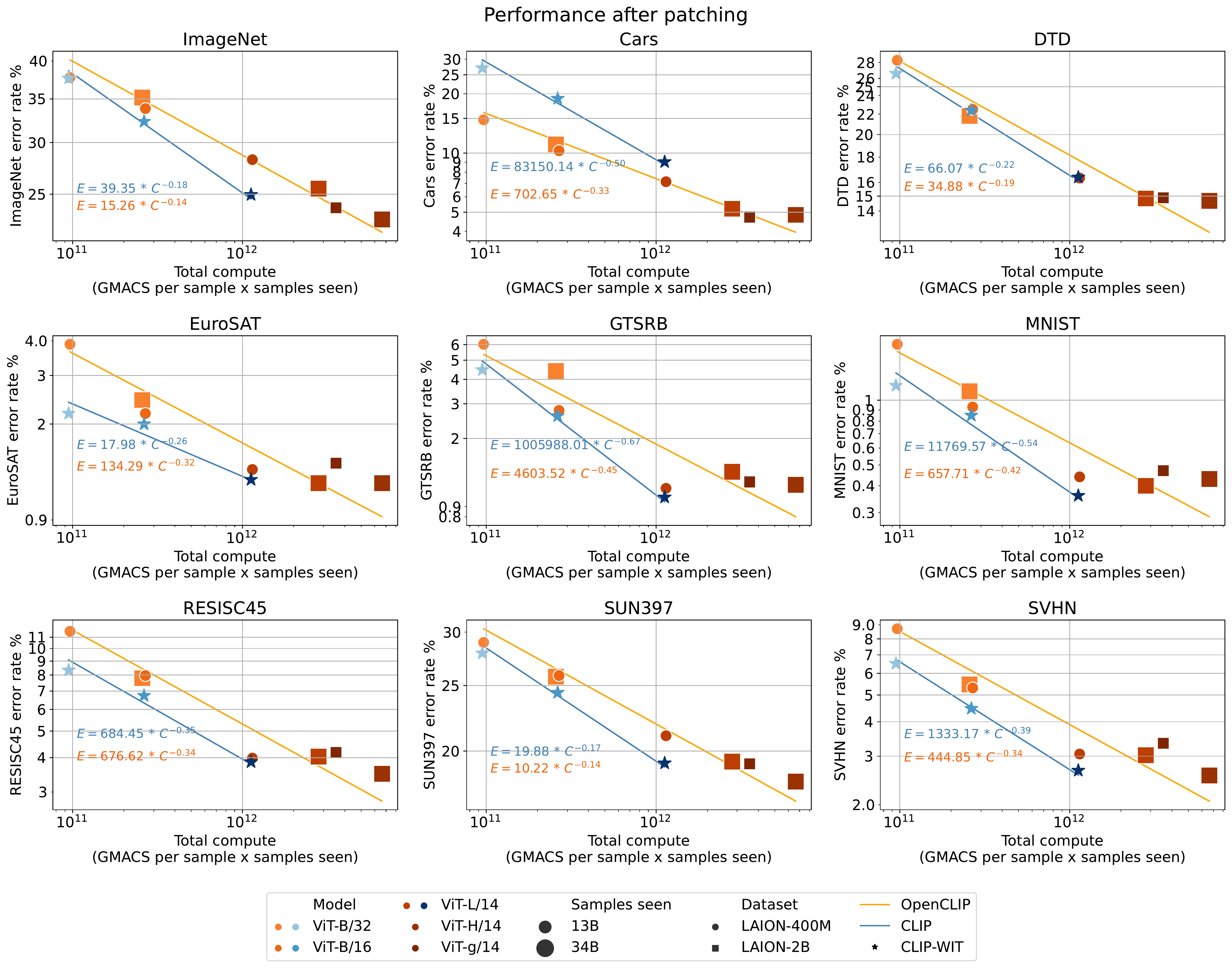}
    \caption{Scaling trends of patched models \cite{ilharco2022patching}, on ImageNet and eight other downstream tasks used for the fine-tuning experiments in Section \ref{subsect:scaling_laws_finetune}.}
    \label{fig:ft-eight-patch}
\end{figure}

Moreover, since fine-tuning on some downstream tasks can decrease accuracy on others, we experiment with model patching by interpolating between the weights of fine-tuned and zero-shot models, as in Ilharco et al. \cite{ilharco2022patching}.\footnote{The weights $\theta_\textrm{patched}$ of the patched model are obtained via the equation $\theta_\textrm{patched} = (1-\alpha)\theta_\textrm{zero-shot} + \alpha \theta_\textrm{fine-tuned}$, where $\alpha\in[0,1]$ is the mixing coefficient.}
We choose the mixing coefficient $\alpha \in {0, 0.1, ..., 1.0}$ that maximizes average accuracy on the eight downstream tasks, while accuracy on ImageNet---used as a control---decreases by one percentage point or less.
In Figure \ref{fig:ft-eight-patch}, we show how scale affects performance on the eight tasks we fine-tune one, along with that on ImageNet. 

Finally, Tables \ref{table:imagenet_ft_hparam} and \ref{table:imagenet_ft12k_hparam} include hparam templates for reproducing ImageNet fine-tune results. Once published, the individual model weights will include their specific training hyper-parameters as there is some variation in specific instances (i.e. at different upscale sizes, from 12k to 1k). Motivated by BEiT \cite{bao2021beit}, all ImageNet fine-tune runs make use of layer-wise learning-rate decay (also known as discriminative fine-tuning \cite{howard2018universal}); this is an important parameter that needs tuning per model size along with the learning-rate itself.

\begin{table}
\footnotesize
\centering
\rowcolors{2}{light-light-gray}{white}
\begin{tabular}{lllrrrrrrr}
\toprule
& & & &  \multicolumn{3}{c}{ImageNet} & \multicolumn{3}{c}{CIFAR100} \\
 Model & Samples Seen & Dataset & VTAB & 10 shot & 25 shot & Full & 10 shot & 25 shot & Full \\
\midrule

           ViT-B/32 &                 13B &         CLIP-WIT &    69.71 &                          59.16 &                          65.27 &                          75.61 &             63.93 &             70.64 &             79.97 \\
           ViT-B/32 &                 13B &       LAION-400M &    71.84 &                          59.36 &                          65.17 &                          74.90 &             70.50 &             75.18 &             82.92 \\
%          ViT-B/32 &                 34B &         LAION-2B &    71.43 &                          61.89 &                          67.49 &                          76.60 &             74.24 &             78.73 &             85.06 \\
           ViT-B/32 &                 34B &         LAION-2B &    71.53 &                          62.40 &                          67.98 &                          76.93 &             75.47 &             79.97 &             85.99 \\
          ViT-B/16 &                 13B &         CLIP-WIT &    71.25 &                          65.42 &                          70.97 &                          79.82 &             68.91 &             74.67 &             82.40 \\
           ViT-B/16 &                 13B &       LAION-400M &    72.72 &                          64.46 &                          69.94 &                          78.74 &             71.96 &             77.21 &             84.07 \\
           ViT-L/14 &                 13B &         CLIP-WIT &    73.77 &                          73.51 &                          77.67 &                          84.39 &             77.57 &             81.91 &             87.14 \\
           ViT-L/14 &                 13B &       LAION-400M &    73.98 &                          70.86 &                          75.02 &                          81.77 &             78.06 &             82.48 &             87.95 \\
           ViT-L/14 &                 34B &         LAION-2B &    74.48 &                          73.94 &                          77.45 &                          83.46 &             82.76 &             86.04 &             90.14 \\
           ViT-H/14 &                 34B &         LAION-2B &    \textbf{75.96} &                          \textbf{75.79} &                          \textbf{79.07} &                          \textbf{84.85} &             \textbf{84.74} &             \textbf{87.82} &             \textbf{91.43} \\
           ViT-g/14 &                 13B &         LAION-2B &    75.18 &                          74.87 &                          78.25 &                          84.09 &             84.66 &             87.76 &             91.09 \\
\bottomrule
\end{tabular}
\caption{
Scaling model and data size leads to lower error linear classifers on ImageNet~\cite{Deng2009a}, CIFAR100~\cite{Krizhevsky2009}, and the visual task adaptation benchmark (VTAB)~\cite{zhai2019large}.
     We train linear probes for models with at least 13B samples seen.
     We train probes by first caching the image features, thus no data augmentation is used.
     $k$ shot denotes that $k$ images per-class are used to train the linear probe.
     }
    \label{tab:lp}
\end{table}

\begin{table}
\scriptsize
\centering
\setlength\tabcolsep{3pt}
\rowcolors{2}{light-light-gray}{white}
\begin{tabular}{lrlrrrrrrrrr}     
\toprule                           

& & & \multicolumn{9}{c}{Top-1 Accuracy (\%)} \\
Arch. & \# samples &  Dataset & ImageNet & Cars & DTD & EuroSAT & GTSRB & MNIST & RESISC45 & SUN397 & SVHN \\                                   
\midrule                                                                                                                                                                           ViT-B/32 &          13B &    CLIP-WIT &         63.35 &     59.73 &    43.99 &        45.81 &      32.56 &      48.25 &         60.65 &       63.18 &     31.61 \\                                  
 ViT-B/32 &          13B &  LAION-400M &         62.94 &     79.24 &    54.47 &        50.89 &      41.98 &      37.44 &         57.62 &       66.28 &     30.36 \\                                  
ViT-B/32 &          34B &    LAION-2B &         65.63 &     84.45 &    54.04 &        47.22 &      36.48 &      63.34 &         67.70 &       67.94 &     41.66 \\                                   
ViT-B/16 &          13B &    CLIP-WIT &         68.33 &     64.61 &    45.11 &        53.96 &      43.34 &      51.80 &         65.76 &       65.50 &     51.98 \\                                  
ViT-B/16 &          13B &  LAION-400M &         67.05 &     83.63 &    51.01 &        49.15 &      43.45 &      66.29 &         64.97 &       67.96 &     34.12 \\          

ViT-L/14 &          13B &    CLIP-WIT &         75.54 &     77.75 &    55.32 &        60.22 &      50.55 &      \textbf{76.36} &         71.05 &       68.28 &     \textbf{58.45} \\                                  
ViT-L/14 &          13B &  LAION-400M &         72.75 &     89.53 &    60.16 &        61.48 &      49.89 &      76.09 &         68.92 &       71.44 &     49.54 \\                                  
ViT-L/14 &          34B &    LAION-2B &         75.26 &     92.71 &    62.82 &        64.44 &      56.14 &      54.10 &         73.25 &       73.56 &     40.84 \\      ViT-H/14 &          34B &    LAION-2B &         \textbf{77.95} &     \textbf{93.50} &    67.50 &        \textbf{71.04} &      \textbf{58.35} &      72.83 &         \textbf{75.87} &       75.23 &     52.51 \\                           
ViT-g/14 &          13B &    LAION-2B &         76.66 &     92.90 &    \textbf{68.24} &        62.70 &      49.87 &      68.46 &         74.57 &       \textbf{75.24} &     39.34 \\                                 
\bottomrule
\end{tabular}
\caption{
    Zero-shot accuracy for various models on downstream tasks from Section \ref{sec:appendix:finetuning}.
    \label{tab:appendix:joint_ft_zs}
    }
\end{table}

\begin{table}
\scriptsize
\centering
\setlength\tabcolsep{5pt}
\rowcolors{2}{light-light-gray}{white}
\begin{tabular}{lrlrrrrrrrr}     
\toprule                           

& & & \multicolumn{8}{c}{Top-1 Accuracy (\%)} \\
Arch. & \# samples &  Dataset & Cars & DTD & EuroSAT & GTSRB & MNIST & RESISC45 & SUN397 & SVHN \\                                      
\midrule                                                                                    ViT-B/32 &          13B &    CLIP-WIT &     72.19 &    79.04 &        98.37 &      98.19 &      99.36 &         93.62 &       73.57 &     95.33 \\                                                  
 ViT-B/32 &          13B &  LAION-400M &     84.14 &    78.99 &        97.96 &      98.37 &      99.55 &         93.54 &       72.76 &     95.67 \\                                                  
 ViT-B/32 &          34B &    LAION-2B &     88.16 &    81.01 &        97.96 &      98.00 &      99.34 &         94.46 &       75.04 &     95.46 \\                             ViT-B/16 &          13B &    CLIP-WIT &     83.10 &    82.45 &        98.26 &      98.84 &      99.44 &         94.90 &       76.94 &     96.33 \\                                                  
ViT-B/16 &          13B &  LAION-400M &     89.22 &    82.34 &        98.22 &      98.82 &      99.54 &         94.67 &       75.81 &     96.18 \\                                                                      
ViT-L/14 &          13B &    CLIP-WIT &     90.87 &    84.63 &        98.78 &      \textbf{99.18} &     99.59 &         96.33 &       81.22 &     \textbf{97.42} \\                                                  
 ViT-L/14 &          13B &  LAION-400M &     92.35 &    84.31 &        98.56 &      99.01 &      99.62 &         96.05 &       79.08 &     96.97 \\                                                  
 ViT-L/14 &          34B &    LAION-2B &     94.42 &    \textbf{85.69} &        \textbf{98.85} &      98.92 &      \textbf{99.67} &         96.06 &       81.10 &     97.06 \\
  ViT-H/14 &          34B &    LAION-2B &     94.80 &    85.32 &        \textbf{98.85} &      99.03 &      99.55 &         \textbf{96.52} &       \textbf{82.08} &     97.40 \\                                                  
 ViT-g/14 &          13B &    LAION-2B &     \textbf{94.84} &    85.48 &        98.44 &      98.95 &      99.54 &         95.95 &       80.82 &     96.67 \\                          
\bottomrule
\end{tabular}
\caption{
    Accuracy after fine-tuning for various models on downstream tasks from Section \ref{sec:appendix:finetuning}. We fine-tune jointly on the eight downstream image classification tasks, alternating batches from each task.
    We fine-tune only the parameters of the vision encoder, using a fixed classification head for each task initialized with the weights from the zero-shot model.
    \label{tab:appendix:joint_ft_ft}
    }
\end{table}

\begin{table}
\scriptsize
\centering
\setlength\tabcolsep{3pt}
\rowcolors{2}{light-light-gray}{white}
\begin{tabular}{lrlrrrrrrrrr}     
\toprule                           

& & & \multicolumn{9}{c}{Top-1 Accuracy (\%)} \\
Arch. & \# samples &  Dataset & ImageNet & Cars & DTD & EuroSAT & GTSRB & MNIST & RESISC45 & SUN397 & SVHN \\                                                    
\midrule                                                                                 ViT-B/32 &          13B &    CLIP-WIT &         62.36 &     72.96 &    73.40 &        97.81 &      95.53 &      98.83 &         91.67 &       72.08 &     93.50 \\
ViT-B/32 &          13B &  LAION-400M &         62.27 &     85.24 &    71.70 &        96.11 &      93.97 &      98.18 &         88.44 &       71.03 &     91.29 \\
ViT-B/32 &          34B &    LAION-2B &         64.84 &     88.93 &    78.19 &        97.56 &      95.60 &      98.90 &         92.21 &       74.22 &     94.53 \\
ViT-B/16 &          13B &    CLIP-WIT &         67.70 &     81.07 &    77.61 &        98.00 &      97.40 &      99.15 &         93.27 &       75.59 &     95.53 \\
ViT-B/16 &          13B &  LAION-400M &         66.18 &     89.73 &    77.50 &        97.81 &      97.22 &      99.07 &         92.03 &       74.13 &     94.69 \\
ViT-L/14 &          13B &    CLIP-WIT &         75.04 &     90.98 &    83.62 &        98.74 &      \textbf{98.99} &      \textbf{99.64} &         96.14 &       80.81 &     97.34 \\
ViT-L/14 &          13B &  LAION-400M &         71.76 &     92.85 &    83.67 &        98.63 &      98.88 &      99.56 &         96.02 &       78.93 &     96.95 \\
ViT-L/14 &          34B &    LAION-2B &         74.50 &     94.79 &    85.16 &        \textbf{98.78} &      98.65 &      99.60 &         95.97 &       80.72 &     96.98 \\
ViT-H/14 &          34B &    LAION-2B &         \textbf{77.12} &     95.15 &    \textbf{85.32} &        \textbf{98.78} &      98.84 &      99.57 &         \textbf{96.51} &       \textbf{81.98} &     \textbf{97.45} \\
ViT-g/14 &          13B &    LAION-2B &         76.16 &     \textbf{95.27} &    85.11 &        98.56 &      98.80 &      99.53 &         95.83 &       80.86 &     96.66 \\        
\bottomrule
\end{tabular}
\caption{
    Accuracy after joint patching \cite{ilharco2022patching} for various models on downstream tasks from Section \ref{sec:appendix:finetuning}. Patching by jointly fine-tuning on the eight tasks with the exception of ImageNet (used only as control), then interpolating the weights of the fine-tuned model with the weights of the zero-shot model. The mixing coefficient for the interpolation is chosen so it maximizes average accuracy on the eight downstream tasks while maintaining ImageNet accuracy within 1 percentage point of the corresponding zero-shot model.
    \label{tab:appendix:joint_ft_patch}
    }
\end{table}

\begin{table}
\scriptsize
\centering
\setlength\tabcolsep{2.5pt}
\rowcolors{2}{light-light-gray}{white}

\begin{tabular}{llllrrrrrrrrr}
\toprule
         & & & & & & & \multicolumn{6}{c}{Top-1 Accuracy (\%)} \\
   Model & Im Size &  Dataset & Extra FT & Params (M) & GMAC & Acts (M) &                 IN & IN-ReaL & IN-V2 &  IN-A &  IN-R & IN-Sketch \\
\midrule
ViT-B/32 &     224 & CLIP-WIT &     None &       88.2 &   4.4 &      5.0 &              81.93 &   87.17 & 70.70 & 22.57 & 55.90 &     45.04 \\
ViT-B/32 &     224 & LAION-2B &     None &       88.2 &   4.4 &      5.0 &              82.58 &   87.54 & 71.21 & 22.85 & 59.16 &     49.07 \\
ViT-B/32 &     224 & LAION-2B &   IN-12k &       88.2 &   4.4 &      5.0 &              83.30 &   87.81 & 72.50 & 30.57 & 57.06 &     45.74 \\
ViT-B/32 &     384 & CLIP-WIT &   IN-12k &       88.3 &  13.1 &     16.5 &              85.11 &   89.04 & 74.53 & 44.75 & 58.21 &     45.75 \\
ViT-B/16 &     224 & CLIP-WIT &     None &       86.6 &  17.6 &     23.9 &              85.28 &   89.16 & 75.57 & 47.23 & 66.02 &     50.94 \\
ViT-B/32 &     384 & LAION-2B &   IN-12k &       88.3 &  13.1 &     16.5 &              85.38 &   89.20 & 75.08 & 47.95 & 60.37 &     47.95 \\
ViT-B/16 &     224 & LAION-2B &     None &       86.6 &  17.6 &     23.9 &              85.47 &   89.43 & 75.13 & 41.57 & 68.75 &     55.40 \\
ViT-B/16 &     384 & CLIP-WIT &     None &       86.9 &  55.5 &    101.6 &              86.24 &   89.71 & 76.68 & 57.55 & 67.22 &     52.15 \\
ViT-B/16 &     384 & LAION-2B &     None &       86.9 &  55.5 &    101.6 &              86.53 &   90.04 & 77.55 & 56.96 & 69.94 &     55.85 \\
ViT-B/16 &     384 & LAION-2B &   IN-12k &       86.9 &  55.5 &    101.6 &              87.17 &   90.11 & 78.16 & 62.61 & 65.53 &     52.62 \\
ViT-L/14 &     224 & LAION-2B &     None &      304.2 &  81.1 &     88.8 &              87.30 &   90.10 & 78.42 & 59.89 & 81.70 &     64.81 \\
ViT-H/14 &     224 & LAION-2B &     None &      632.0 & 167.4 &    139.4 &              87.59 &   90.17 & 79.36 & 65.56 & \textbf{83.28} &     \textbf{67.41} \\
ViT-L/14 &     336 & LAION-2B &     None &      304.5 & 191.1 &    270.2 &              87.78 &   90.30 & 79.07 & 69.03 & 82.60 &     64.79 \\
ViT-L/14 &     224 & CLIP-WIT &     None &      304.2 &  81.1 &     88.8 &              87.85 &   90.31 & \textbf{79.59} & 71.79 & 82.32 &     62.63 \\
ViT-L/14 &     224 & LAION-2B &   IN-12k &      304.2 &  81.1 &     88.8 &              87.89 &   90.30 & 78.51 & 67.01 & 78.26 &     62.06 \\
ViT-L/14 &     336 & LAION-2B &   IN-12k &      304.5 & 191.1 &    270.2 &              88.17 &   90.43 & 78.84 & 73.64 & 77.68 &     60.97 \\
ViT-L/14 &     224 & CLIP-WIT &   IN-12k &      304.2 &  81.1 &     88.8 &              88.17 &   90.37 & 79.38 & 72.33 & 78.68 &     61.40 \\
ViT-H/14 &     224 & LAION-2B &   IN-12k &      632.0 & 167.4 &    139.4 &              88.25 &   90.41 & 79.22 & 70.72 & 82.82 &     65.32 \\
ViT-H/14 &     336 & LAION-2B &   IN-12k &      632.5 & 391.0 &    407.5 &              \textbf{88.50} &   \textbf{90.49} & 79.55 & \textbf{75.68} & 82.26 &     64.62 \\
\bottomrule
\end{tabular}
\caption{Fine-tune results for ImageNet-1k and associated robustness test sets (ImageNet-ReaL~\citep{beyer2020we}, ImageNet-V2~\citep{pmlr-v97-recht19a}, ImageNet-A~\citep{imageneta}, Imagenet-R~\citep{imagenetr}, and ImageNet-Sketch~\citep{imagenetsketch}). Rows with the 'Extra FT' set to IN-12k were fine-tuned on a 12k class subset of ImageNet-22k before fine-tuning on ImageNet.}
\label{table:finetune_results}
\end{table}

\begin{table}
\footnotesize
\centering
\rowcolors{2}{light-light-gray}{white}
\begin{tabular}{lllll}
\toprule
        Hyperparameter &                B/32 &                B/16 &                L/14 &                H/14 \\
\midrule
    Peak Learning-rate &            1.00E-03 &            3.00E-04 &            6.00E-05 &            5.00E-05 \\
            Batch Size &                4096 &                2048 &                2048 &                2048 \\
                Epochs &                  50 &                  50 &                  50 &                  50 \\
         Warmup Epochs &                  10 &                  10 &                  10 &                  10 \\
   Layer-wise LR Decay &                0.65 &                 0.7 &                 0.8 &                0.82 \\
  EMA Weight Smoothing &              0.9998 &              0.9998 &              0.9997 &              0.9998 \\
          Weight Decay &                0.05 &                0.05 &                0.01 &                0.02 \\
       Label Smoothing &                 0.1 &                 0.1 &                 0.1 &                 0.1 \\
          Stoch. Depth &                 0.1 &                 0.1 &                 0.2 &                 0.2 \\
               Dropout &                   0 &                   0 &                   0 &                   0 \\
     Gradient Clipping &                   3 &                   3 &                   3 &                   2 \\
Rand Augment (Uniform) &      M=U(0, 8), N=2 &      M=U(0, 9), N=2 &      M=U(0, 9), N=3 &     M=U(0, 8), N=4  \\
     Random Erase Prob &                 0.3 &                 0.3 &                 0.3 &                 0.3 \\
     Random Resize Crop &                 Yes &                 Yes &                 Yes &                Yes \\
           Mixup Alpha &                   0 &                   0 &                   0 &                   0 \\
          Cutmix Alpha &                   0 &                   0 &                   0 &                   0 \\
          Color Jitter &                   0 &                   0 &                   0 &                   0 \\
\bottomrule
\end{tabular}
\caption{ImageNet fine-tune hyper-parameters.}
\label{table:imagenet_ft_hparam}
\end{table}

\begin{table}
\footnotesize
\centering
\rowcolors{2}{light-light-gray}{white}
\begin{tabular}{lllll}
\toprule
        Hyperparameter &           B/32 &           B/16 &           L/14 &           H/14 \\
\midrule
    Peak Learning-rate &       1.00E-03 &       5.00E-04 &       5.00E-04 &       4.00E-04 \\
            Batch Size &           4096 &           4096 &           4096 &           4096 \\
                Epochs &             60 &             60 &             60 &             60 \\
         Warmup Epochs &             10 &             10 &             10 &             10 \\
   Layer-wise LR Decay &           0.65 &            0.7 &            0.8 &           0.86 \\
  EMA Weight Smoothing &         0.9998 &         0.9998 &         0.9999 &         0.9999 \\
          Weight Decay &           0.05 &           0.05 &           0.02 &           0.02 \\
       Label Smoothing &            0.1 &            0.1 &            0.1 &            0.1 \\
          Stoch. Depth &            0.1 &            0.1 &            0.2 &            0.2 \\
               Dropout &              0 &              0 &              0 &              0 \\
     Gradient Clipping &              3 &              3 &              3 &              2 \\
Rand Augment (Uniform) & M=U(0, 8), N=2 & M=U(0, 8), N=2 & M=U(0, 9), N=2 & M=U(0, 8), N=3 \\
     Random Erase Prob &            0.3 &            0.3 &            0.3 &            0.3 \\
    Random Resize Crop &            Yes &            Yes &            Yes &            Yes \\
           Mixup Alpha &              0 &              0 &              0 &              0 \\
          Cutmix Alpha &              0 &              0 &              0 &              0 \\
          Color Jitter &              0 &              0 &              0 &              0 \\
\bottomrule
\end{tabular}
\caption{ImageNet-12k intermediate fine-tune hyper-parameters.}
\label{table:imagenet_ft12k_hparam}
\end{table}

\subsubsection{Control experiments}
\label{sec:appendix:control_exp}

\textbf{Batch size during pre-training.} To be able to train efficiently on a large number of GPUs (up to 1520 in this work), it is desired to maximize the local batch size for each GPU worker for performing data parallel distributed training. For this large amount of GPUs, it leads to training with global batch sizes of 86K-88K. As we would like to also re-use experiments that were already performed with smaller batch sizes of 32K-45K, we execute control experiments to reassure that varying batch size in those ranges does not alter observed model performance on downstream tasks strongly. The experiments summarized in Table~\ref{tab:batch_size} provide evidence that performance variation due to changes in batch size is small, in the range of $0.2-0.5\%$ across different settings, which is small enough not to distort the trends  observed in the effect of scale, where the changes are substantially larger.

\begin{table}[!htb]
\setlength\tabcolsep{3.0pt}
    \centering
    \small
    \begin{tabular}{lcc}\toprule
       Batch size /  & 32k/38k (L/14) & 64k (B/16) / \\
      Model & & 86k (+lr tune) \\ \midrule
    ViT B/32 & 62.9 & 63.37 \\
    %\midrule
%    Ours & LAION-400M & L/14 &  &  &  \\
    ViT B/16 & 67.34 & 67.86 \\
    ViT L/14 & 72.8 & 72.98 \\\bottomrule
    \end{tabular}
    \caption{
    Batch size control experiments, zero-shot ImageNet top-1 accuracy. Executed on LAION-400M, 13B samples seen (32 full epochs).
    \vspace{-1em}
    }
    \label{tab:batch_size}
\end{table}

% Testing dependence of performance on batch size during pre-training.

% The main difference in training recipes is the batch size due to different compute environments, and our experiments with varying batch sizes suggest that the batch size changes do not explain the change in scaling trends.

% Compared to the original CLIP training procedure\citep{radford2021learning}, we work with larger batch sizes and adapt the learning rate accordingly.
%We opt for larger batch sizes to allow for more efficient distributed training; maximizing the local batch size per GPU and using close to one thousand GPUs lead us to global batch sizes in the range of 86-88K samples.
% In order to assess the validity of re-using measurements obtained with different batch sizes, we perform a number of control experiments varying batch size from 32K to 86-88K, and observe a difference of $0.2-0.5\%$ across different settings (see Appendix Sec. \ref{sec:appendix:control_exp}), which is small enough not to confound observations on the effect of scale.

\textbf{LAION-400M and 400M subset of LAION-2B size.} For 400M data scale, we are using LAION-400M dataset, as it was already validated by numerous previous works. This is not a subset of LAION-2B, as both were obtained by the same, but separately executed composition procedure using Common Crawl. To test that LAION-400M and LAION-2B can be considered as two different scale of the same data distribution, we extracted a random 400M subset from LAION-2B and conducted a pre-training experiment using our reference OpenCLIP ViT-B/32 model, 13B samples seen scale. We evaluated the pre-trained model on ImageNet zero-shot classification task, comparing it to same model pre-trained on LAION-400M. The outcome shows no significant difference between the performance of both models. This provides evidence that LAION-400M is comparable to a 400M subset extracted from LAION-2B, and can be thus considered to be a smaller scale of same data distribution.

\begin{table}[!htb]
\setlength\tabcolsep{4.0pt}
    \centering
    \small
    \begin{tabular}{lcc}\toprule
    Model/Dataset & 400M LAION-2B subset &  LAION-400M \\ \midrule
    ViT B/32 & 63.56 & 63.37 \\\bottomrule
    \end{tabular}
    \caption{
    400M data scale subset control experiments, zero-shot ImageNet top-1 accuracy. Executed either on 400M subset of LAION-2B or on LAION-400M, 13B samples seen (32 full epochs).
    \vspace{-1em}
    }
    \label{tab:subset_400M}
\end{table}

\textbf{Pre-training trial-to-trial variance.} To have a sanity check of trial-to-trial variance for model pre-training, we trained our reference ViT-B/32 model, 13B samples seen scale, for two trials using exactly the same hyper-parameters (lr=$0.001$, batch size 86K, warm up 2K). We evaluated the two trials on ImageNet zero-shot classification task. The result suggests a small variance of around 0.1\%, which is much smaller than variations observed when changing the scales. This allows us to conclude that scaling trends we observe are not distorted by variance caused by trial-to-trial pre-training.   
\begin{table}[!htb]
\setlength\tabcolsep{3.0pt}
    \centering
    \small
    \begin{tabular}{lcc}\toprule
       Trial  & ImageNet zero-shot top-1  \\ \midrule
        1 & 63.28 \\
        2 & 63.67 \\\bottomrule
    \end{tabular}
    \caption{
    Trial-to-trial variance control experiment. Executed on LAION-400M, 13B samples seen (32 full epochs) using ViT B/32 model.
    \vspace{-1em}
    }
    \label{tab:trial_to_trial}
\end{table}

\textbf{Resampling vs full shuffled training.} During our larger scale pre-training experiments featuring LAION-2B, it became important to allow for frequent checkpoint saving. Saving within a running epoch would require to memorize which samples were already seen, to be able to resume training in such a way that only previously not seen samples would be taken. To simplify the procedure, we have tested a version that does not perform epoch-wise training, taking a pre-defined number of samples instead for a virtual "step" through data. Such a resampling procedure can have repeated samples in the subset of data that contains in total the number of samples equal to number of samples in one full epoch through the dataset. As such training procedure differs from standard epoch-wise training, we conducted test experiments to check whether this results in differences in performance of pre-trained models when comparing to standard epoch-wise shuffling training. We trained our reference ViT-B/32 model and ViT-B/16 model on LAION-400M either using standard epoch-wise training with shuffling or the training that involves described resampling procedure. We observed only negligible differences of 0.1\%-0.3\%, concluding that using simple resampling cannot distort scaling trends observed in the study. 

\begin{table}[!htb]
\setlength\tabcolsep{3.0pt}
    \centering
    \small
    \begin{tabular}{lcc}\toprule
    Model & Resampling &  Full shuffling \\ \midrule
    ViT B/32 & 63.37 & 63.28; 63.67 \\\bottomrule
    \end{tabular}
    \caption{
    Resampling vs. full shuffling control experiments, zero-shot ImageNet top-1 accuracy. Executed on LAION-400M, 13B samples seen (32 full epochs).
    \vspace{-1em}
    }
    \label{tab:resampling_shuffling}
\end{table}

% corresponding to the number of samples that equals number of samples in 1 epoch through the dataset.

\subsubsection{Further detailed results and analysis}
\label{sec:appendix:further_results}

\begin{table}[!tbh]
\centering
\small
\rowcolors{2}{light-light-gray}{white}
%\setlength{\tabcolsep}{2pt}
%\resizebox{\linewidth}{!}{
\begin{tabular}{@{}lclll@{}}
\toprule
\textbf{Model}                     & \textbf{Samples seen} & \textbf{LAION-80M} & \textbf{LAION-400M} & \textbf{LAION-2B} \\ \midrule
\textbf{ViT-B/32} & 3B           & 38.05     & 41.53      & 43.66       \\
                          & 13B          & 42.30     & 46.18       & 45.50       \\
                          & 34B          & 42.10         & 46.41      &50.69        \\ \midrule
\textbf{ViT-B/16} & 3B           & 43.48     & 45.14      & 46.93       \\
                          & 13B          & 44.42     & 48.39       & 48.72       \\
                          & 34B          & 44.45         & 48.31      & 52.60        \\ \midrule                          
\textbf{ViT-L/14} & 3B           & 45.69     & 50.50      & 51.64       \\         
& 13B          & 46.36         & 51.51      & 53.01           \\
                          & 34B          & 45.70         & 52.83       & 54.63       \\ \midrule 
\textbf{ViT-H/14} & 34B           & -     & -      & 56.43       \\         \midrule 
\textbf{ViT-g/14} & 13B           & -     & -      & 56.54       \\         \bottomrule 

\end{tabular}

%}
\caption{Detailed results on VTAB+~\cite{Schuhmann2022} zero-shot classification, where we average over 35 tasks. 
}
\label{table:zeroshot_vtab_plus}
\end{table}

%TODO: here extended large tables with full details belong (eg VTAB+ full table as in NeurIPS paper) (DONE)

\begin{table}[ht]
\centering
\small
\rowcolors{2}{light-light-gray}{white}
\begin{tabular}{@{}lcccccc@{}}
\toprule
Dataset & B/32 (34B)&B/16 (34B)&L/14 (34B)&g/14 (13B)&H/14 (34B)\\  \midrule
INet & 66.47 &70.22 &75.20 &76.66 &\textbf{77.97 }\\
INet-v2 & 58.16 &62.28 &67.69 &69.61 &\textbf{70.82 }\\
INet-R & 76.47 &80.59 &87.41 &88.65 &\textbf{89.32 }\\
INet-S & 53.72 &56.09 &63.28 &65.22 &\textbf{66.57 }\\
ObjNet & 48.78 &56.05 &65.50 &67.47 &\textbf{69.70 }\\
INet-A & 25.43 &38.23 &53.88 &57.11 &\textbf{59.23 }\\
CIFAR-10 & 93.65 &94.94 &96.64 &97.05 &\textbf{97.42 }\\
CIFAR-100 & 75.47 &76.83 &83.36 &83.91 &\textbf{84.68 }\\
MNIST & 67.73 &65.99 &54.87 &69.04 &\textbf{72.94 }\\
Flowers102 & 72.35 &71.23 &75.90 &77.61 &\textbf{80.21 }\\
Cars & 86.15 &88.50 &92.61 &92.77 &\textbf{93.46 }\\
SVHN & 43.51 &51.39 &46.30 &\textbf{60.33 }&56.13 \\
FER2013 & 46.02 &51.78 &\textbf{53.71 }&46.57 &51.76 \\
RenderedSST2 & 57.17 &59.80 &59.31 &\textbf{64.58 }&64.09 \\
Pets & 89.81 &90.52 &93.21 &94.28 &\textbf{94.39 }\\
Caltech-101 & 83.50 &83.83 &85.04 &\textbf{85.22 }&85.04 \\
VOC2007-Cl & 79.75 &78.85 &80.52 &\textbf{81.03 }&77.61 \\
SUN397 & 68.57 &70.85 &74.33 &\textbf{75.40 }&75.22 \\
FGVC Aircraft & 24.06 &27.00 &36.93 &37.80 &\textbf{42.75 }\\
Country211 & 16.78 &20.31 &26.36 &28.73 &\textbf{30.01 }\\
DTD & 55.64 &56.33 &62.77 &\textbf{68.14 }&67.87 \\
GTSRB & 49.49 &48.24 &56.10 &49.74 &\textbf{58.45 }\\
STL10 & 96.55 &97.86 &\textbf{98.86 }&98.59 &98.44 \\
Retino & \textbf{73.42 }&67.96 &21.06 &43.42 &23.80 \\
EuroSAT & 46.94 &53.46 &65.15 &64.80 &\textbf{71.74 }\\
RESISC45 & 60.71 &62.76 &66.67 &\textbf{71.71 }&69.57 \\
PCAM & \textbf{59.44 }&56.37 &55.26 &55.09 &53.63 \\
CLEVR Counts & 15.02 &21.49 &31.09 &\textbf{33.19 }&27.84 \\
CLEVR Dist & 14.54 &\textbf{21.07 }&16.10 &17.73 &16.77 \\
DSPRITES Orient & \textbf{3.77 }&2.68 &2.00 &3.08 &2.61 \\
DSPRITES pos & 2.80 &3.30 &3.15 &\textbf{3.54 }&3.14 \\
SmallNORB Elv & \textbf{11.70 }&11.30 &10.95 &11.34 &11.13 \\
SmallNORB Azim & 5.86 &5.67 &5.63 &\textbf{5.88 }&5.50 \\
DMLAB & 17.48 &19.93 &\textbf{22.43 }&19.02 &14.20 \\
KITTI Dist & \textbf{27.14 }&17.16 &22.93 &14.63 &11.11 \\

\midrule
\textbf{VTAB+ (Avg.)} & 50.69 &52.60 &54.63 &\textbf{56.54} &56.43 \\
\bottomrule
\end{tabular}
\caption{Detailed zero-shot top-1 classification results of LAION-2B models on VTAB+ 35 tasks. We highlight the best results for each downstream dataset.}
\label{table:vtab_plus_details}
\end{table}

\textbf{Consistency of scaling trends for CLIP and openCLIP}

\begin{figure*}
    \centering
    \includegraphics[width=\linewidth]{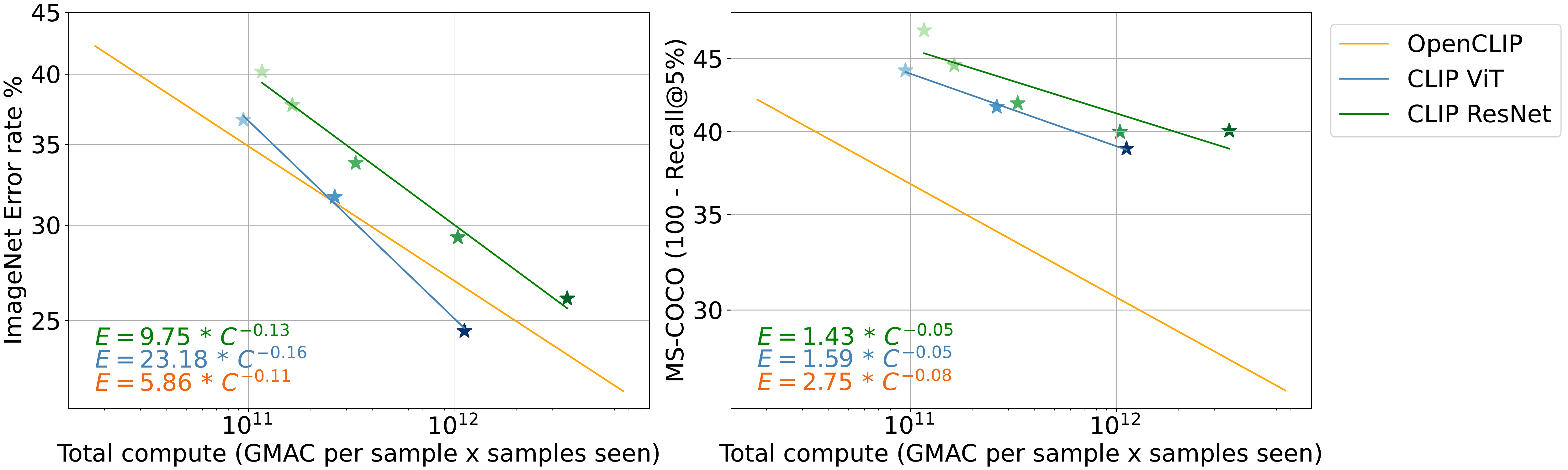}
    \caption{Consistency of scaling trends for CLIP and openCLIP with additional CLIP ResNet models from OpenAI and trained on WIT. Relationship between total compute (GMAC) and ImageNet zero-shot classification error rate \textbf{(Left)} and with MS-COCO image retrieval at Recall@5 \textbf{(Right)}.}
    \label{fig:resnet_scaling_trends}
\end{figure*}

In Fig.~\ref{fig:resnet_scaling_trends}, we complement the results we found in Fig.~\ref{fig:fig1} and  show scaling trends with additional ResNet models from OpenAI trained on the WebImageText (WIT) dataset. Despite different model architectures, we see the same consistent pattern - both OpenAI’s ResNets and ViT trained on WIT demonstrate stronger scaling on zero-shot classification but worse scaling on retrieval than openCLIP trained on LAION datasets.

\textbf{Details of zero-shot classification results.} Complementing results from the Section \ref{subsect:scaling_laws}, we provide summary tables for the performance measure on different downstream tasks:  ImageNet (Tab.~\ref{table:zeroshot_im1k}), ImageNet robustness(Tab.~\ref{table:zeroshot_imrobustness}), MS-COCO image retrieval (Tab.~\ref{table:zeroshot_mscoco_image_retrieval}) and text retrieval (Tab.~\ref{table:zeroshot_mscoco_text_retrieval}), Flickr30K image retrieval (Tab.~\ref{table:zeroshot_flickr30k_image_retrieval}) and text retrieval (Tab.~\ref{table:zeroshot_flickr30k_text_retrieval}), and VTAB+ (Tab.~\ref{table:zeroshot_vtab_plus} and ~\ref{table:vtab_plus_details}).

\textbf{Details of linear probing results}. To supplement Figures~\ref{fig:lp1} and \ref{fig:lp2}, we provide the corresponding Table~\ref{tab:lp} with detailed results.

%\textbf{Details of fine-tuning results}. To supplement Figure, we provide the corresponding Tablw~\ref{table:fi} with full details.
%ToDO: GMAC for architecture hyperparam table (DONE)

%TODO: full table on model/data/sample seen scales for zero shot im1k (DONE)

%ToDo : full table on model/data/sample seen scales for zero shot retrieval  (Flickr30k) (DONE)

%ToDo Mehdi, Romain, Ross, Jenia: table on training hyperparams. Table \ref{table:openclip_hyper_parameters}. Introduce samples seen, and GPU-hours (DONE)

\textbf{Architecture and training hyperparameters.} We provide overview for architecture (Tab.~\ref{table:openclip_architectures}) and pre-training hyper-parameters (Tab.~\ref{table:openclip_hyper_parameters}) that we have used in the experiments. 

\begin{table}[ht]
\centering
\setlength\tabcolsep{5pt}
\footnotesize
\rowcolors{2}{light-light-gray}{white}
\begin{tabular}{@{}llrrrrrrr@{}}
\toprule
\textbf{Model} & \textbf{Dataset} & \textbf{BS. (global)} & \textbf{LR.} & \textbf{Warm.} & \textbf{\#samples}. &  \textbf{\#GPUs} & \textbf{Time (hrs.)} & \textbf{GPU-h}/\textbf{MWh} \\ \midrule
%B/32 (400M)                & 256 (32768)   & $5\text{e-}4$          & 2K         & 32   & 128 & 36          & 4608         \\
%B/32 (2B)                  & 416 (46592)   & $5.5\text{e-}4$        & 10K        & 16   & 112    & 210      &  23520          \\
%B/16 (400M)                & 192 (33792)   & $5\text{e-}4$          & 2K         & 32   & 176  & 61 &    10736               \\
%L/14 (400M)                & 96 (38400)    & $6\text{e-}4$          & 5K         & 32  & 400 & 88   &     35200             \\ \bottomrule
%L/14 (2B)                  & 224 (8601)         & 384    & $1\text{e-}3$          & 10K        & 16     & 88                     \\ \bottomrule
B/32 & LAION-80M & 256(32768) & 5e-4 & 2K & 3B & 128 & 7 & 836/0.29 \\
B/32 & LAION-80M & 256(32768) & 5e-4 & 2K & 13B & 128 & 33 & 4181/1.46 \\
B/32 & LAION-80M & 256(88064) & 1e-3 & 10K & 34B & 344 & 96 & 32953/11.53 \\
B/32 & LAION-400M & 256(88064) & 1e-3 & 10K & 3B & 344 & 3 & 1063/0.37 \\
B/32 & LAION-400M & 672(86016) & 1e-3 & 2K & 13B & 128 & 70 & 8912/3.12 \\
B/32 & LAION-400M & 256(32768) & 5e-4 & 2K & 34B & 128 & 87 & 11177/3.91 \\
B/32 & LAION-2B & 256(88064) & 1e-3 & 10K & 3B & 344 & 3 & 1121/0.39 \\
B/32 & LAION-2B & 256(32768) & 5e-4 & 2K & 13B & 128 & 39 & 4954/1.73 \\
B/32 & LAION-2B & 96(79104) & 1e-3 & 2K & 34B & 824 & 51 & 42307/14.81 \\
B/16 & LAION-80M & 256(88064) & 1e-3 & 10K & 3B & 344 & 6 & 1900/0.66 \\
B/16 & LAION-80M & 512(90112) & 1e-3 & 10K & 13B & 176 & 71 & 12518/4.38 \\
B/16 & LAION-80M & 256(88064) & 1e-3 & 10K & 34B & 344 & 70 & 24032/8.41 \\
B/16 & LAION-400M & 256(88064) & 1e-3 & 10K & 3B & 344 & 5 & 1713/0.60 \\
B/16 & LAION-400M & 192(33792) & 5e-4 & 10K & 13B & 176 & 61 & 10736/3.76 \\
B/16 & LAION-400M & 512(90112) & 1e-3 & 10K & 34B & 176 & 148 & 26009/9.10 \\
B/16 & LAION-2B & 256(88064) & 1e-3 & 10K & 3B & 344 & 5 & 1822/0.64 \\
B/16 & LAION-2B & 512(90112) & 1e-3 & 10K & 13B & 176 & 66 & 11675/4.09 \\
B/16 & LAION-2B & 256(88064) & 1e-3 & 10K & 34B & 344 & 121 & 41726/14.60 \\
L/14 & LAION-80M & 224(88704) & 1e-3 & 10K & 3B & 396 & 18 & 7243/2.54 \\
L/14 & LAION-80M & 448(89600) & 1e-3 & 10K & 13B & 200 & 102 & 20393/7.14 \\
L/14 & LAION-80M & 224(89600) & 1e-3 & 10K & 34B & 400 & 227 & 90647/31.73 \\
L/14 & LAION-400M & 224(88704) & 1e-3 & 10K & 3B & 396 & 17 & 6717/2.35 \\
L/14 & LAION-400M & 112(86016) & 1e-3 & 2K & 13B & 768 & 61 & 46735/16.36 \\
L/14 & LAION-400M & 84(86016) & 1e-3 & 10K & 34B & 1024 & 122 & 124727/43.65 \\
L/14 & LAION-2B & 224(88704) & 1e-3 & 10K & 3B & 396 & 18 & 7055/2.47 \\
L/14 & LAION-2B & 84(86016) & 1e-3 & 10K & 13B & 1024 & 52 & 53599/18.76 \\
L/14 & LAION-2B & 224(86016) & 1e-3 & 10K & 34B & 384 & 319 & 122509/42.88 \\
H/14 & LAION-2B & 96(79104) & 5e-4 & 2K & 34B & 824 & 279 & 229665/80.38 \\
g/14 & LAION-2B & 80(64000) & 5e-4 & 2K & 13B & 800 & 137 & 109392/38.29 \\ \midrule
Total &  &  &  &  &  &  &  & 1058318/370.41 \\

\bottomrule
\end{tabular}
\caption{Training hyper-parameters and resources used to for pre-training our models on LAION 80M, 400M, and 2B subsets. Note that \textbf{BS} refer to batch size per GPU worker (with \textbf{global} the corresponding global batch size), \textbf{LR}  to base learning rate, \textbf{Warm} to the total number of warmup steps, \textbf{Time} to total training time in hours, \textbf{GPU-h} to GPU hours, \textbf{MWh} to the total energy consumed in Megawatt hours.}
\label{table:openclip_hyper_parameters}
\end{table}

\begin{table}[tb]
\small
\centering
\rowcolors{2}{light-light-gray}{white}
%\setlength{\tabcolsep}{2pt}
% \resizebox{\linewidth}{!}{
\small
%\subfloat[width=.5\linewidth]{
\begin{tabular}{@{}lclll@{}}
\toprule
\textbf{Model}                     & \textbf{Samples seen} & \textbf{LAION-80M} & \textbf{LAION-400M} & \textbf{LAION-2B} \\ \midrule
\textbf{ViT-B/32} & 3B           & 51.94     & 57.12      & 57.36       \\
                          & 13B          & 56.46     & 63.23       & 62.53       \\
                          & 34B          & 56.43         & 64.06      &66.47        \\ \midrule
\textbf{ViT-B/16} & 3B           & 57.55     & 62.68      & 61.82       \\
                          & 13B          & 60.24     & 67.00       & 68.13       \\
                          & 34B          & 61.28         & 69.00      & 70.22        \\ \midrule                          
\textbf{ViT-L/14} & 3B           & 61.14     & 69.31      & 68.93       \\         
& 13B          & 63.96         & 73.06      & 73.10           \\
                          & 34B          & 64.83         & 73.94       & 75.20       \\ \midrule 
\textbf{ViT-H/14} & 34B           & -     & -      & 77.97       \\         \midrule 
\textbf{ViT-g/14} & 13B           & -     & -      & 76.66       \\         \bottomrule 

\end{tabular}
%}
%\subfloat[]{
%}
% }
\caption{Detailed results on ImageNet zero-shot accuracy.
}
\label{table:zeroshot_im1k}
\end{table}

\begin{table}[tb]
\small
\centering
\rowcolors{2}{light-light-gray}{white}
%\setlength{\tabcolsep}{2pt}
% \resizebox{\linewidth}{!}{
\small
\begin{tabular}{@{}lclll@{}}
\toprule
\textbf{Model}                     & \textbf{Samples seen} & \textbf{LAION-80M} & \textbf{LAION-400M} & \textbf{LAION-2B} \\ \midrule
\textbf{ViT-B/32} & 3B           & 37.95     & 41.60      & 42.44       \\
                          & 13B          & 42.23     & 48.97       & 48.83       \\
                          & 34B          & 43.01         & 50.12      &52.51        \\ \midrule
\textbf{ViT-B/16} & 3B           & 43.48     & 47.82      & 48.07       \\
                          & 13B          & 47.29     & 54.89       & 55.89       \\
                          & 34B          & 49.29         & 57.14      & 58.65        \\ \midrule                          
\textbf{ViT-L/14} & 3B           & 48.26     & 57.53      & 57.56       \\         
& 13B          & 52.23         & 63.84      & 64.61           \\
                          & 34B          & 54.23         & 65.25       & 67.55       \\ \midrule 
\textbf{ViT-H/14} & 34B           & -     & -      & 71.13       \\         \midrule 
\textbf{ViT-g/14} & 13B           & -     & -      & 69.61       \\         \bottomrule 

\end{tabular}
% }
\caption{Detailed results on ImageNet five robustness datasets zero-shot accuracy (average over the five datasets is reported).
}
\label{table:zeroshot_imrobustness}
\end{table}

\begin{table}[tb]

\rowcolors{2}{light-light-gray}{white}
%\setlength{\tabcolsep}{2pt}
% \resizebox{\linewidth}{!}{
\small
\centering
\begin{tabular}{@{}lclll@{}}
\toprule
\textbf{Model}                     & \textbf{Samples seen} & \textbf{LAION-80M} & \textbf{LAION-400M} & \textbf{LAION-2B} \\ \midrule
\textbf{ViT-B/32} & 3B           & 51.04     & 56.29      & 57.01       \\
                          & 13B          & 54.67     & 61.90       & 61.66       \\
                          & 34B          & 54.72         & 62.28      &65.05        \\ \midrule
\textbf{ViT-B/16} & 3B           & 55.83     & 60.85      & 61.08       \\
                          & 13B          & 57.83     & 63.64       & 66.11       \\
                          & 34B          & 58.84         & 65.81      & 67.73        \\ \midrule                          
\textbf{ViT-L/14} & 3B           & 58.42     & 65.63      & 66.21       \\         
& 13B          & 59.18         & 68.40      & 69.16           \\
                          & 34B          & 59.84         & 68.62       & 71.08       \\ \midrule 
\textbf{ViT-H/14} & 34B           & -     & -      & 73.43       \\         \midrule 
\textbf{ViT-g/14} & 13B           & -     & -      & 72.40       \\         \bottomrule 

\end{tabular}

% }
\caption{Detailed results on MS-COCO image retrieval Recall@5. 
%We vary model scale (ViT-B/32, ViT-B/16 and ViT-L/14), samples seen budget (3B, 13B, and 34B images seen) and LAION dataset scale (80M, 400M, 2B). When investing enough into training compute, seeing same number of samples on larger data scale leads consistently to better zero-shot transfer performance. Bottlenecks.
}
\label{table:zeroshot_mscoco_image_retrieval}
\end{table}

\begin{table}[tb]

\rowcolors{2}{light-light-gray}{white}
%\setlength{\tabcolsep}{2pt}
% \resizebox{\linewidth}{!}{
\centering
\small
\begin{tabular}{@{}lclll@{}}
\toprule
\textbf{Model}                     & \textbf{Samples seen} & \textbf{LAION-80M} & \textbf{LAION-400M} & \textbf{LAION-2B} \\ \midrule
\textbf{ViT-B/32} & 3B           & 67.16     & 73.38      & 73.10       \\
                          & 13B          & 70.32     & 77.60       & 77.04       \\
                          & 34B          & 70.78         & 77.46      &79.58        \\ \midrule
\textbf{ViT-B/16} & 3B           & 72.22     & 77.18      & 76.72       \\
                          & 13B          & 73.84     & 79.62       & 81.00       \\
                          & 34B          & 74.12         & 80.52      & 81.78        \\ \midrule                          
\textbf{ViT-L/14} & 3B           & 74.90     & 80.78      & 79.86       \\         
& 13B          & 76.24         & 82.12      & 82.94           \\
                          & 34B          & 75.96         & 83.44       & 84.00       \\ \midrule 
\textbf{ViT-H/14} & 34B           & -     & -      & 86.04       \\         \midrule 
\textbf{ViT-g/14} & 13B           & -     & -      & 85.36       \\         \bottomrule 

\end{tabular}

% }
\caption{Detailed results on MS-COCO text retrieval Recall@5. 
%We vary model scale (ViT-B/32, ViT-B/16 and ViT-L/14), samples seen budget (3B, 13B, and 34B images seen) and LAION dataset scale (80M, 400M, 2B). When investing enough into training compute, seeing same number of samples on larger data scale leads consistently to better zero-shot transfer performance. Bottlenecks.
}
\label{table:zeroshot_mscoco_text_retrieval}
\end{table}

\begin{table}[tb]

\rowcolors{2}{light-light-gray}{white}
%\setlength{\tabcolsep}{2pt}
% \resizebox{\linewidth}{!}{
\small
\centering
\begin{tabular}{@{}lclll@{}}
\toprule
\textbf{Model}                     & \textbf{Samples seen} & \textbf{LAION-80M} & \textbf{LAION-400M} & \textbf{LAION-2B} \\ \midrule
\textbf{ViT-B/32} & 3B           & 76.00     & 80.50      & 82.16       \\
                          & 13B          & 78.46     & 85.20       & 85.36       \\
                          & 34B          & 78.98         & 85.90      &88.26        \\ \midrule
\textbf{ViT-B/16} & 3B           & 80.78     & 85.84      & 85.12       \\
                          & 13B          & 84.76     & 88.16       & 89.90       \\
                          & 34B          & 84.38         & 89.58      & 90.32        \\ \midrule                          
\textbf{ViT-L/14} & 3B           & 84.16     & 89.14      & 89.82       \\         
& 13B          & 84.86         & 91.04      & 91.72           \\
                          & 34B          & 85.70         & 91.28       & 92.92       \\ \midrule 
\textbf{ViT-H/14} & 34B           & -     & -      & 94.10       \\         \midrule 
\textbf{ViT-g/14} & 13B           & -     & -      & 93.48       \\         \bottomrule 

\end{tabular}

% }
\caption{Detailed results on Flickr30K image retrieval Recall@5.
}
\label{table:zeroshot_flickr30k_image_retrieval}
\end{table}

\begin{table}[tb]

\rowcolors{2}{light-light-gray}{white}
%\setlength{\tabcolsep}{2pt}
% \resizebox{\linewidth}{!}{
\centering
\small
\begin{tabular}{@{}lclll@{}}
\toprule
\textbf{Model}                     & \textbf{Samples seen} & \textbf{LAION-80M} & \textbf{LAION-400M} & \textbf{LAION-2B} \\ \midrule
\textbf{ViT-B/32} & 3B           & 88.20     & 91.60      & 92.70       \\
                          & 13B          & 91.30     & 95.60       & 94.50       \\
                          & 34B          & 90.70         & 95.60      &96.10        \\ \midrule
\textbf{ViT-B/16} & 3B           & 91.90     & 95.60      & 94.60       \\
                          & 13B          & 94.90     & 96.80       & 97.60       \\
                          & 34B          & 94.80         & 97.40      & 98.00        \\ \midrule                          
\textbf{ViT-L/14} & 3B           & 93.60     & 97.80      & 96.70       \\         
& 13B          & 95.00         & 98.30      & 98.40           \\
                          & 34B          & 96.90         & 97.70       & 98.70       \\ \midrule 
\textbf{ViT-H/14} & 34B           & -     & -      & 99.30       \\         \midrule 
\textbf{ViT-g/14} & 13B           & -     & -      & 99.10       \\         \bottomrule 

\end{tabular}

% }
\caption{Detailed results on Flickr30K text retrieval Recall@5. 
}
\label{table:zeroshot_flickr30k_text_retrieval}
\end{table}

\iffalse
\begin{table}[tb]

\rowcolors{2}{light-light-gray}{white}
%\setlength{\tabcolsep}{2pt}
\resizebox{\linewidth}{!}{
\small
\begin{tabular}{@{}lclll@{}}
\toprule
\textbf{Model}                     & \textbf{Samples seen} & \textbf{LAION-80M} & \textbf{LAION-400M} & \textbf{LAION-2B} \\ \midrule
\textbf{ViT-B/32} & 3B           & 32.28     & 34.92      & 37.50       \\
                          & 13B          & 36.63     & 38.75       & 37.92       \\
                          & 34B          & 36.00         & 38.55      &43.56        \\ \midrule
\textbf{ViT-B/16} & 3B           & 38.16     & 37.90      & 40.70       \\
                          & 13B          & 37.55     & 40.00       & 39.20       \\
                          & 34B          & 37.69         & 38.06      & 44.43        \\ \midrule                          
\textbf{ViT-L/14} & 3B           & 39.56     & 41.61      & 43.62       \\         
& 13B          & 38.79         & 41.42      & 43.26           \\
                          & 34B          & 37.15         & 43.20       & 44.39       \\ \midrule 
\textbf{ViT-H/14} & 34B           & -     & -      & 44.98       \\         \midrule 
\textbf{ViT-g/14} & 13B           & -     & -      & 46.75       \\         \bottomrule 

\end{tabular}
}
\caption{Detailed results on VTAB~\cite{Zhai2019} zero-shot classification, where we average over 19 tasks. 
}
\label{table:zeroshot_vtab}
\end{table}
\fi

\begin{table}[h!]
\centering
\rowcolors{2}{light-light-gray}{white}

\begin{tabular}{@{}lllllll@{}}
%\begin{NiceTabular}{@{}lllllll@{}}
%\CodeBefore
%\Body
\toprule
\textbf{Name} & \textbf{Width}  & \textbf{Emb.} & \textbf{Depth} & \textbf{Acts.} & \textbf{Params} & \textbf{GMAC} \\ \midrule
ViT-B/32      & 768 / 512                 & 512                & 12 / 12    & 10 M  & 151 M & 7.40 \\
ViT-B/16      & 768 / 512                 & 512                & 12 / 12    & 29 M  & 150 M & 20.57\\
ViT-L/14     & 1024 / 768                 & 768                & 24 / 12    & 97 M  & 428 M & 87.73 \\ 
ViT-H/14     & 1280 / 1024                & 1024               & 32 / 24    & 161 M  & 986 M & 190.97\\ 
ViT-g/14     & 1408 / 1024                & 1024               & 40 / 24    & 214 M  & 1.37 B & 290.74 \\ 
ViT-G/14     & 1664 / 1280                & 1280               & 48 / 32    & 310 M  & 2.54 B & 532.92 \\ 
\bottomrule
%\end{NiceTabular}
\end{tabular}
\caption{Hyper-parameters of different architectures we consider. \textbf{Emb} refers to embedding size, \textbf{Acts} refers to the number of activations in millions, and \textbf{Params} refers to the number of parameters in millions. \textbf{GMAC} refers to giga multiply–accumulates. All entries in the form of A / B denote image and text parameters respectively.} 
\label{table:openclip_architectures}
\end{table}

\iffalse
\begin{table}[h]
%\vspace{-6pt}
\begingroup
%\setlength{\tabcolsep}{3pt}
\begin{tabular}{l@{\hskip 10pt}lllll@{\hskip 15pt}rrrr}
\toprule
& \multicolumn{2}{c}{Architecture} & Batch & Examples seen & Parameters& \multicolumn{4}{c}{ImageNet top-1 \%} \\
                    & Image & Text & size  &               & per token & Test & V2 & R & A \\
\midrule
COCA$^\mathsection$~\cite{yu2022coca} & ViT-g  & T-g & 65k & 32.8B & 2.1B & 86.3   & 80.7      & 96.5    & 90.2   \\
BASIC~\cite{pham2022basic}             & CoAtNet-7$^*$ & T-H$^{*}$ & 65k  & 19.7B\textsuperscript{PT} +32.8B & 3B  & 85.7 & 80.6      & 95.7    & 85.6   \\
LIT~\cite{zhai2021lit} & ViT-g$^*$ & T-g & 32k & 25.8B\textsuperscript{PT} + 18.2B & 2.1B & 84.5   & 78.7      & 93.9    & 79.4\\
ALIGN~\cite{jia2021align}  & EffNet-L2 & T-L$^*$ & 16k & 19.8B & $\sim$ 820M & 76.4        & 70.1       & 92.2       & 75.8        \\
CLIP~\cite{radford2021clip} & ViT-L/14$^\dagger$  & T-B & 32k & 12.8B & $\sim$ 400M & 76.2        & 70.1       & 88.9       & 77.2        \\
\midrule
\texttt{LIMoE}                                  & \multicolumn{2}{c}{H/14} & 21k & 23.3B & 675M & -       & 77.7       & 94.9       & 78.7       \\
\bottomrule
\end{tabular}
\endgroup
\centering
\caption{From \url{https://arxiv.org/pdf/2206.02770.pdf}}%
\label{tab:sota_models}

\end{table}
\fi

% \clearpage
\section{Code and Data availability}
\label{appendix:code}
We will provide source code used for running experiments and producing figures in this study at \url{https://github.com/LAION-AI/scaling-laws-openclip}. Links to pre-trained models obtained in this study and links to instructions for obtaining LAION-400m and LAION-5B used for pre-training experiments will be also made available there. All datasets used in the study are openly available and are listed together with references to the original work in Table \ref{tab:datasets_list}.

\begin{table}[!h]
\centering
\rowcolors{2}{light-light-gray}{white}
\begin{tabular}{@{}llll@{}}
%\begin{NiceTabular}{@{}llll@{}}
%\CodeBefore
%\rowcolors{2}{light-light-gray}{white}
%\Body
\toprule
\bf{Dataset} & \bf{Abbr.}  & \bf{Test size} & \bf{\#Classes}\\\midrule
ImageNet & INet & 50,000 &1,000\\
ImageNet-v2 & INet-v2 & 10,000 &1,000\\
ImageNet-R & INet-R & 30,000 &200\\
ImageNet Sketch & INet-S & 50,889 &1,000\\
ObjectNet & ObjNet & 18,574 &113\\
ImageNet-A & INet-A & 7,500 &200\\
CIFAR-10 & - & 10,000 &10\\
CIFAR-100 & - & 10,000 &100\\
MNIST & - & 10,000 &10\\
Oxford Flowers 102 & Flowers102 & 6,149 &102\\
Stanford Cars & Cars & 8,041 &196\\
SVHN & - & 26,032 &10\\
Facial Emotion Recognition 2013 & FER2013 & 7,178 &7\\
RenderedSST2 & - & 1,821 &2\\
Oxford-IIIT Pets & Pets & 3,669 &37\\
Caltech-101 & - & 6,085 &102\\
Pascal VOC 2007 Classification & VOC2007-Cl & 14,976 &20\\
SUN397 & - & 108,754 &397\\
FGVC Aircraft & - & 3,333 &100\\
Country211 & - & 21,100 &211\\
Describable Textures & DTD & 1,880 &47\\
GTSRB & - & 12,630 &43\\
STL10 & - & 8,000 &10\\
Diabetic Retinopathy & Retino & 42,670 &5\\
EuroSAT & - & 5,400 &10\\
RESISC45 & - & 6,300 &45\\
PatchCamelyon & PCAM & 32,768 &2\\
CLEVR Counts & - & 15,000 &8\\
CLEVR Object Distance & CLEVR Dist & 15,000 &6\\
DSPRITES Orientation & DSPRITES Orient & 73,728 &40\\
DSPRITES Position & DSPRITES pos & 73,728 &32\\
SmallNORB Elevation & SmallNORB Elv & 12,150 &9\\
SmallNORB Azimuth & SmallNORB Azim & 12,150 &18\\
DMLAB & - & 22,735 &6\\
KITTI closest vehicle distance & KITTI Dist & 711 &4\\ \midrule
%\midrule
%\multicolumn{2}{l}{Retrieval datasets} & & & \\
MS-COCO & - & 5,000 & -\\
Flickr30K & - & 1,000 & -\\
\bottomrule
%\end{NiceTabular}
\end{tabular}
\caption{Datasets used for evaluating downstream performance. Adapted from~\citep{Schuhmann2022}.}
\label{tab:datasets_list}
\end{table}

% Further details on the usage of the datasets in the conducted experiments are also provided in the repository linked above.
% \url{Link will be made available in final version}

% Only for final version
%\section*{Author Contributions}

% Only for final version
%\section*{Further Acknowledgements}

\section*{Broader and Social Impact}
\label{sec:impact}
\textbf{Safety aspect.} Our work deals with studying function and properties of pre-trained models on large scales. Releasing these models to public can have both positive and negative implications, like with any research artefact that possesses generic functionality. We would like to stress that we consider the released pre-trained language-vision models as research artefacts that are there to advance the studies of scaling laws and allow analysis of the properties and behavior of such models for the broader research community. These models are not meant to be incorporated into end products or even used for applications in sensitive areas like interpretation of medical imaging in hospitals or security surveillance. There is potential for abuse of technology based on large-scale pre-trained generalist models, and it is the task of democratic institutions to work out rules for sensitive applications that might involve those. Open release of models gives the broad research community also opportunity to  study safety related aspects of such models, such to preventively design measures that make such abuse by malicious parties less probable, in a common transparent effort. Same applies to the common effort of studying yet not systematically understood biases that such models may contain due to pre-training on either largely uncurated, imbalanced data or on data filtered by models that already contain unknown biases (like OpenAI's CLIP that was trained on the private WIT-400M dataset), and due to the simplistic nature of the contrastive InfoNCE loss that drives learning.

\textbf{Energy cost.} There is high computational cost bound to pre-training experiments on large scale. Supercomputers used in our studies are highly ranked in the Green Top-500 list, ensuring that energy costs are dampened. In addition, strongly transferable pre-trained models save energy on numerous downstream tasks where they can perform in data-efficient and thus in an energy saving manner. Releasing such pre-trained models to public incurs additional energy savings, as research community can re-use already validated models without necessity to train those from scratch again.

% Only for final version
\section*{Author Contributions}
\begin{itemize}
\item \textbf{Mehdi Cherti}: Planned and executed experiments on JUWELS Booster and Stability AI HPC, coordinated and performed results collection, distillation and analysis, performed deduplication analysis on LAION and downstream targest datasets, manuscript writing and revision.
\item \textbf{Romain Beaumont}: Planned and executed experiments on JUWELS Booster and Stability AI HPC, pioneered training of large scale openCLIP ViT H-14 and g-14, manuscript revision.
\item \textbf{Ross Wightman}: Planned and executed experiments on JUWELS Booster and Stability AI HPC, performed full fine-tuning experiments, manuscript writing and revision.
\item \textbf{Mitchell Wortsman}: experiment design and planning, performed experiments evaluating linear probing fine-tuning performance and robustness on ImageNet and other downstream datasets, manuscript writing and revision.
\item \textbf{Gabriel Ilharco}: experiment design and planning, performed experiments evaluating full fine-tuning performance and robustness on ImageNet and other downstream datasets, manuscript writing and revision.
\item \textbf{Cade Gordon}: implemented local contrastive loss for efficient distributed training, manuscript writing and revision.
\item \textbf{Christoph Schuhmann}: supervised larger scale experiments, resource and community organization, compute and storage resource acquisition, manuscript revision.
\item \textbf{Ludwig Schmidt}: provided advice on experiment design and study directions, manuscript writing and revision, general supervision
\item \textbf{Jenia Jitsev}: led the project; conducted experiments on various scales using JUWELS Booster and Stability AI HPC, scientific organization \& manuscript writing, ethical and social content, experiments planning and design, compute and storage resource acquisition, general supervision.

\end{itemize}

% Only for final version
% \section*{Acknowledgements}

\end{appendix}

\end{document}